\documentclass[journal]{IEEEtran}
\PassOptionsToPackage{hyphens}{url}
\newcommand{\underreview}{0}
\newcommand{\final}{0}

\usepackage[pagebackref=false,breaklinks=true,colorlinks,bookmarks=false,citecolor=blue,linkcolor=blue]{hyperref}
\usepackage{times}
\usepackage{epsfig}
\usepackage{graphicx}
\usepackage{amsmath}
\usepackage{amssymb}
\usepackage{ifthen}
\usepackage{float}
\usepackage{morefloats}
\usepackage{alltt}
\usepackage{amsmath}
\usepackage{amssymb}
\usepackage{amsthm}
\usepackage{ulem}
\usepackage{mathrsfs}
\usepackage[utf8]{inputenc}
\usepackage{newlfont} 
\usepackage{floatflt}
\usepackage{wrapfig}
\usepackage[ruled]{algorithm2e} 
\usepackage{booktabs} 
\usepackage{gensymb}
\usepackage{fixltx2e}
\usepackage{subfig} 
\usepackage{multirow}
\usepackage{cleveref}
\Crefname{equation}{Eq.}{Eqs.}
\Crefname{figure}{Fig.}{Figs.}
\Crefname{section}{Sec.}{Sec.}
\usepackage[T1]{fontenc} 
\usepackage{etoolbox}
\usepackage{xfrac}
\usepackage{booktabs} 
\usepackage[export]{adjustbox}
\usepackage{algpseudocode}
\usepackage[acronym,nomain]{glossaries}
\usepackage{adjustbox}
\usepackage{xcolor}
\usepackage{wrapfig}
\usepackage[percent]{overpic}
\usepackage{soul}
\usepackage{paralist}
\usepackage{pict2e}

\newcommand{\etal}{{et~al.}}

\definecolor{YodaColor}{rgb}{0,0.6,0} 
\definecolor{ChiakaiColor}{rgb}{0.30, 0.66, 0.37} 
\definecolor{YichangColor}{rgb}{0.70, 0.30, 0.20} 
\definecolor{MinghsuanColor}{rgb}{0.8, 0.50, 0.10} 
\definecolor{JasonColor}{rgb}{0.84 0.31 0.87} 
\definecolor{ReviseColor}{rgb}{0 0 1} 
\definecolor{DarkGreen}{rgb}{0.008, 0.294, 0.188}

\newcommand{\ckliang}[1]{{\color{ChiakaiColor} Chia-Kai: #1 $\qed$}}
\newcommand{\yichang}[1]{{\color{YichangColor} Yichang: #1 $\qed$}}
\newcommand{\minghsuan}[1]{{\color{MinghsuanColor} Ming-Hsuan: #1 $\qed$}}
\newcommand{\wslai}[1]{{\color{JasonColor} Jason: #1 $\qed$}}
\newcommand{\reviewer}[2][]{{\color{YodaColor} Reviewer#1: #2 $\qed$}}

\newcommand{\warning}[1]{{\it\color{red} #1}}
\newcommand{\note}[1]{{\it\color{blue} #1}}
\newcommand{\nothing}[1]{}

\definecolor{AudioColor}{rgb}{0, 0, 0}
\newcommand{\audio}[2]{{\color{AudioColor} [#1] #2 $\qed$}}
\definecolor{VideoColor}{rgb}{0.44,0.66,0.38}
\newcommand{\video}[1]{{\color{VideoColor} Video: #1 $\qed$}}
\definecolor{DeadlineColor}{rgb}{0.9,0.4,0}
\newcommand{\deadline}[1]{{\bf\color{DeadlineColor} ETA: #1}}

\definecolor{OldColor}{rgb}{0.5,0.5,0.5}
\newcommand{\old}[1]{{\color{OldColor} #1}}
\definecolor{NewColor}{rgb}{0.9,0.4,0}

\definecolor{DeleteColor}{rgb}{0.1,0.6,1.0}
\newcommand{\delete}[1]{{\color{DeleteColor} #1}}
\definecolor{MoveColor}{rgb}{0.5,0.1,0.5}

\definecolor{figred}{rgb}{1,0,0}
\definecolor{figgreen}{rgb}{0,0.6,0}
\definecolor{figblue}{rgb}{0,0,1}
\definecolor{figpink}{rgb}{1,0.63,0.63}

\ifthenelse{\equal{\final}{1}}
{
\renewcommand{\ckliang}[1]{}
\renewcommand{\yichang}[1]{}
\renewcommand{\minghsuan}[1]{}
\renewcommand{\wslai}[1]{}
\renewcommand{\reviewer}[2][]{}

\renewcommand{\warning}[1]{}
\renewcommand{\note}[1]{}
\renewcommand{\old}[1]{}
\renewcommand{\audio}[2][]{}
\renewcommand{\video}[1]{}
\renewcommand{\deadline}[1]{}
}
{}

\newcommand{\pseudocode}{Pseudocode}
\floatstyle{plain}
\newfloat{algorithm}{tbhp}{lop}
\floatname{algorithm}{\pseudocode}

\ifthenelse{\isundefined{\diff}}
{

\renewcommand{\delete}[1]{}

}
{
\ifthenelse{\equal{\diff}{0}}
{

\renewcommand{\delete}[1]{}

}
{}
}

\ifdefined\google

\else

\fi

\newcommand{\figref}[1]{Figure~\ref{fig:#1}}
\newcommand{\eqnref}[1]{\eqref{eq:#1}}

\newcommand{\blue}[1]{\textcolor{blue}{#1}}
\newcommand{\red}[1]{\textcolor{red}{#1}}
\newcommand{\green}[1]{\textcolor{DarkGreen}{#1}}
\newcommand{\white}[1]{\textcolor{white}{#1}}
\newcommand{\black}[1]{\textcolor{black}{#1}}

\newcommand{\Paragraph}[1]{\vspace{2mm}\noindent\textbf{#1}} 
\renewcommand{\paragraph}[1]{\vspace{2mm} \noindent\textbf{#1}}

\newcommand{\colvector}[1]{\textbf{#1}}
\newcommand{\normtwo}[1]{\|{#1}\|^2_2}

\def\imagewidth{W}

\def\mesh{M}
\def\vertex{\colvector{v}}
\def\stereo{u}
\def\perspective{p}

\def\stereovertex{\colvector{\stereo}}
\def\perspectivevertex{\colvector{\perspective}}

\def\energy{E}
\def\faceterm{f}
\def\boundaryterm{b}
\def\spatialregterm{s}
\def\gridedgeterm{e}
\def\temporalterm{t}
\def\coherentterm{c}
\def\lineterm{l}
\def\mask{P}
\def\spatial{sp}

\def\facebox{\textbf{B}}
\def\facescalemat{\textbf{S}}
\def\facescale{a}
\def\facetranslation{\textbf{t}}
\def\facetargetscale{s_f}
\def\faceweight{w_{k}}
\def\facescaleweight{w_{s}}

\newcommand{\neighbor}[1]{A(#1)}

\def\leftborder{\partial_{\text{left}}}
\def\rightborder{\partial_{\text{right}}}
\def\bottomborder{\partial_{\text{bottom}}}
\def\topborder{\partial_{\text{top}}}

\def\unitvector{\hat{\colvector{e}}}

\def\line{\colvector{d}}
\def\quadd{\colvector{q}}
\def\pa{\colvector{a}}
\def\pb{\colvector{b}}
\def\coeff{\colvector{w}}

\ifCLASSOPTIONcompsoc
  \usepackage[nocompress]{cite}
\else
  \usepackage{cite}
\fi

\begin{document}
\normalem
\title{Correcting Face Distortion in Wide-Angle Videos}
\ifthenelse{\equal{\underreview}{1}}
{}
{
\author{Wei-Sheng Lai, YiChang Shih, Chia-Kai Liang, Ming-Hsuan Yang\\
Google\\
{\tt\small \{wslai, yichang, ckliang, minghsuan\}@google.com}
}
}

\author{Wei-Sheng Lai, YiChang Shih, Chia-Kai Liang, Ming-Hsuan Yang
\IEEEcompsocitemizethanks{
	\IEEEcompsocthanksitem 
	All authors are with Google LLC,  Mountain View, CA 94043. Email: $\{$wslai$|$yichang$|$ckliang$|$minghsuan$\}$@google.com
	%
	%
}
}

\maketitle

\begin{abstract}
\label{sec:abstract}
Video blogs and selfies are popular social media formats, which are often captured by wide-angle cameras to show human subjects and expanded background.
Unfortunately, due to perspective projection, faces near corners and edges exhibit apparent distortions that stretch and squish the facial features, resulting in poor video quality.
In this work, we present a video warping algorithm to correct these distortions.
Our key idea is to apply stereographic projection locally on the facial regions. 
We formulate a mesh warp problem using spatial-temporal energy minimization and minimize background deformation using a line-preservation term to maintain the straight edges in the background.
To address temporal coherency, we constrain the temporal smoothness on the warping meshes and facial trajectories through the latent variables.
For performance evaluation, we develop a wide-angle video dataset with a wide range of focal lengths.
The user study shows that 83.9\% of users prefer our algorithm over other alternatives based on perspective projection.
The video results can be found at \url{ https://www.wslai.net/publications/video_face_correction/}.
\end{abstract}

\begin{IEEEkeywords}
Wide-angle videos, video warping, face distortion.
\end{IEEEkeywords}
\IEEEpeerreviewmaketitle


\section{Introduction}
\label{sec:intro}
\begin{figure*}
    \captionsetup{type=figure}
	\footnotesize
	\renewcommand{\tabcolsep}{1pt} 
	\renewcommand{\arraystretch}{0.8} 
    \begin{tabular}{ccc}
        \begin{overpic}[width=0.32\textwidth,tics=10]{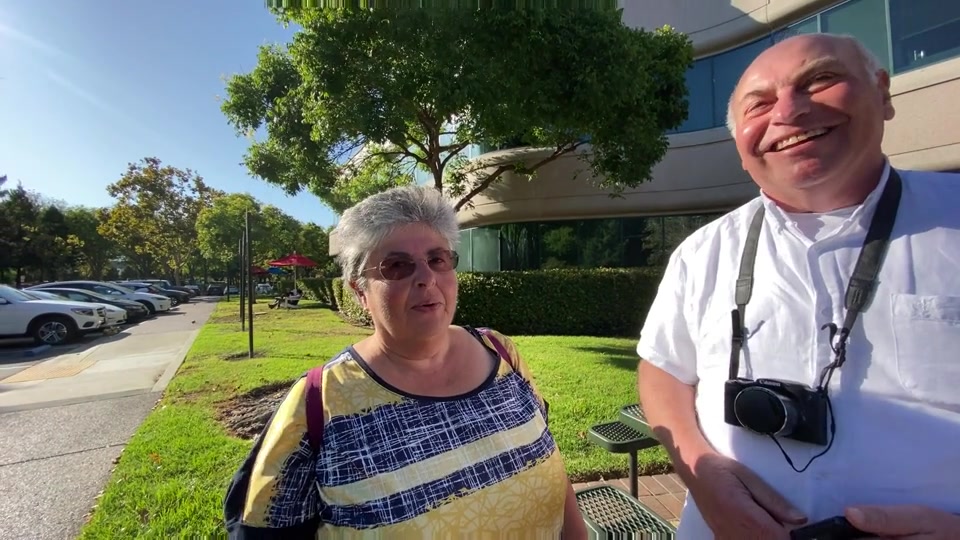}
            \put (3,50) {\white{$105\degree$ FOV}}
            \put (3,45) {\white{Frame $n$}}
        \end{overpic} &
        \begin{overpic}[width=0.32\textwidth,tics=10]{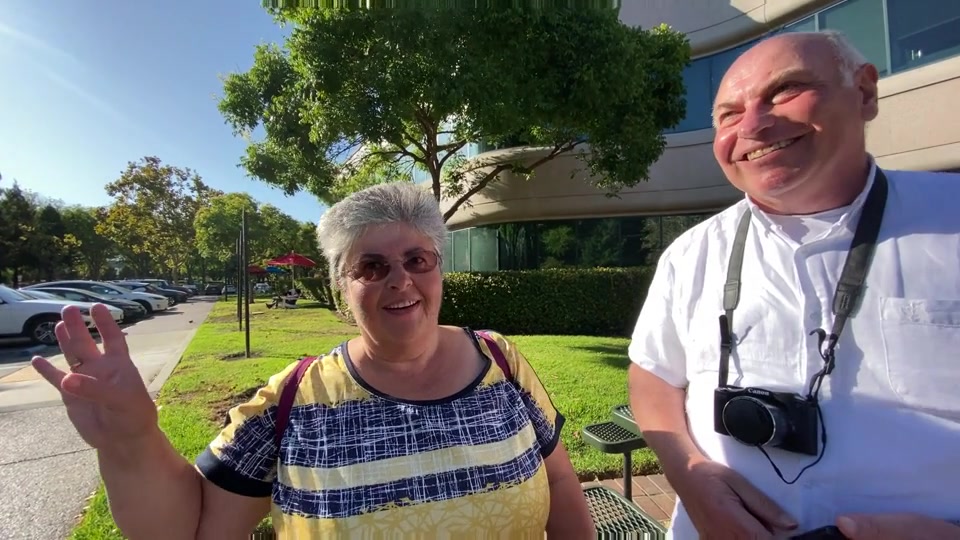}
            \put (3,50) {\white{$105\degree$ FOV}}
            \put (3,45) {\white{Frame $n + 50$}}
        \end{overpic} &
        \begin{overpic}[width=0.32\textwidth,tics=10]{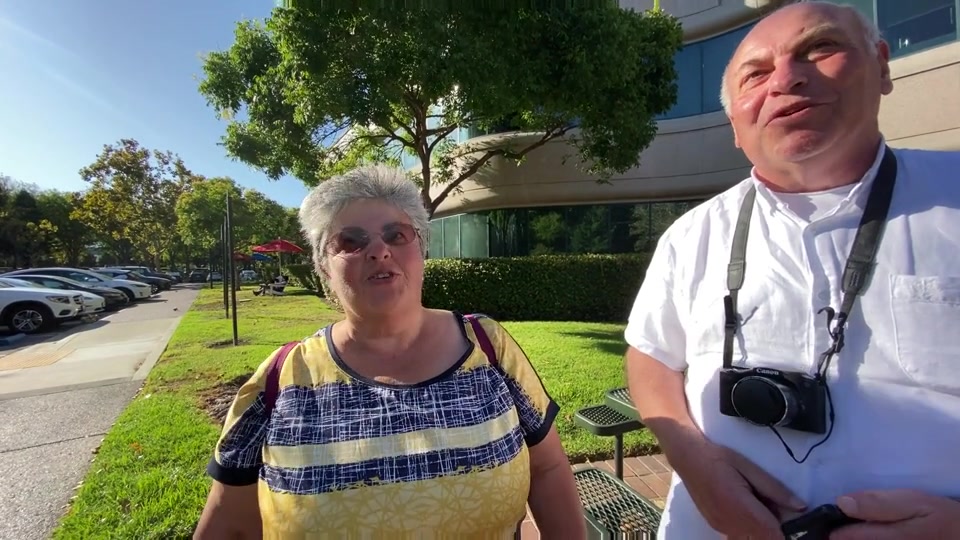}
            \put (3,50) {\white{$105\degree$ FOV}}
            \put (3,45) {\white{Frame $n + 100$}}
        \end{overpic}
        \\
        \includegraphics[width=0.32\textwidth]{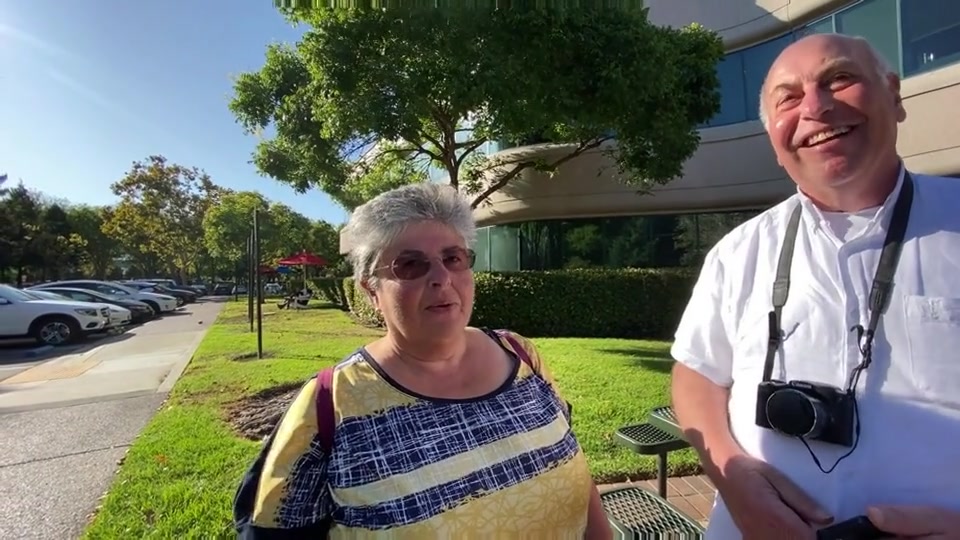} & 
        \includegraphics[width=0.32\textwidth]{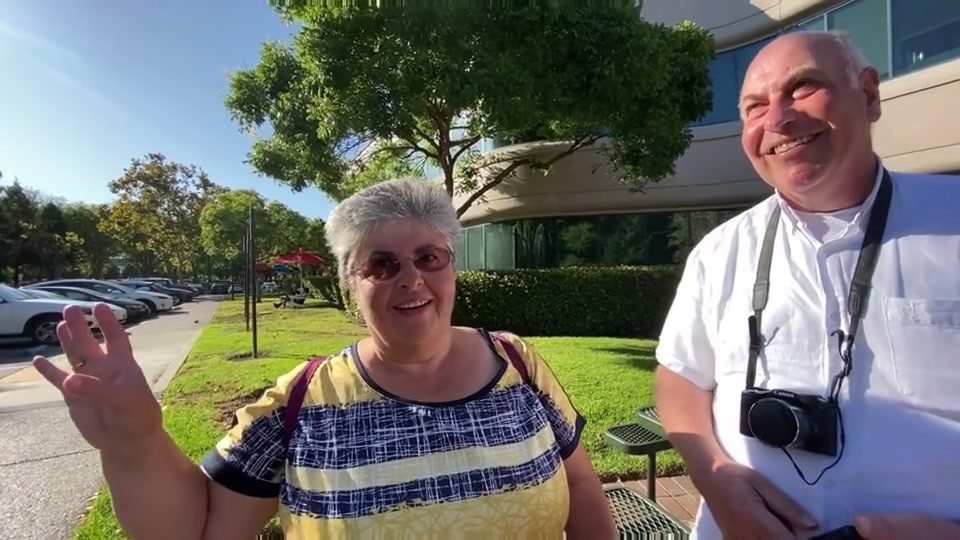} & 
        \includegraphics[width=0.32\textwidth]{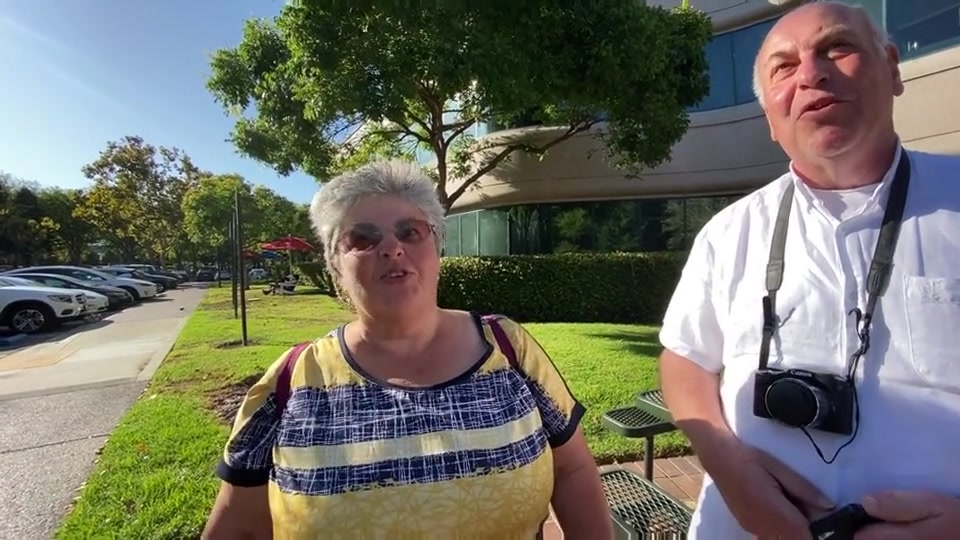} 
    \end{tabular}
	\captionof{figure}{
        \textbf{Top:} 
        Input video frames captured by the Apple iPhone 11 wide-angle camera (105\degree~diagonal field-of-view, 30 FPS).
        Wide-angle camera's perspective projection introduces apparent distortion of the subjects near corners. The facial features are stretched and squished. 
        \textbf{Bottom:} 
        Our approach corrects facial distortion, preserves the scene geometry, and maintains temporal consistency.
	}
	\label{fig:teaser}
\end{figure*}

\IEEEPARstart{M}{any} mobile videos are recorded by wide-angle lenses to include both the narrating subjects and background.
Recently, the camera field-of-view (FOV) on mobile phones ranges from $80\degree$ to $130\degree$.
Due to the camera's perspective projection, video playback exhibits visual distortion that stretches subjects near the image corners~\cite{Dhanraj:2005:WPL} (\figref{teaser} Top).
%
As such, the subjects look vastly different than in real life, and various computer vision tasks may not perform well if the algorithms are trained or tuned on the images acquired from cameras of narrower FOVs. 
%
%

Professional cinematography circumvents this problem by staging the subjects near the camera center or using longer focal lenses. 
The latter approach is expensive and cumbersome for video photographers using hand-held cameras, especially when the filming space is limited. 
This raises the need for post-processing to remove the apparent distortion caused by wide-angle lenses.
Existing approaches use stereographic projection~\cite{Zorin:1995:CGP} to reduce perspective distortion, but result in fisheye artifacts and lose scene realism.
Advanced photo processing algorithms exploit a local mesh warp adapted to facial regions, and generate a natural look on both subjects and backgrounds~\cite{Carroll:2009:OCP, Shih:2019:DFW}. 
Extending those algorithms for videos is challenging, as subtle changes on the mesh grid create severe temporal flickering and wobbling effects~\cite{Niu2010:WPF}.

In this paper, we address the wide-angle face distortion on videos using spatial-temporal mesh optimization.
%
%
Our goal is to generate a video with a natural look on human faces since faces draw significant attention from the viewers~\cite{Goferman:2011:CAS,Judd:2009:LTP,Pan:2016:SAD}.
Our algorithm generates a warping mesh that locally adapts to the stereographic projection on the facial region.
Temporally, we regularize the mesh and the facial similarity transforms across all the video frames. 
To preserve the scene geometry, we introduce a line-preservation term using a line tracking algorithm across multiple frames.
Finally, temporal coherence is achieved by propagating the facial information and warping mesh from the future to the current frame in a non-causal fashion through spatial-temporal mesh optimization.

Our method is validated on a database consisting of $126$ hand-held videos, whose diagonal FOVs range from $97\degree$ to $105\degree$, and evaluated against existing approaches via a user study.
%
The perspective distortion addressed in this work is different from lens distortion, which is often corrected through camera calibration and post-processing software, e.g., Adobe Premier.
We assume the input videos to our algorithm follow the perspective projection and correct the lens distortion during a pre-processing step.

Our contributions in this work are:
\begin{compactitem}
\item A novel video algorithm to recover facial distortion deformed by perspective projection. 
In contrast to other alternatives, our method is fully automatic without user intervention. 
\item A benchmark dataset of $126$ wide-angle videos collected from Google Pixel 3, GoPro, and iPhone 11 cameras for thorough performance evaluation. 
\end{compactitem}

Our work is built on top of Shih et al.~\cite{Shih:2019:DFW} but with the following main differences:
\begin{enumerate}
\item We introduce temporal smoothness and coherent embedding terms to preserve the temporal consistency of optimized meshes.
\item We adopt a line-preservation term to \emph{explicitly} preserve the straight lines in the background.
\item We extend the single-image mesh optimization of Shih et al.~\cite{Shih:2019:DFW} to handle videos by introducing face tracking, video subject mask segmentation, line tracking, and full-volume optimization.
\end{enumerate}

\section{Related Work}
\label{sec:related_work}
\subsection{Video Warping} 
Our method builds on mesh-based video warping algorithms used by numerous video editing applications, such as retargeting~\cite{Krahenbuhl:2009:ASF, Lin:2013:CAV, Wang:2009:MAT, Wang:2011:SAC, Wang:2010:MBV,  Wolf:2007:NHC}, stabilization~\cite{Grundmann:2011:ADV, Liu:2009:CPW, Liu:2013:BCP}, rolling shutter removal~\cite{Grundmann:2012:CFR, Karpenko:2011:DVS}, stitching~\cite{Jiang:2015:VSW, Nie:2017:DVS}, stereoscopic video retargeting~\cite{Li:2018:DAS,Liu:2015:ARM}, and lens undistortion~\cite{Wei:2012:FVC}.
Our work is specific to face-centric video warping~\cite{Shi:2019:SRT, Yu:2018:SVS} and takes the semantic information from face tracking and  segmentation~\cite{Wadhwa:2018:SDW}. 
Instead of translating or cropping an input video, our method removes the apparent distortion locally on faces while minimizing the impact on the rest of the scene.
The proposed method extends the photo-based perspective distortion correction~\cite{Shih:2019:DFW} to videos.
Spatially, we use stereographic projection~\cite{Zorin:1995:CGP} and grid edge-bending terms~\cite{Chang:2011:CAD,Wang:2008:OSS}. 
We address temporal consistency challenges~\cite{Niu2010:WPF} by adding a temporal smoothness term to and enforcing coherent embedding on the face similarity transform. 
We then adopt a full-volume optimization on the entire video.
Unlike user-assisted video warping~\cite{Krahenbuhl:2009:ASF, Wei:2012:FVC} or control-point-based facial animation~\cite{Liao:2014:SAV}, our method is designed for automatic processing.

Some recent approaches~\cite{Valente:2015:PDM,fried2016perspective,nagano2019deep,zhao2019learning} aim to remove perspective distortion in near-range portraits, where the nose and eyes tend to look larger and the ears vanish altogether.
Such distortions occur when the subject is close to the camera.
On the other hand, we address perspective distortion due to the wide-angle camera, which appears when a subject is far from the camera center.

\subsection{Distortion Correction}
Our work is different from existing methods that address optical distortion~\cite{Wei:2012:FVC, Hugemann:2010:CLD}.
In this paper, we assume the camera has been calibrated, and the optical distortion is corrected using parametric warping, or via commercial software such as Adobe Premiere Pro. 
We address perceptual distortion due to the perspective image projection, which becomes more prominent when the field-of-view is more than 72\degree. 
While existing approaches focus on rectifying perspective distortion on planar objects such as documents or photos~\cite{Li:2019:DRI, Markovitz:2020:CYR}, our work focuses on human faces. 
For static images, perceived distortion is often addressed through the combination of subject segmentation and either planar~\cite{Tehrani:2016:CPP} or stereographic projection~\cite{Shih:2019:DFW}.
However, such a method does not maintain temporal coherence when extending to video inputs.
Measuring the amount of perceived perspective distortion is essential to distortion correction and remains an active research problem~\cite{Bousaid:2020:PDM}. 
We assume the perceived distortion is correlated to the loss of local conformality when mapping objects from the 3D world to 2D image space~\cite{Carroll:2009:OCP}, and employ a stereographic projection to preserve the local conformality. 
While some recent approaches use end-to-end deep neural networks for image rectification~\cite{Yin:2018:FAM, Del:2020:BFO}, 
our method employs a deep neural network for subject segmentation and solves an optimization problem to generate temporally consistent warps.

\subsection{Straight Line Preservation} 
Preserving both object shapes and straight edges under mesh-based image warping is a challenging task, especially when the edges are near the face region.
Image-based methods typically rely on user annotation~\cite{Carroll:2010:IWF,Carroll:2009:OCP} to preserve salient edges. 
However, labeling straight edges on every frame is tedious and impractical for video processing. 
Recent approaches automate the process using a line detector~\cite{Von:2010:LSD} and a line preservation term~\cite{Chang:2012:ALA, He:2013:RPI, Li:2015:AGP} during the mesh optimization step.
A straightforward extension of these terms for videos is likely to cause temporal stability issues since missing lines will result in unexpected changes to the warping mesh. 
In this work, we use the pyramidal Lucas-Kanade method~\cite{bouguet2001pyramidal} to track the end-points of line segments.
By jointly optimizing the entire video sequence with a temporal smoothness term, we obtain temporally stable results with corrected faces.

%


\begin{figure*}[t!]
    \centering
    \includegraphics[width=1.0\textwidth]{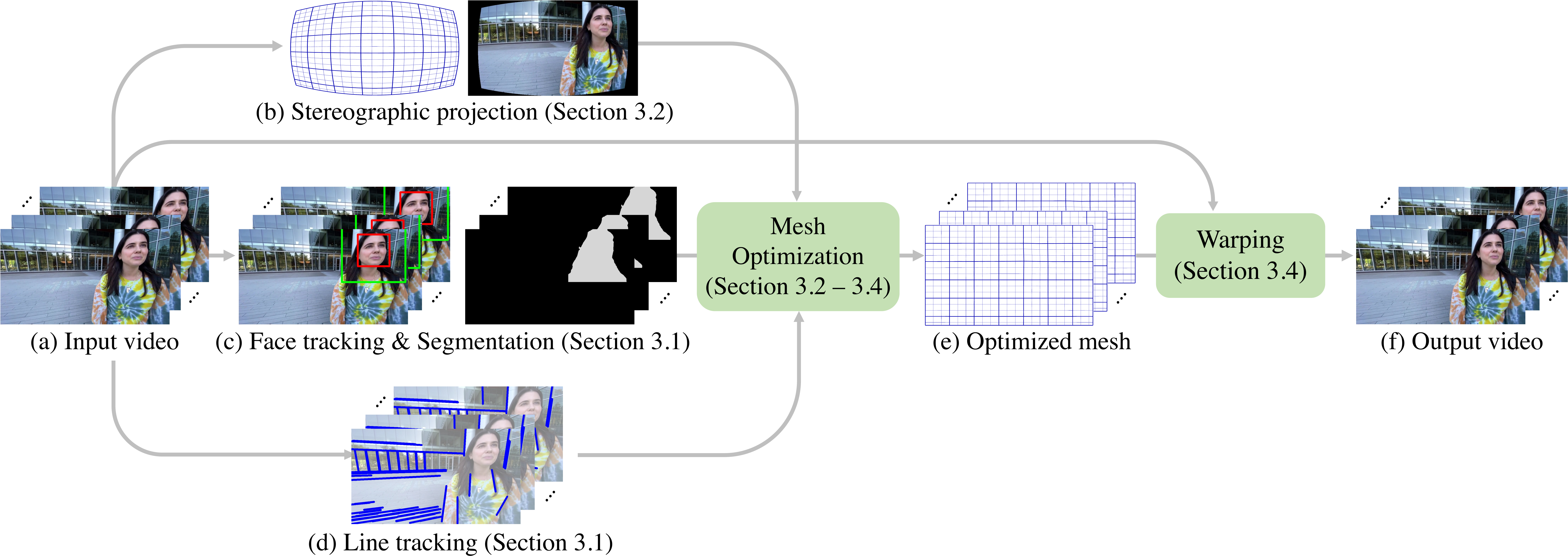}
    \caption{
        \textbf{Overview of the proposed algorithm}.
        Given (a) a wide-angle input video, we first (b) compute warping mesh for stereographic projection, (c) identify face regions, and (d) track background lines across the entire video.
        We then (e) jointly optimize our spatial and temporal energy functions to find a set of optimal warping meshes that locally apply the stereographic projection to the face regions while keeping the background as close as the input projection (i.e., perspective projection).
        Finally, we (f) warp the input video based on the warping mesh to generate the output video.
        %
        %
        %
    }
    \label{fig:overview}
\end{figure*}

\section{Algorithm}
\label{sec:algorithm}

In this section, we first provide an overview of the proposed algorithm and pipeline.
We then introduce our spatial and temporal energy terms for mesh optimization.
Finally, we present the details of our mesh optimization for video warping.

\subsection{Overview}

\figref{overview} shows an overview of the proposed algorithm.
Given a wide-angle video, our goal is to generate an image sequence where the human faces look natural without distorting the background scene.
To this end, we first track subject faces and line segments across the entire video.
We then jointly optimize the warping mesh across the spatio-temporal domain and return a set of warping mesh $\{\mesh^{(n)}\}$, where $n$ indexes the frame of the video. 
Each mesh consists of a set of vertices $\{\vertex_i\}$ on a 2D grid sharing the same dimension across the video. 
The length of $\{\mesh^{(n)}\}$ is equivalent to the total frame number $N$ in the input video. 
The final output video is generated by applying frame-based warping using $\{\mesh^{(n)}\}$ on the input video. 

\paragraph{Face tracking and segmentation.}
To track the faces in a video, we use a single-shot detector~\cite{liu2016ssd}\footnote{We use the pre-trained model (COCO model SSD512) from \url{https://github.com/weiliu89/caffe/tree/ssd}, but our method is compatible with any off-the-shelf face detector.} trained on face images, and use optical flow to predict where a set of faces may appear in a frame given the results from the previous frame.
We use a subject segmentation network with a U-Net architecture~\cite{Ronneberger:2015:UNC, Tkachenka:2019:RTH} to identify facial regions.
The segmentation network processes a video frame-by-frame.
In addition, we use the subject segmentation result from the previous frame to generate the mask for the current frame and maintain temporal consistency.
To cover the correction area on hair and chin, we expand the face bounding box size by a factor of two.
The face regions are the intersection between the expanded face bounding boxes and subject masks.

\paragraph{Line tracking.}
We use the line segment detection method~\cite{Von:2010:LSD} to identify the straight edges in the background.
Then, we track the two end-points of each line segment via the pyramidal Lucas-Kanade method~\cite{bouguet2001pyramidal}.
Specifically, we compute the forward and backward optical flows of the end-points and check the re-projection consistency.
We continue tracking the line segment if the re-projection errors of both end-points are within $2$ pixels.
With these steps, we observe that the lines in the background roughly maintain the same orientation across the video frames, and discard the detected lines if the orientation variation is larger than $1\degree~$ between the adjacent frames, indicating unreliable tracking results.
%
Our line tracking excludes the facial regions, which will be corrected by the proposed warping method.

\subsection{Spatial Energy Function}
\label{sec:spatia-energy}

Our spatial term $\energy_{\spatial}$ builds on the image-based energy terms described in Shih~\etal~\cite{Shih:2019:DFW}, which consists of a face mask $P$ and stereographic projection mesh $\mesh_\stereo$ as:
\begin{align} \label{eq:total_cost_photo}
    \energy_{\spatial} = 
    \lambda_\faceterm \energy_\faceterm(\mask, \mesh_\stereo) +
    \lambda_\spatialregterm E_\spatialregterm +
    \lambda_\gridedgeterm E_\gridedgeterm +
    \lambda_\boundaryterm E_\boundaryterm,
\end{align}
where $\energy_\faceterm$, $\energy_\spatialregterm$, $\energy_\gridedgeterm$, and $\energy_\boundaryterm$ are the face, spatial smoothness, grid edge-bending, and boundary terms, respectively; $\lambda_\faceterm$, $\lambda_\spatialregterm$, $\lambda_\gridedgeterm$, and $\lambda_\boundaryterm$ are the weights for the corresponding energy terms. 
The spatial energy term $\energy_{\spatial}$ is computed for every frame as the subject mask $\mask$ varies across the video. 
For readability, we omit the temporal index $n$ in the spatial terms. 

\paragraph{Stereographic projection.}
We compute the stereographic projection mesh using a radial mapping:
\begin{align}
    r_u = r_0 \tan\left( 0.5 \tan^{-1}\left(\frac{r_p}{f}\right)\right), 
    \label{eq:stereo}
\end{align}
where $f$ is the camera focal length described with the same unit as $r_p$, $r_u$ and $r_p$ are the radial distances to the optical center under stereographic projection and perspective projection, respectively.
The scaling factor $r_0$ is chosen such that $r_u = r_p$ at the image boundary:
\begin{align}
    r_0 = \frac{d}{2 \tan\left(0.5 \tan^{-1}\left(\frac{d}{2f}\right) \right)},
\end{align}
where $d = \min(W, H)$ is the minimum of the image width and height.
Given a perspective projection mesh vertex $\{x_p, y_p\}$ in Cartesian coordinates, we first compute $r_p$ by:
\begin{align}
    r_p = \sqrt{(x_p - W/2)^2 + (y_p - H/2)^2}.
\end{align}
We then calculate $r_u$ using \eqref{eq:stereo} and convert back to Cartesian coordinates via:
\begin{align}
    x_u = r_u / r_p \cdot x_p + W/2, \\
    y_u = r_u / r_p \cdot y_p + H/2.
\end{align}

\paragraph{Face term.}
We associate each subject in a video frame with an energy function $\energy_k$:
\begin{align} \label{eq:face_term}
    \energy_k = 
    \faceweight \sum_{i \in {\facebox_k}} \normtwo{ \vertex_i - \left( \facescalemat_k \stereovertex_i + \facetranslation_k \right) }
    + \facescaleweight ( \facescale_k - \facetargetscale )^2 \,,
\end{align}
where $k$ indexes the detected subject face, $\facebox_k$ denotes the set of vertices on the $k$-th face, $\{\stereovertex_i\}$ are vertices of the stereographic mesh $\mesh_\stereo$, $\facescalemat_k$ and $\facetranslation_k$ are the parameters of the similarity transform.
The total face term $\energy_\faceterm = \sum_{k=0}^{K-1} \energy_k$, where $K$ denotes the total number of faces, is the sum of the energy associated with all the subjects in a frame. 
The first term in~\eqnref{face_term} enforces the vertices on face regions to be similar to those vertices on the stereographic mesh, while allowing the faces to be relaxed through the similarity transform as:
\begin{align}
    \facescalemat_k = 
        \begin{bmatrix}
            a_k & b_k  \\
            -b_k & a_k
        \end{bmatrix}\,,
    \facetranslation_k = 
        \begin{bmatrix}
            t_{x,k}  \\
            t_{y,k}
        \end{bmatrix}\,.
    \label{eq:face_affine}
\end{align}
The second term in~\eqnref{face_term} regularizes the scale of the face, where the target face scale $\facetargetscale$ is set to $1$ and the weight $\facescaleweight$ is set to $1$.
As the image corners have stronger perspective distortions, we set the per-face weight $\faceweight = \tanh{(2 r_k / r_{\text{max}})}$, where $r_k$ and $r_{\text{max}}$ are the distances from the image center to the face and image corners, respectively.

\paragraph{Spatial smoothness term.}
We impose spatial smoothness using the following energy function:
\begin{align}
    E_\spatialregterm = \sum_i \sum_{j \in \neighbor{i}} \normtwo{\vertex_i - \vertex_j}\,,
\end{align}
where $\neighbor{i}$ denotes the 4-way adjacent vertices of $\vertex_i$.

\paragraph{Grid edge-bending term.}
We use the following energy function to minimize the distortion on grid edges:
\begin{align} \label{eq:grid_edge_bending_term}
    E_\gridedgeterm 
    &= \sum_i \sum_{j \in \neighbor{i}} \normtwo{ (\vertex_i - \vertex_j) \times \unitvector_{ij} }, 
\end{align}
where $\times$ is the cross-product and $\unitvector_{ij}$ represents the unit vector along the direction of the source uniform grid $\perspectivevertex_i - \perspectivevertex_j$.
The energy function in~\eqnref{grid_edge_bending_term} penalizes the component perpendicular to the grid edge $\colvector{e}_{ij}$ to reduce shearing deformation on the grid.
We note that this term is called the ``line-bending term'' in~\cite{Shih:2019:DFW}.
To avoid confusion with our line-preservation term, we refer to it as the ``grid edge-bending term'' in this work.

\paragraph{Boundary term.}
We apply a boundary cost term to the vertices on the mesh boundary to avoid the trivial null solution and keep the resolution of the video frame unchanged as:
\begin{align}\label{eq:boundary_cost}
    E_b &= \sum_{i \in \leftborder} \normtwo{\vertex_{i,x}}
         + \sum_{i \in \rightborder} \normtwo{\vertex_{i,x} - W} \nonumber \\
        &+ \sum_{i \in \topborder} \normtwo{\vertex_{i,y}}
         + \sum_{i \in \bottomborder} \normtwo{\vertex_{i,y} - H }\,,
\end{align}
where $\partial_{\ast}$ denotes the input mesh boundary, and $W$ and $H$ are the image width and height, respectively. 
%
Note that in Shih et. al.~\cite{Shih:2019:DFW}, the asymmetric boundary condition includes a non-linear and discontinuous step function that restricts the range of $\vertex_{i,x}$.
As stated in their paper, the constrained optimization requires the trust-region optimizer to have a good guess on the initial trust regions. 
In addition, the non-linear constraints do not guarantee a globally optimal solution. 
For a 30 fps video, it is difficult to make a consistent guess on every frame.
In contrast, we address the numerical stability with unconstrained boundary conditions. The quadratic formulation used in our boundary term guarantees the globally optimal solution using commonly available numerical optimization software.

\subsection{Temporal Consistency}
\label{sec:objective-functions}

The spatial energy $\energy_{\spatial}$ in~\eqnref{total_cost_photo} is not sufficient to deal with temporal flickering caused by drastic movements across the field of view or panning camera motion.
When faces are missing or falsely detected by the detector, the mesh changes between stereographic and perspective projections on the facial areas, resulting in an unpleasant viewing experience.
%
%
The flickering mesh is particularly noticeable when the scene contains rich geometric cues.

\paragraph{Temporal smoothness term.}
We enforce smoothness between each vertex and its temporal neighbors with the following energy function:
\begin{align} \label{eq:temporal_term}
    \energy_\temporalterm = \sum_{n=1}^{N-1} \sum_{i} \normtwo{ \vertex_i^{(n)} - \vertex_i^{(n-1)} }\,,
\end{align}
where $N$ is the total number of frames in the input video.
In this work, $\energy_\temporalterm$ connects every vertex in the spatial-temporal mesh volume and propagates the critical mesh deformation information from one frame to the entire video. 
This results in temporally stable warping meshes.
We show the effectiveness of the temporal smoothness term in the supplementary video.

\paragraph{Coherent embedding on face similarity transform.}
The face term $\energy_\faceterm$ in~\eqnref{total_cost_photo} enforces the vertices in face regions to be similar to those vertices in the stereographic mesh while allowing faces to be matched through the similarity transform.
For video inputs, the similarity transform parameters of the same face may change dramatically between consecutive frames.
Therefore, to improve temporal stability, we introduce a coherent embedding term to enforce the smoothness on the similarity transform embedding:
\begin{align} \label{eq:coherence_embedding_term}
    \energy_\coherentterm = \sum_{n=1}^{N-1} \sum_{k} \normtwo{ \facescalemat_k^{(n)} - \facescalemat_k^{(n - 1)}} + \normtwo{\facetranslation_k^{(n)}  - \facetranslation_k^{(n - 1)}} \,.
\end{align}
We apply the energy term in~\eqnref{coherence_embedding_term} when the same subject appears in the two consecutive frames according to the face tracking.

\paragraph{Line-preservation term.}
When faces are moving, straight lines in the background and near facial regions deform the shape inconsistently in the temporal domain.
The spatial and temporal energy functions, $\energy_{\spatial}$ and $\energy_\temporalterm$ in \eqnref{total_cost_photo} and \eqnref{temporal_term}, respectively, guide the proposed method to generate meshes in the opposite directions: while $\energy_{\spatial}$ adapts to the current face information, $\energy_{\temporalterm}$ retains the mesh identical across the temporal domain. 
The conflicting goals distort straight lines and result in significant flickering artifacts. 
To address this issue, we introduce a line-preservation energy term to maintain the geometry structure in the background.

\begin{figure}
    \centering
    \includegraphics[width=0.8\linewidth]{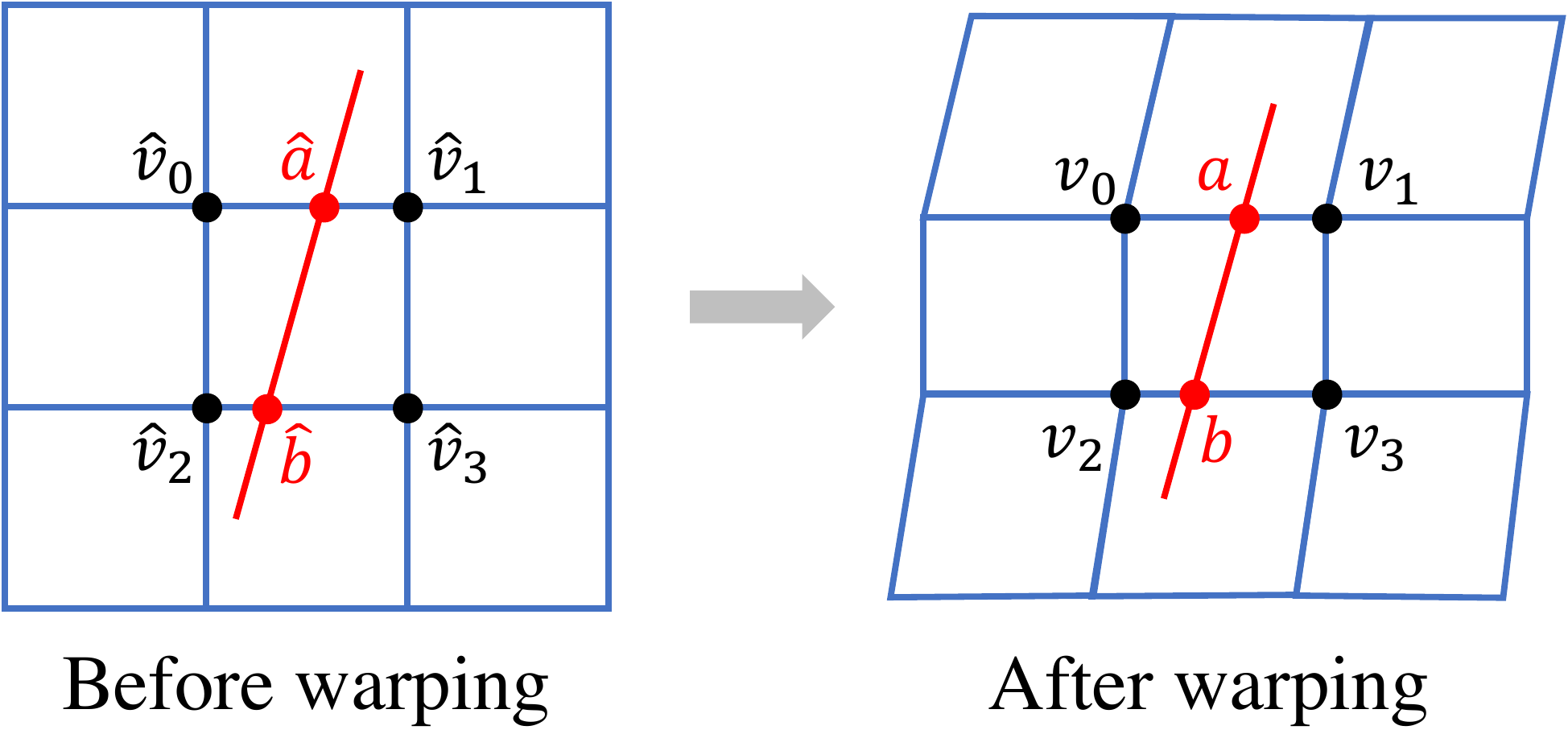}
    \caption{
        \textbf{Line segment representation.}
        Our line-preservation cost preserves the line orientation after the warping, i.e., the direction of $\line = \pb - \pa$ remains the same as $\hat{\line} = \hat{\pb} - \hat{\pa}$.
    }
    \label{fig:line_segement}
\end{figure}

Consider a quad that intersects a line segment $l$ as illustrated in~\figref{line_segement}.
We represent the quad by the vertices of the four corners as a 2-by-4 matrix $\quadd=[\vertex_{0},\vertex_{1},\vertex_{2}, \vertex_{3}]$ and denote the two intersection points of line segment $l$ on quad $\quadd$ as $\pa$ and $\pb$. 
The point $\pa$, and similarly for $\pb$, can be represented through the Barycentric coordinate system of the quad: $\pa=\quadd\coeff_a$, where $\coeff_{a}$ is a 4-by-1 vector whose $\ell_1$ norm equals to $1$. 
We denote line segment $l$ on quad $q$ as $\line_{l,q}=\pb-\pa=\quadd_{q}\coeff_{l,q}$, where $\coeff_{l,q}=\coeff_b-\coeff_a$. 
The coefficient $\coeff_{l,q}$ remains unchanged after the mesh deformation. 
This allows us to regularize deformation of quad $\quadd_{q}$ using the line-preservation energy function:
\begin{align} 
\label{eq:single_line_term}
   \energy_\lineterm = \sum_{l=1}^{l=L}\sum_{q\in l} \normtwo{ \line_{l,q} - s_{l,q}\hat{\line}_{l,q}}\,,
\end{align}
where $s_{l,q}$ is a scaling factor as a latent variable, $\hat{\line}$ denotes the line segment before warping as constraints, and $\line$ represents the line segment after the warping as optimization variables. 
%
For every frame, we compute $\energy_\lineterm$ across all detected lines $l \in L$ in the frame.  
The scaling factor $s_{l,q}$ preserves the background shape by penalizing the quad deformation orthogonal to the line direction.
It can be shown that $\energy_\lineterm$ is a quadratic function to $\line$ by substituting $s_{l,d}$ in~\eqnref{single_line_term} with the optimal values $ (\hat{\line}_{l,q}^{\top} \hat{\line}_{l,q} )^{-1} \hat{\line}_{l,q}^{\top} \line_{l,q}$~\cite{He:2013:RPI}. 
%

Our line-preservation term is motivated by prior approaches for image retargeting~\cite{He:2013:RPI, Chang:2012:ALA}. 
To handle video inputs, we constrain the rotation of the line segments to be identical before and after warping. 
Compared to image resizing~\cite{Chang:2012:ALA} and panorama rectangling~\cite{He:2013:RPI}, the effect of our warp is more local on the facial area only.
The additional freedom on rotation causes extra flickering in the videos.
Furthermore, the line orientation is non-linear to the rotation angle and makes the optimization process unstable.

\subsection{Mesh Optimization}
\label{sec:mesh-optimization}
The overall spatial-temporal energy function combines the spatial term $\energy_{\spatial}$, temporal smoothness term $\energy_{\temporalterm}$, coherent embedding term $\energy_{\coherentterm}$, and line preservation term $\energy_{\lineterm}$ in~\eqnref{total_cost_photo},~\eqnref{temporal_term},~\eqnref{coherence_embedding_term}, and ~\eqnref{single_line_term}, respectively, over the entire video frames indexed by $n$:
\begin{align} \label{eq:total_cost_video}
    \sum_{n=0}^{N-1}\left(\energy_{\spatial}^{(n)}  +\lambda_\lineterm\energy_{\lineterm}^{(n)}\right) +
    \lambda_\coherentterm\energy_\coherentterm +
    \lambda_\temporalterm\energy_\temporalterm .
\end{align}
We set $\lambda_\lineterm=64$, $\lambda_\coherentterm=4$, and $\lambda_\temporalterm=16$ in~\eqnref{total_cost_video}, and $\lambda_\faceterm=4$, $\lambda_\spatialregterm=1$, $\lambda_\gridedgeterm=2$, $\lambda_\boundaryterm=4$ in~\eqnref{total_cost_photo} for all the experiments.
%

\begin{figure*}[t!]
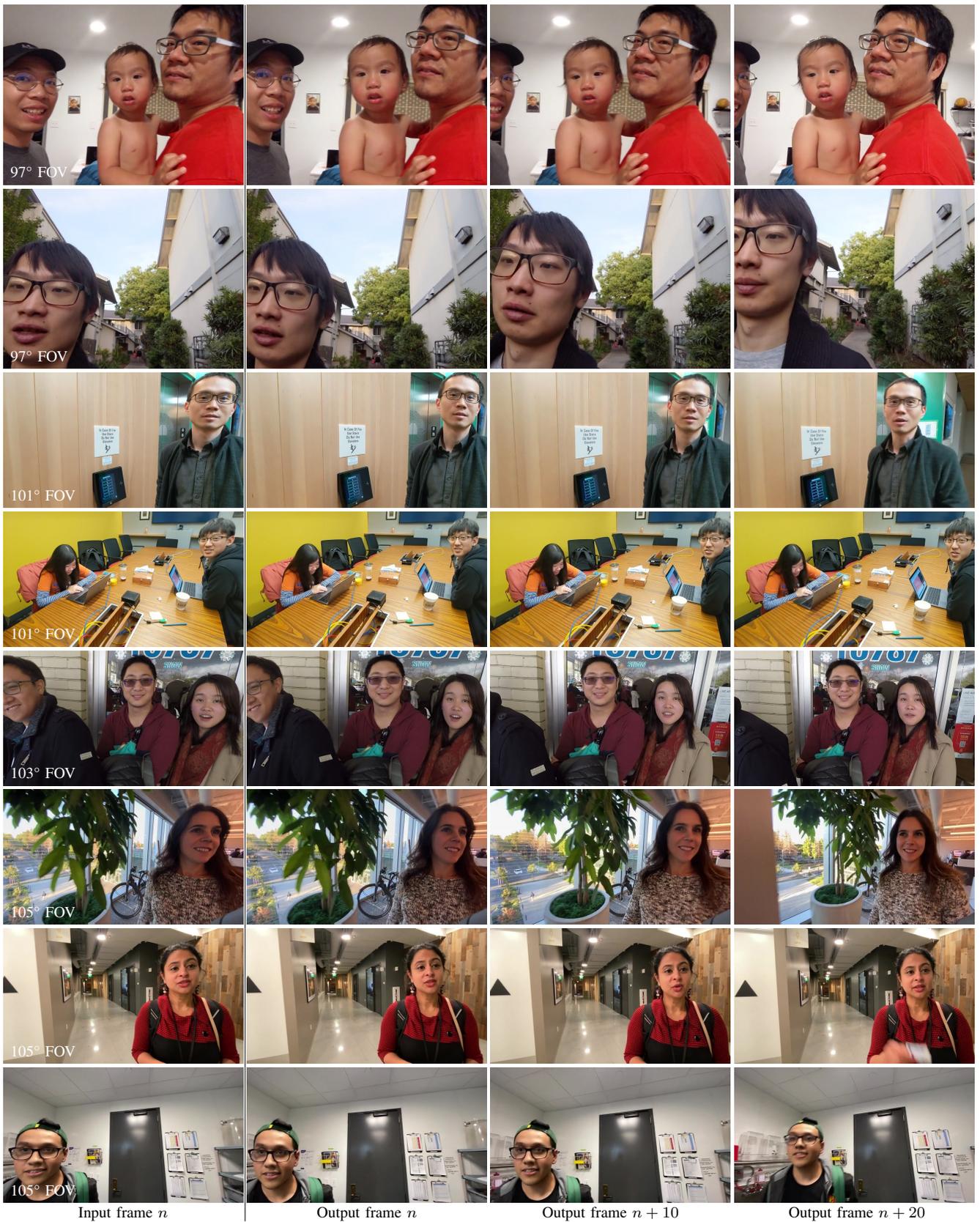

    \centering
    \footnotesize
    \renewcommand{\tabcolsep}{1pt} 
	\renewcommand{\arraystretch}{0.8} 
	\newcommand{\listcases}[7]{
	    \begin{overpic}[width=#1\linewidth,tics=10]{figures/comparisons/#2/#3.jpg}
            \put (3,3) {\white{#4$\degree$ FOV}}
        \end{overpic} &
	    \includegraphics[width=#1\textwidth]{figures/comparisons/#2/#5.jpg} &
	    \includegraphics[width=#1\textwidth]{figures/comparisons/#2/#6.jpg} &
	    \includegraphics[width=#1\textwidth]{figures/comparisons/#2/#7.jpg}
	}
    \begin{tabular}{c|ccc}
        \listcases{0.24}{pixel3-front-osmo-group3-20190419-18}
        {000520_input}{97}{000520}{000530}{000540}
        \\
        \listcases{0.24}{pixel3-front-osmo-group1-20190419-17-1}
        {000370_input}{97}{000370}{000375}{000385}
        \\
        \listcases{0.24}{pixel3-rear-osmo-group1-20190508-2}
        {000250_input}{101}{000250}{000265}{000280}
        \\
        \listcases{0.24}{pixel3-rear-osmo-group2-20190203-1}
        {000300_input}{101}{000300}{000310}{000320}
        \\
        \listcases{0.24}{GOPR2439-group7}
        {000160_input}{103}{000160}{000170}{000180}
        \\
        \listcases{0.24}{10012019_174927}
        {000620_input}{105}{000620}{000630}{000643}
        \\
        \listcases{0.24}{20191010_015}
        {000060_input}{105}{000060}{000070}{000080}
        \\
        \listcases{0.24}{10032019_160352}
        {001910_input}{105}{001910}{001920}{001930}
        \\
        Input frame $n$ &
        Output frame $n$ &
        Output frame $n + 10$ &
        Output frame $n + 20$\\
    \end{tabular}
    \caption{
        \textbf{Results of our method.} For each wide-angle video, we show one input frame in the left-most column and three output frames in the right three columns. 
        The faces look more natural and exhibit less distortion than those in the input.
        We demonstrate the effect of temporal coherence by showing the results every 10 frames for 30 frames-per-second videos. 
    }
    \label{fig:results}
\end{figure*}

\paragraph{Implementations.}
We implement our algorithm in Python and solve the least-squares optimization problem described  in~\eqnref{total_cost_video} with the LSMR sparse solver~\cite{Fong:2011:LAI} in Scipy.
We sequentially determine the weights of the face term, spatial smoothness term, grid edge-bending term, boundary term, temporal smoothness term, coherence-embedding term, and the line-preservation term.
We analyze the effect of each term on a small validation set (with 10 videos) and empirically adjust the weights sampling from $1$ to $128$ by power of $2$, until visually pleasing results are achieved.
We set the mesh grid dimensions to be $33 \times 25$ for efficient optimization. 
While this is coarser than that in imaging retargeting applications, empirically we find it sufficient for our application.

Note that the focal length is the only camera parameter required by our algorithm, which is used in computing the stereographic projection meshes.
Incorrect focal length leads to distortion and unnatural face shapes in the output frames.
The focal length can be obtained from the image EXIF data or through camera calibration.

\begin{table}[t]
    \centering
        \caption{
        \textbf{Datasets.}
        Our datasets cover various resolutions and diagonal FOVs for quality evaluation and user study.}
    \label{tab:dataset}
    \footnotesize
    \begin{tabular}{c|ccc}
        \toprule
        Camera Model &  DFOV(\degree) & Resolution & $\#$Videos \\
        \midrule
        GoPro Hero 5 & 103 & $1920 \times 1080$ & 22 \\
        Pixel 3 front & 97 & $1024 \times 768$ & 44 \\
        Pixel 3 rear & 101 & $1920 \times 1080$ & 14 \\
        iPhone11 Pro Max & 105 & $1920 \times 1080$ & 46 \\
        \bottomrule
    \end{tabular}

\end{table}

\begin{figure*}
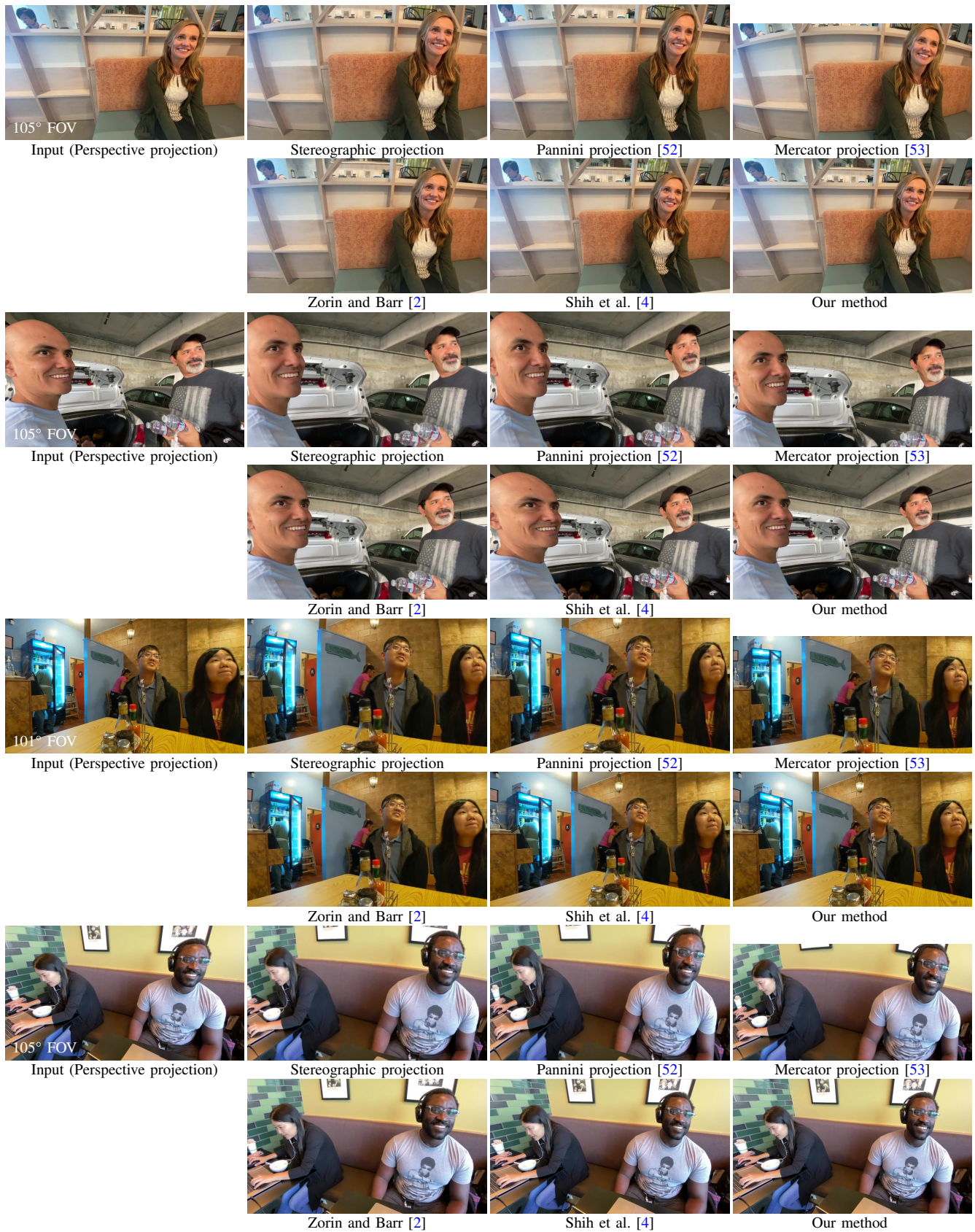

    \centering
    \footnotesize
    \renewcommand{\tabcolsep}{1pt} 
	\renewcommand{\arraystretch}{0.8} 
    \newcommand{\case}[3]{\includegraphics[width=#1\textwidth]{figures/comparisons/#2/#3.jpg}}
	\newcommand{\listmethods}[4]{
	    \begin{overpic}[width=#1\textwidth,tics=10]{figures/comparisons/#2/#3_input.jpg}
            \put (3,3) {\white{#4$\degree$ FOV}}
        \end{overpic} &
	    \case{#1}{#2}{#3_stereo_crop} & 
	    \case{#1}{#2}{#3_pannini_crop} & 
	    \case{#1}{#2}{#3_mercator_crop} \\
        Input (Perspective projection) & 
        Stereographic projection & 
        Pannini projection~\cite{Sharpless:2010:PAN} & 
        Mercator projection~\cite{mercator} \\
	    &
	    \case{#1}{#2}{#3_zorin_crop} & 
	    \case{#1}{#2}{#3_shih19} &
	    \case{#1}{#2}{#3_ours} \\
	    &
        Zorin and Barr~\cite{Zorin:1995:CGP} & 
        Shih et al.~\cite{Shih:2019:DFW} &
        Our method
	}
    \begin{tabular}{ccccc}
        \listmethods{0.24}{20191015_009}{001242}{105}
        \\
        \listmethods{0.24}{10032019_162237}{000460}{105}
        \\
        \listmethods{0.24}{pixel3-rear-osmo-group3-20190511-3}{000155}{101}
        \\
        \listmethods{0.24}{20191010_006_2}{000006}{105}
    \end{tabular}
    \caption{
        \textbf{Comparisons to the existing methods.}
        Existing global projection methods and adaptive projection approaches~\cite{Zorin:1995:CGP} distort straight lines in the background and create irregular boundaries.
        Contents may be missing after rectangular cropping on the warped videos.
        In the second example, Shih et al.~\cite{Shih:2019:DFW} fail to correct the perspective distortion as the face detector misses the face on the left subject.
        In our method, thanks to the temporal smoothness constraint, we successfully correct the distortion on the left face by leveraging the temporal information across the video.
        %
        Overall, the proposed algorithm renders natural-looking faces, preserves the geometry of the background, and generates regular boundaries.
    }
    \label{fig:comparisons}
\end{figure*}

\begin{figure*}[t!]
    \centering
    \footnotesize
    \renewcommand{\tabcolsep}{1pt} 
	\renewcommand{\arraystretch}{1.0} 
	\newcommand{\listcases}[6]{
	}
    \begin{tabular}{c|ccc}
        Input frame $n$ &
        Output frame $n$ &
        Output frame $n + 2$ &
        Output frame $n + 4$
        \\
	    \begin{overpic}[width=0.24\linewidth,tics=10]{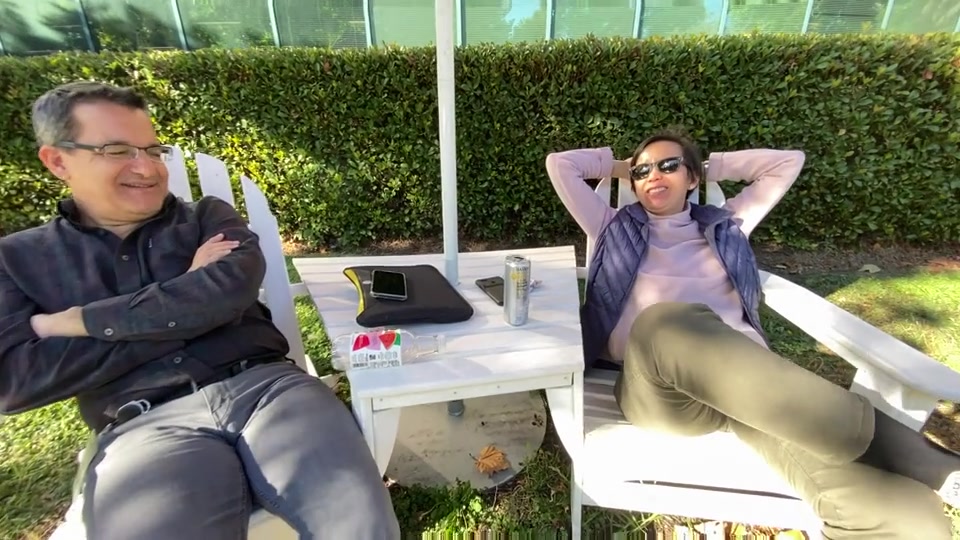}
            \put (3,3) {\white{105$\degree$ FOV}}
        \end{overpic} &
	    \includegraphics[width=0.24\textwidth]{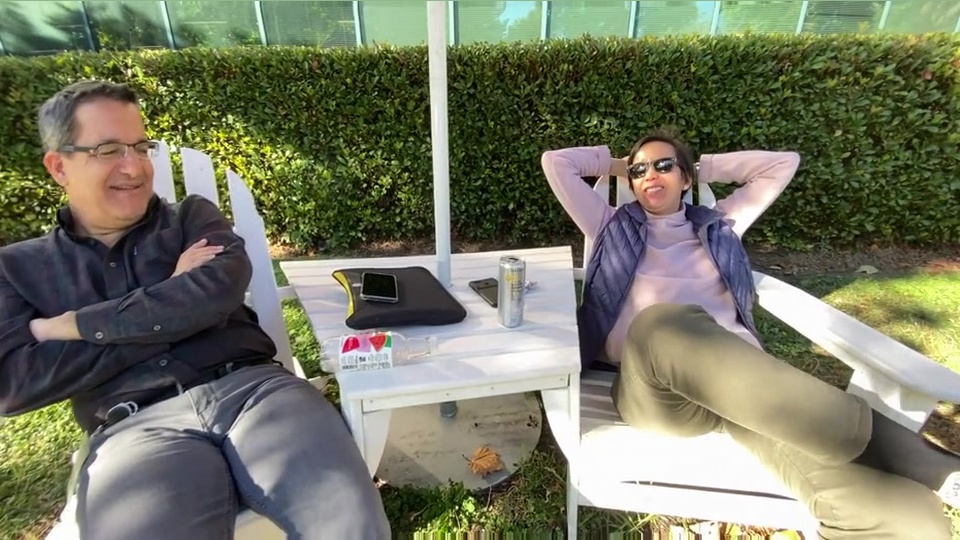} &
	    \begin{overpic}[width=0.24\linewidth]{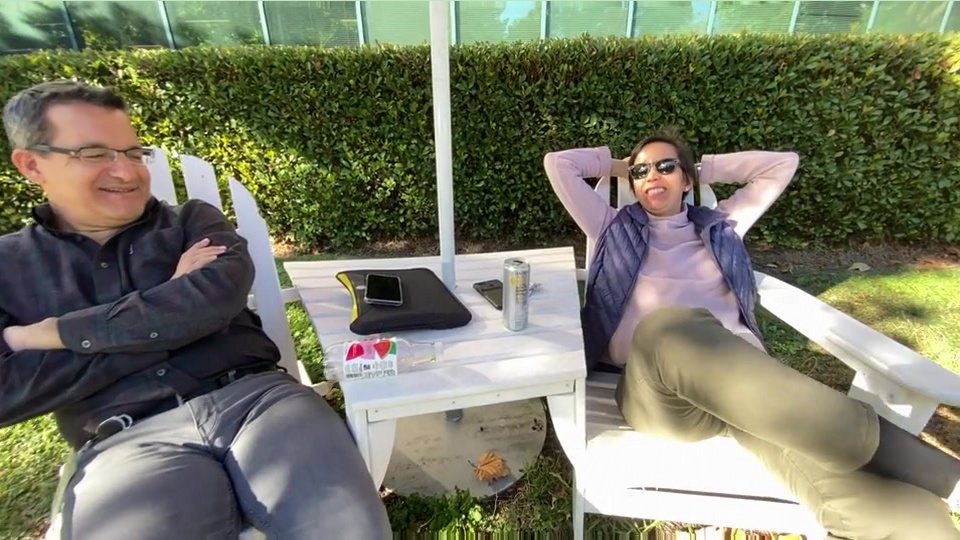}
	        \linethickness{1pt}
            \put(25,55){\color{red}\vector(-1,-1){10}}
        \end{overpic} &
	    \includegraphics[width=0.24\textwidth]{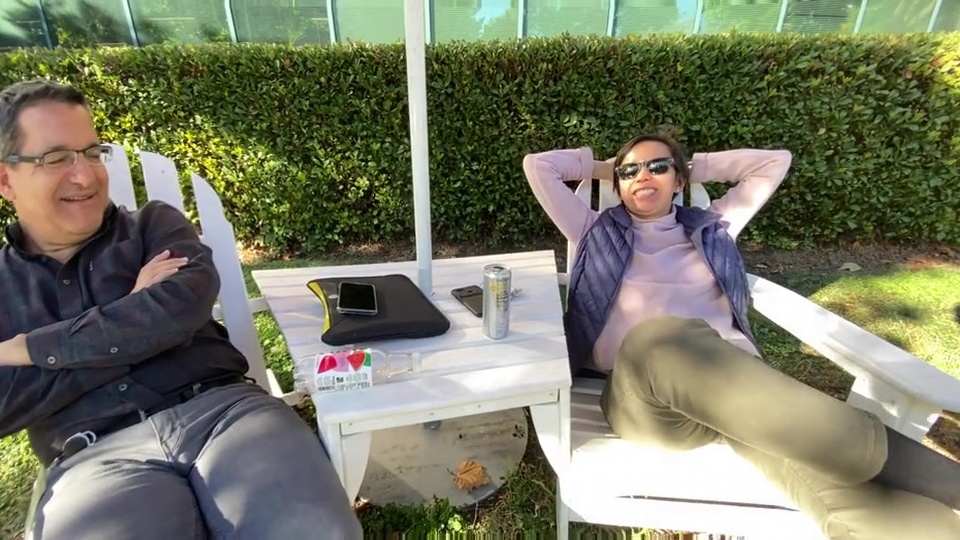} \\
	    & \multicolumn{3}{c}{Shih et al.~\cite{Shih:2019:DFW}} \\
	     &
	    \includegraphics[width=0.24\textwidth]{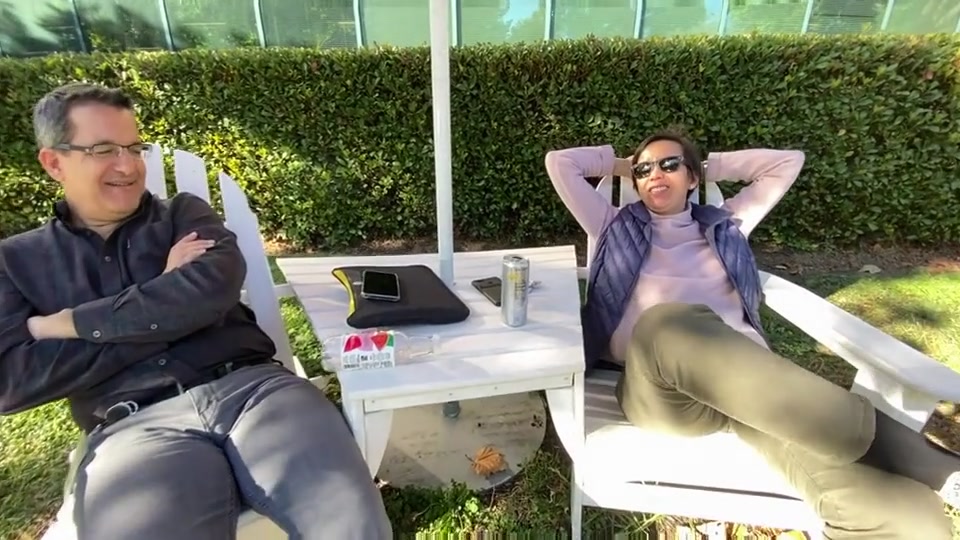} &
	    \includegraphics[width=0.24\textwidth]{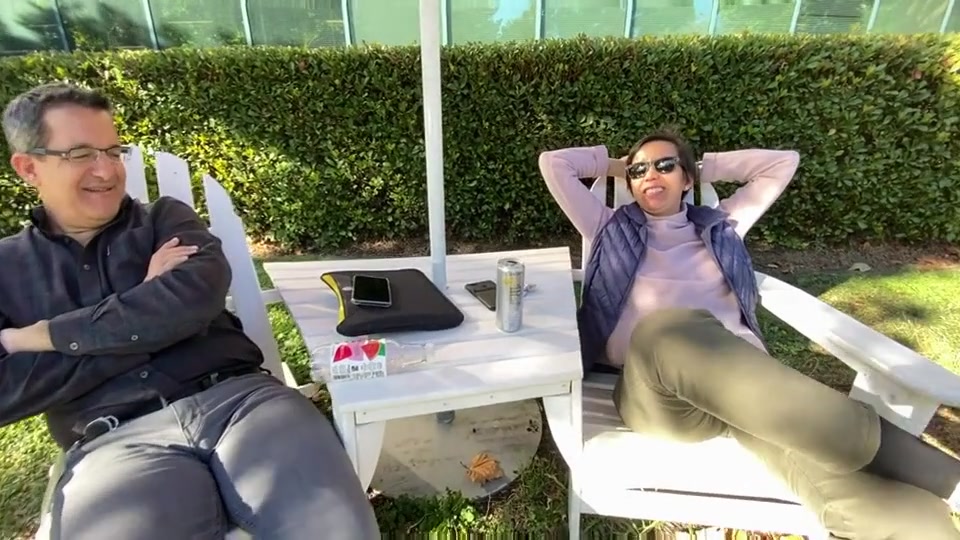} &
	    \includegraphics[width=0.24\textwidth]{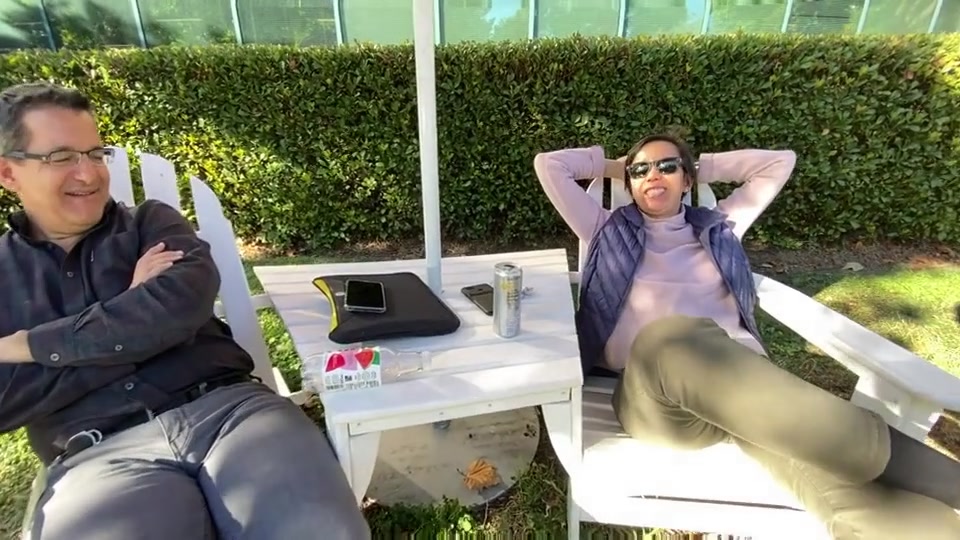}
        \\
	    & \multicolumn{3}{c}{Our method} 
	    \\ 
	    \begin{overpic}[width=0.24\linewidth,tics=10]{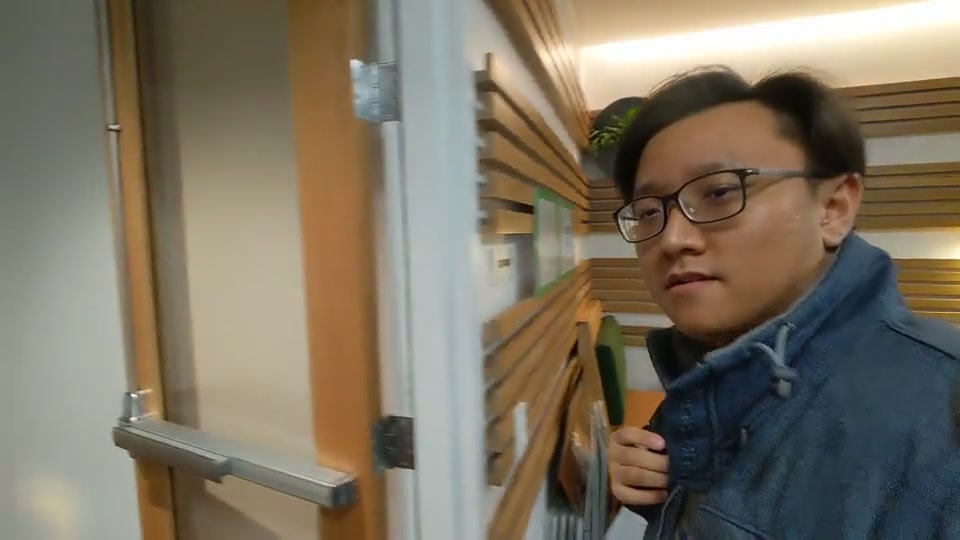}
            \put (3,3) {\white{101$\degree$ FOV}}
        \end{overpic} &
	    \includegraphics[width=0.24\linewidth]{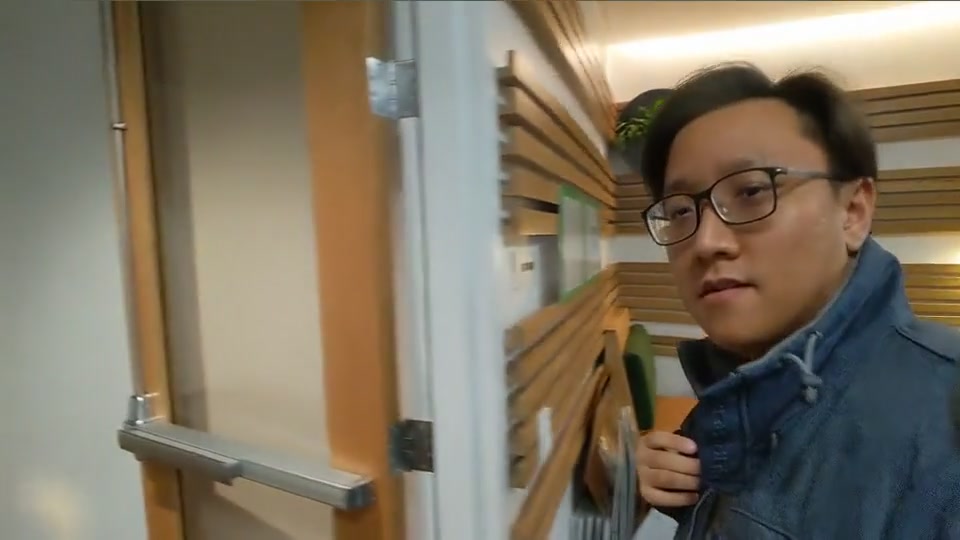} &
	    \begin{overpic}[width=0.24\linewidth]{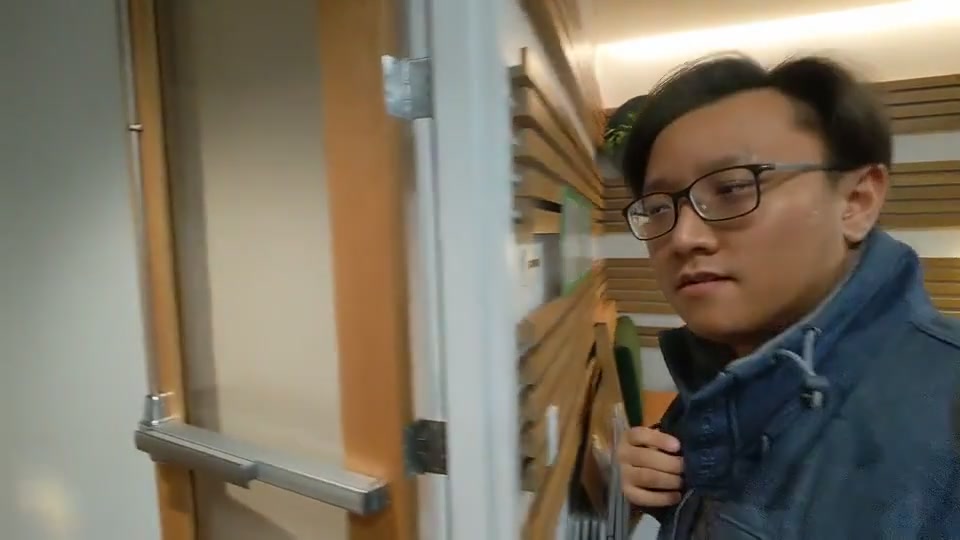}
	        \linethickness{1pt}
            \put(95,56){\color{red}\vector(-1,-1){10}}
        \end{overpic} &
	    \includegraphics[width=0.24\linewidth]{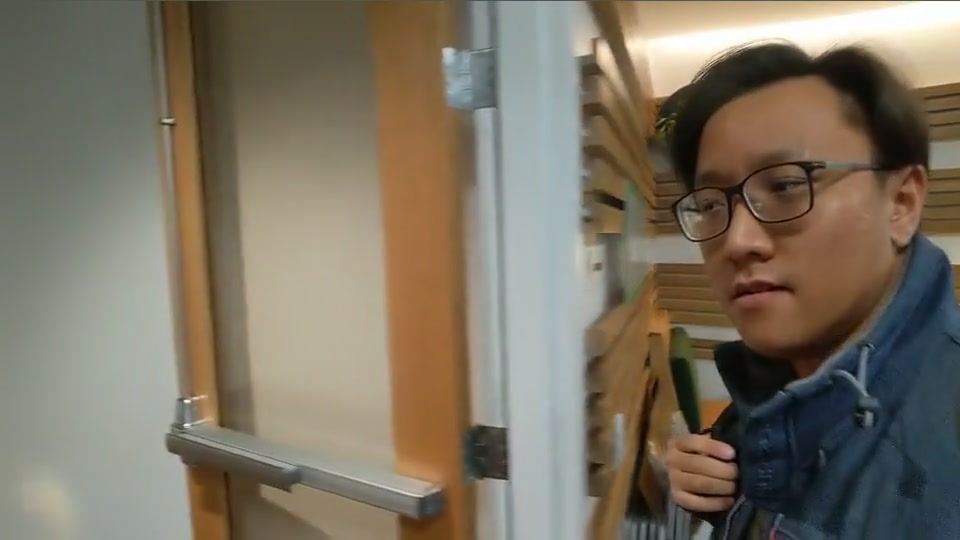}
	    \\
	    & \multicolumn{3}{c}{Shih et al.~\cite{Shih:2019:DFW}} \\
	     &
	    \includegraphics[width=0.24\textwidth]{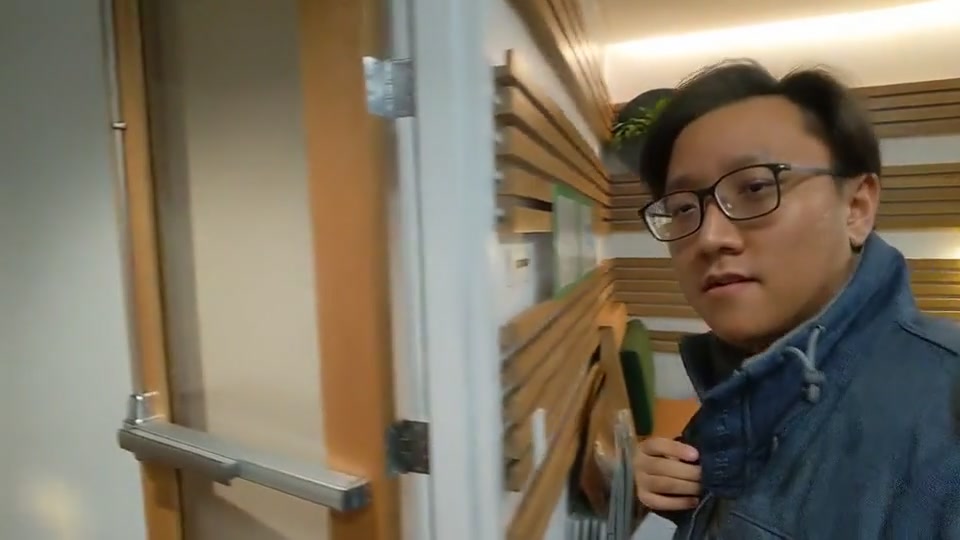} &
	    \includegraphics[width=0.24\textwidth]{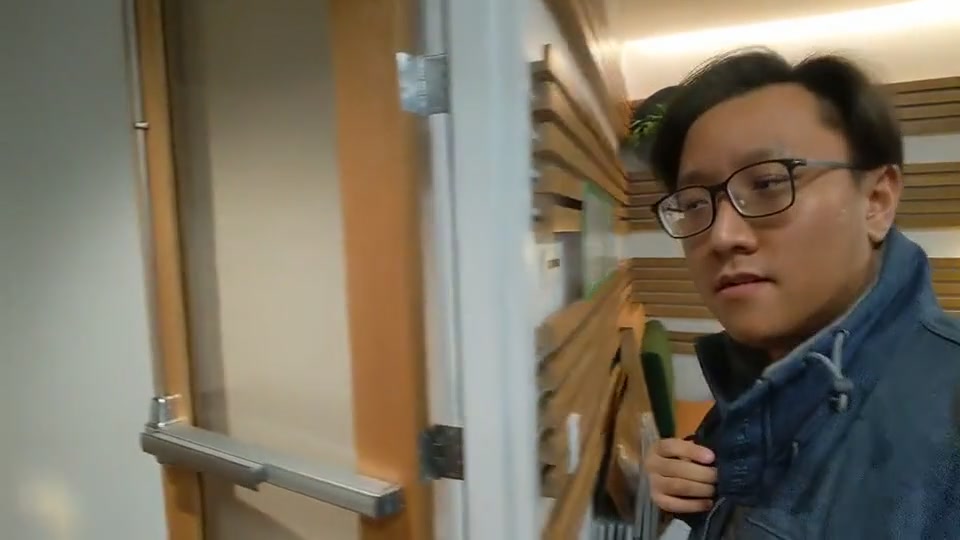} &
	    \includegraphics[width=0.24\textwidth]{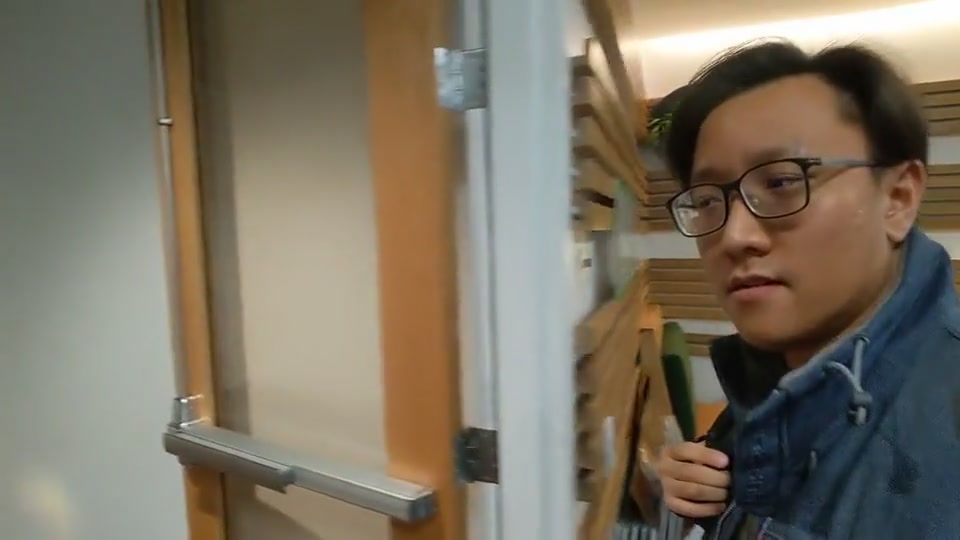}
        \\
	    & \multicolumn{3}{c}{Our method} 
	    \\
	    \begin{overpic}[width=0.24\linewidth,tics=10]{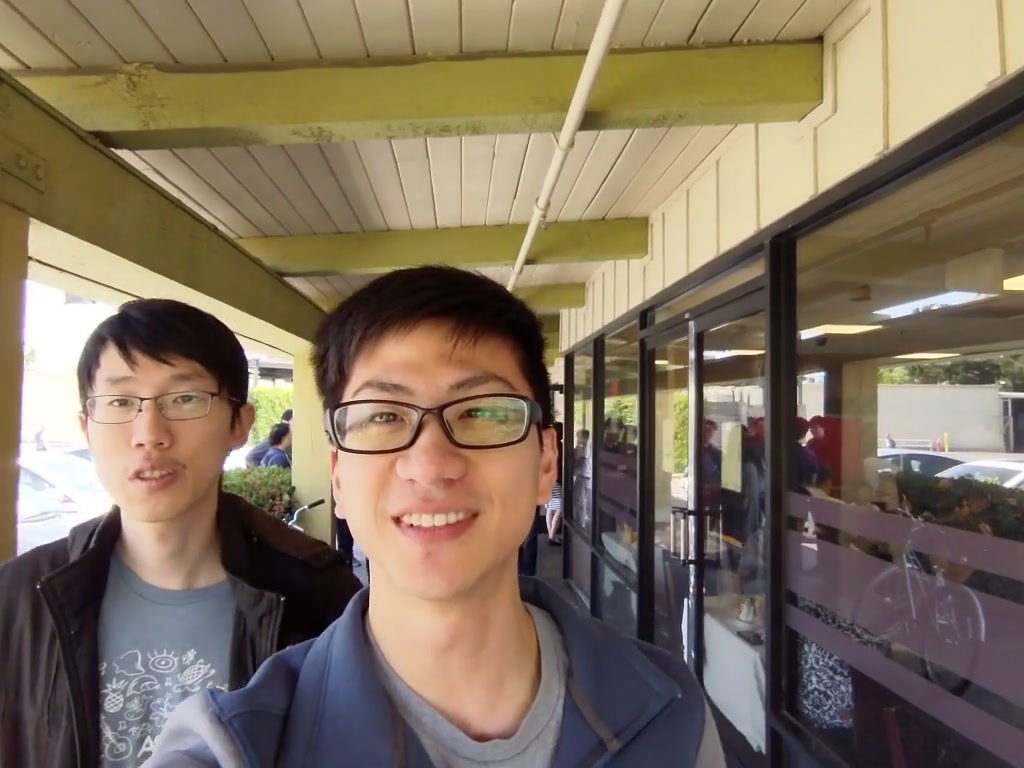}
            \put (3,3) {\white{97$\degree$ FOV}}
        \end{overpic} &
	    \begin{overpic}[width=0.24\linewidth]{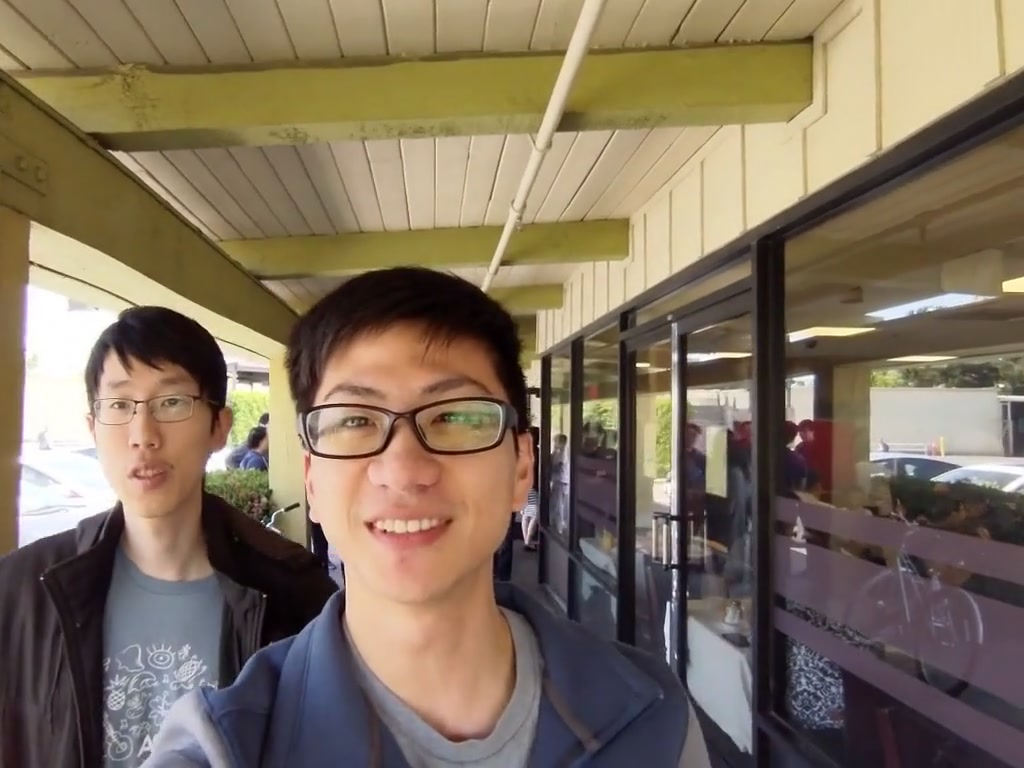}
	        \linethickness{1pt}
            \put(30,56){\color{red}\vector(-1,-1){10}}
        \end{overpic} &
	    \begin{overpic}[width=0.24\linewidth]{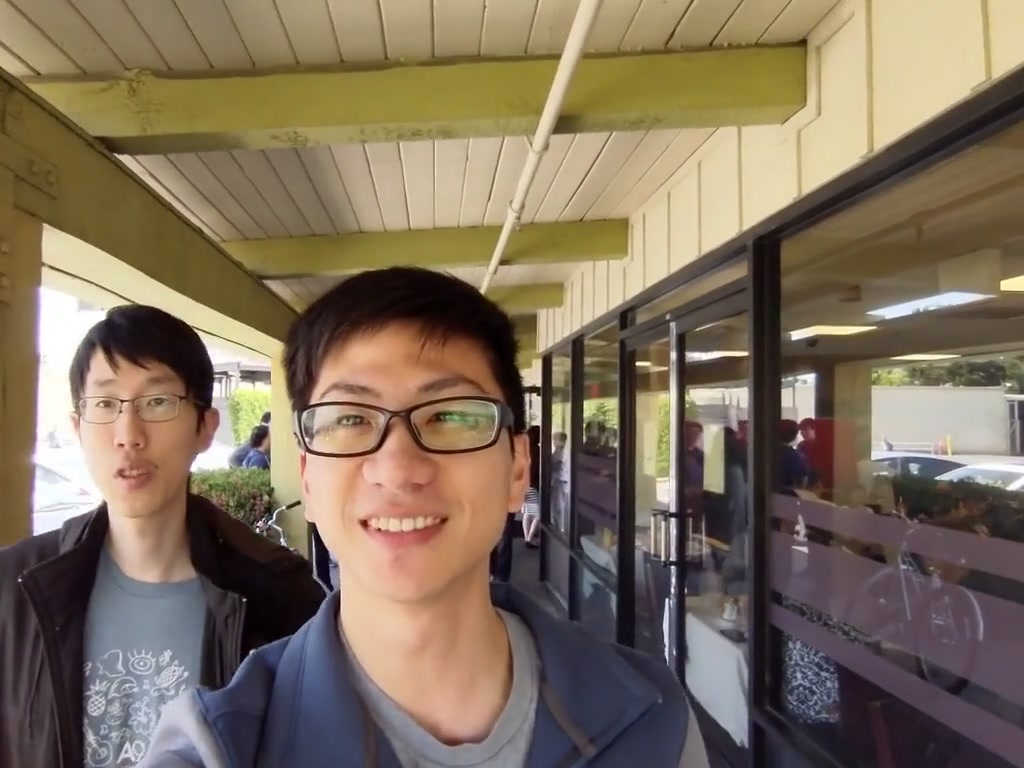}
	        \linethickness{1pt}
            \put(30,56){\color{red}\vector(-1,-1){10}}
        \end{overpic} &
	    \begin{overpic}[width=0.24\linewidth]{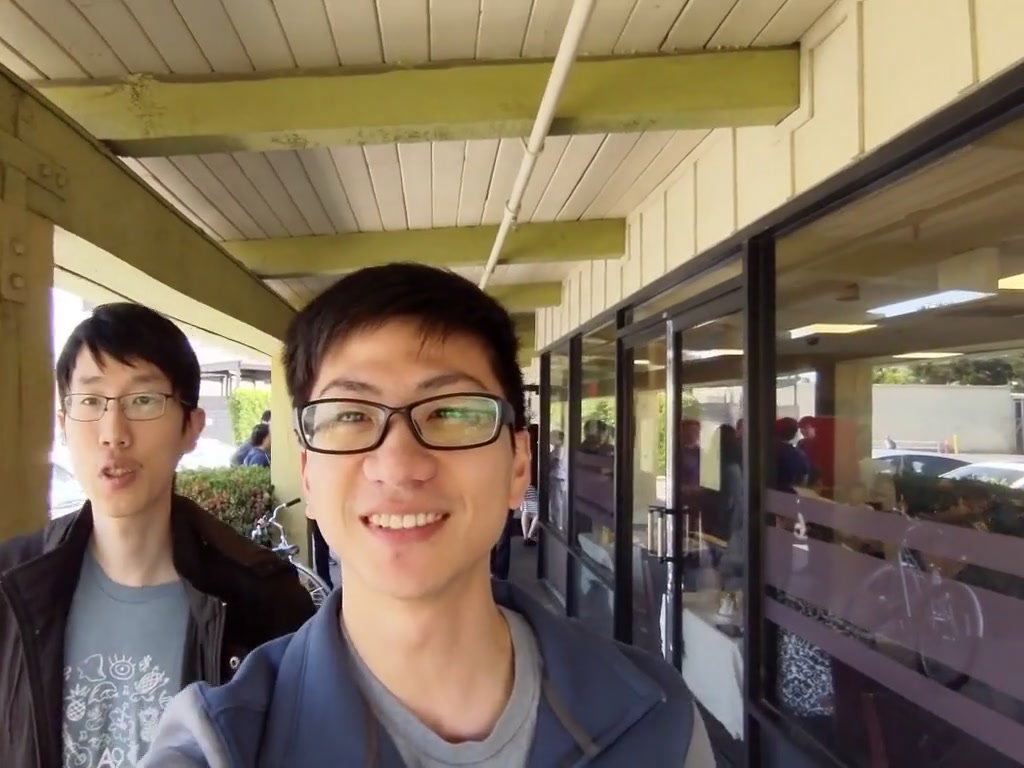}
	        \linethickness{1pt}
            \put(30,56){\color{red}\vector(-1,-1){10}}
        \end{overpic} \\
	    & \multicolumn{3}{c}{Shih et al.~\cite{Shih:2019:DFW}} \\
	     &
	    \includegraphics[width=0.24\textwidth]{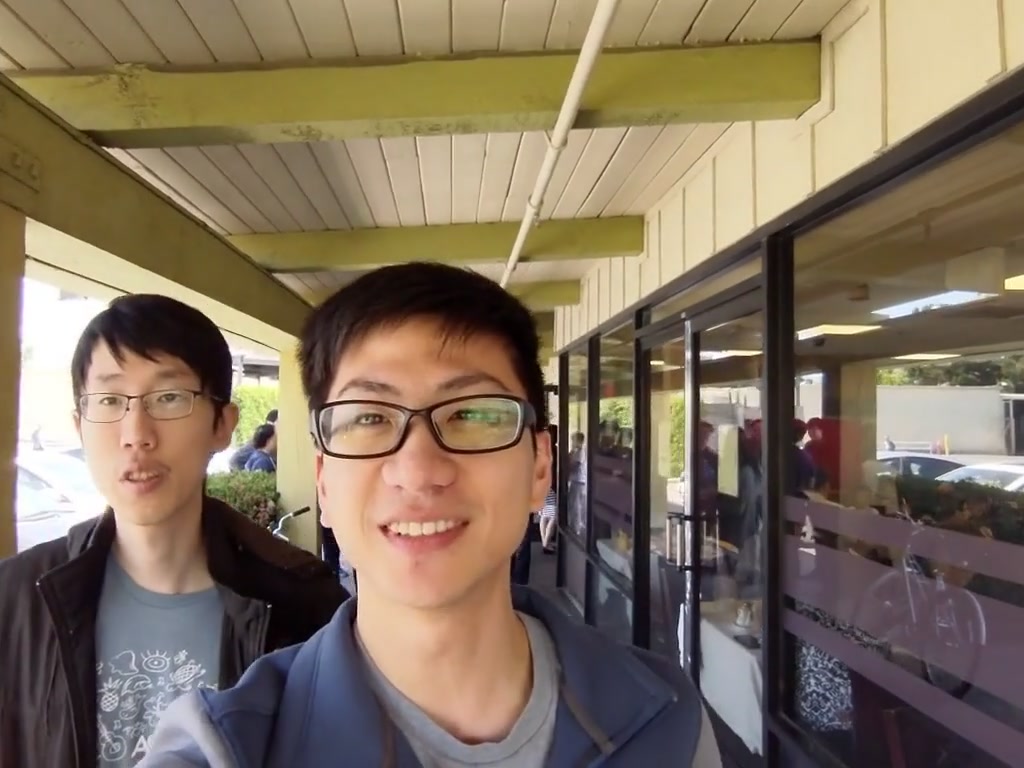} &
	    \includegraphics[width=0.24\textwidth]{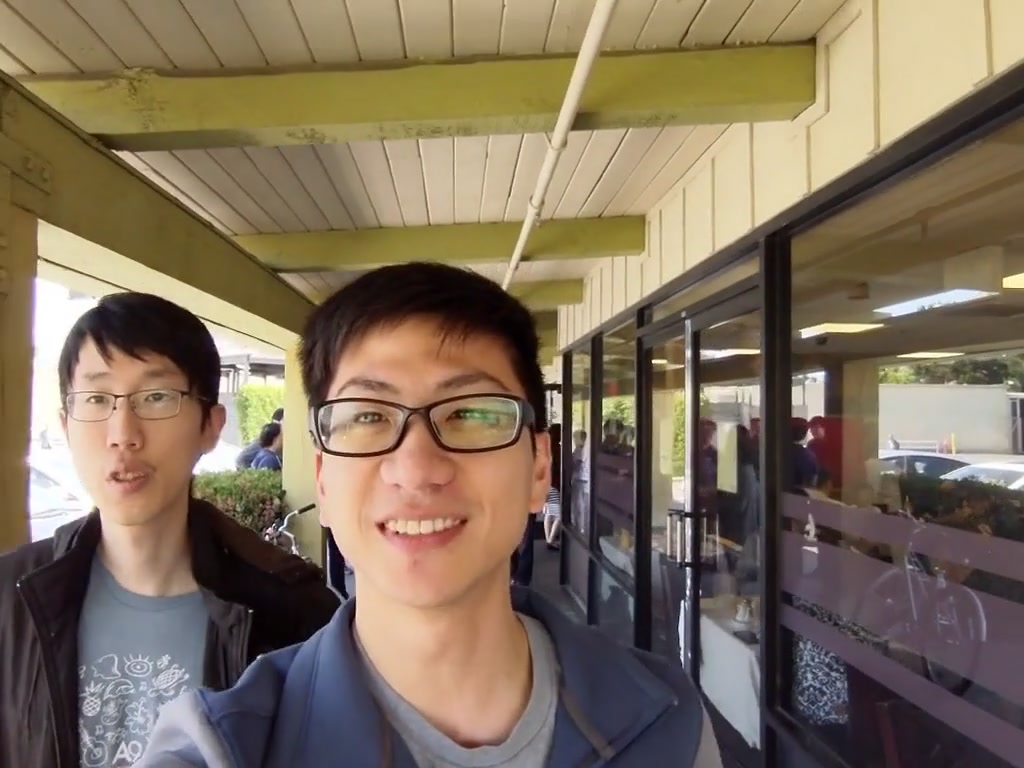} &
	    \includegraphics[width=0.24\textwidth]{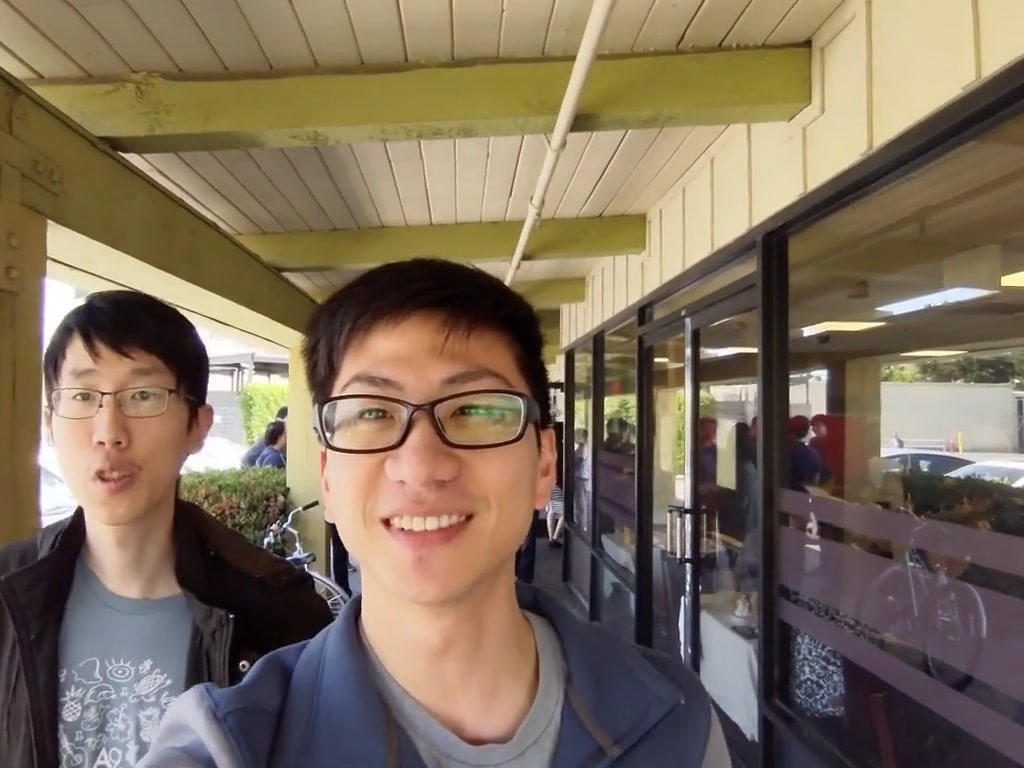}
        \\
	    & \multicolumn{3}{c}{Our method} 
    \end{tabular}
    \caption{
        \textbf{Comparisons with Shih et al.~\cite{Shih:2019:DFW}.}
        The results of the frame-based method~\cite{Shih:2019:DFW} are temporally unstable and exhibit distorted lines. 
        As shown in frame $n + 2$ of the first and second examples, the face detector misses one of the faces. 
        Therefore, the frame-based method falls back to the perspective projection and results in temporal flickering. 
        In the third example, the straight line on the beam is bent by the face on the left. 
        In contrast, our algorithm is able to maintain temporal stability and preserve straight lines.}
    \label{fig:compare_with_shih}
\end{figure*}

\section{Experimental Results}
\label{sec:results}

In this section, we first introduce our wide-angle video database for quality evaluation.
We then evaluate the proposed algorithm against existing methods for correcting face distortion in videos and conduct a subjective user study.
Finally, we discuss the limitations of our method.
More results and videos can be found in the supplementary material. 


\subsection{Wide-Angle Video Database}

We collect a database of 126 wide-angle videos for performance evaluation. 
To cover the common use cases by casual users, we capture videos with both consumer and mobile phone cameras, including the GoPro Hero5 (103\degree~FOV), Pixel 3 rear camera mounted with the  Moment Lens extension~\cite{moment} (101\degree~FOV), Pixel 3 front-facing camera (97\degree~FOV), and iPhone 11 Pro Max rear camera (105\degree~FOV).
For each camera, we collect 14 to 46 video clips.
Table~\ref{tab:dataset} lists the distribution of the database.
The duration of the videos ranges from 10 to 60 seconds, which is well suited to conventional film editing pace.
We first rectify the input videos using the lens calibration parameters embedded in the EXIF meta-data to remove lens distortion.
In addition, the camera is mounted on a DJI Osmo camera gimbal~\cite{osmo} to stabilize the captured videos.
%
Using a stabilizer mimics video blog usages, where mechanical stabilization is frequently used to reduce unwanted motion such as handshakes.
The stabilized videos raise challenges for our method, as subtle temporal inconsistency becomes clearly noticeable.
Our database covers a wide range of complicated scenes under indoor and outdoor conditions.
The number of subjects ranges from 1 to 7 in a scene.
The videos also cover subject activities and camera motions in typical video blogs, such as walking, narrating, moving dynamically to various locations across the camera FOV, interactions between a large group of subjects, camera panning, and transitions from background to human subjects.
The videos are captured in front of challenging man-made objects such as statues, building facades, interiors, and natural scenes like trees and sky. 
In addition, some subjects wear accessories like glasses and hats to raise the complexity of face occlusion handling.
%
In several cases, faces may not be detected as the subjects move back-and-forth quickly in the camera FOV.

\figref{results} shows the results of our distortion correction algorithm on our video database.
For each video, we show the snapshots of the input and three consecutive output frames. 
Our method recovers the natural looks for the subjects and preserves temporal coherence and the shape of background objects. 
When subjects frequently move toward and away from the camera, our warping mesh efficiently adapts to the presence of the subjects without distorting the background.
When subjects are present in front of the architecture of many straight edges, our method preserves the line structures and corrects the distortion for the subjects.
%
More results are presented in the supplementary video. 
%
%

\subsection {Performance Evaluations}

\figref{comparisons} shows the results of our algorithm and other methods based on stereographic, Pannini~\cite{Sharpless:2010:PAN}, Mercator~\cite{mercator}, and adaptive~\cite{Zorin:1995:CGP} projections. 
In the first row, the horizontal structure in the background is clearly distorted in the stereographic and Mercator projection but is rendered as horizontal edges by our method.
In the second example, the face of the left subject in the Pannini projection appears elongated and unnatural, while our method corrects the perspective distortion.
These methods generate irregular boundaries and may lose video content after cropping the output videos. 
In contrast, our algorithm corrects the wide-angle distortions and renders natural looks of both the subjects and background structures. 

We compare the rendered results by our algorithm and the frame-based method by Shih et al.~\cite{Shih:2019:DFW} in~\figref{compare_with_shih} and the accompanying video. 
Since the frame-based method is designed to process one single photo at a time, the rendered images contain significant flickering artifacts. 
We further apply a simple FIR filter on the warping mesh generated by~\cite{Shih:2019:DFW}, but find that such a post-processing step is not able to remove the artifacts. 
In contrast, our approach renders temporally consistent results.
When the face detection is missing or occluded, or the segmentation mask is inaccurate, the spatial-temporal optimization approach smoothly interpolates the mesh to reduce the flickering effects effectively.



\begin{table}[]
    \centering
        \caption{\textbf{User study.} Each number shows the percentage of users that favor our method when compared against the result processed by the other method.}
    \label{tab:user-study}
    \footnotesize
    \begin{tabular}{l|ccc}
        \toprule
        The other method & 97\degree & 101\degree & 103\degree \\
        \midrule
        Input (perspective) & 
        $80.0\%$ & $91.7\%$ & $80.0\%$ \\
        Stereographic & 
        $90.9\%$ & $50.0\%$ & $85.7\%$ \\
        Pannini~\cite{Sharpless:2010:PAN} & 
        $85.7\%$ & $75.0\%$ & $50.0\%$ \\
        Zorin and Barr~\cite{Zorin:1995:CGP} & 
        $57.1\%$ & $83.3\%$ & $71.4\%$ \\
        Shih~et.~al.~\cite{Shih:2019:DFW} & 
        $100.0\%$ & $100.0\%$ & $100.0\%$ \\
        \bottomrule
    \end{tabular}

\end{table}

\nothing{
\begin{figure}[t!]
    \centering
    \includegraphics[width=0.9\linewidth]{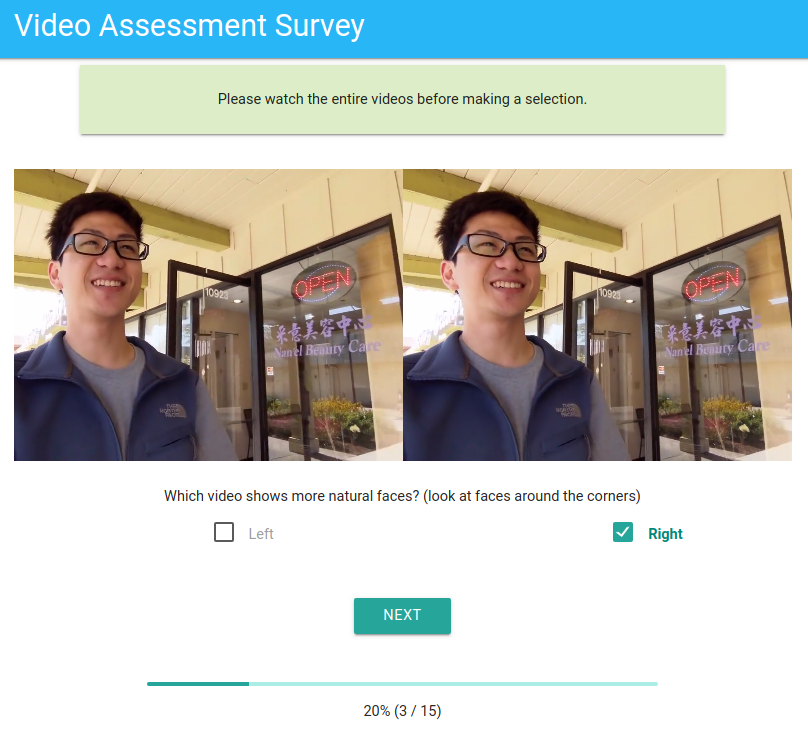}
    \caption{
        \textbf{Snapshot of the user study interface.}
    }
    \label{fig:user-study}
\end{figure}
}

\subsection{User Study}
We conduct a user study to evaluate our algorithm against the video-based approaches based on perspective projection, stereographic projection, Pannini projection~\cite{Sharpless:2010:PAN}, and adaptive projection~\cite{Zorin:1995:CGP}.
In addition, we evaluate our method against the frame-based method by Shih~\etal~\cite{Shih:2019:DFW}. 
The study includes 36 videos randomly sampled from our wide-angle video dataset with different FOVs.
We play the videos processed by our approach and one of the other methods in random left-right order and ask the participants to vote for the preferred result.
The user study includes 20 participants and each participant is asked to vote on 24 video pairs.
Overall, the results rendered by our method are favored over the other schemes, as illustrated in Table~\ref{tab:user-study}. 
%
%
When comparing to Shih~\etal~\cite{Shih:2019:DFW}, all the users prefer our results due to the temporal stability.

\nothing{\ckliang{Need more elaboration?}\yichang{Brought some contexts from the table to the main texts.}}

\subsection{Ablation Study}

\begin{figure}
    \centering
    \footnotesize
    \renewcommand{\tabcolsep}{1pt} 
    \begin{tabular}{cc}
        \begin{overpic}[width=0.49\linewidth,tics=10]{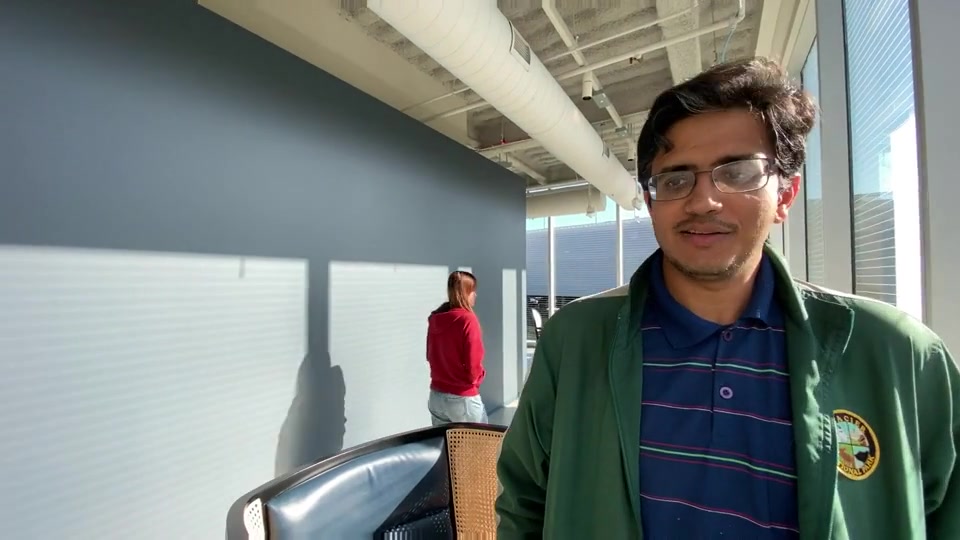}
            \put (3,48) {\white{$105\degree$ FOV}}
            \put (3,40) {\white{Frame $n$}}
        \end{overpic} &
        \begin{overpic}[width=0.49\linewidth,tics=10]{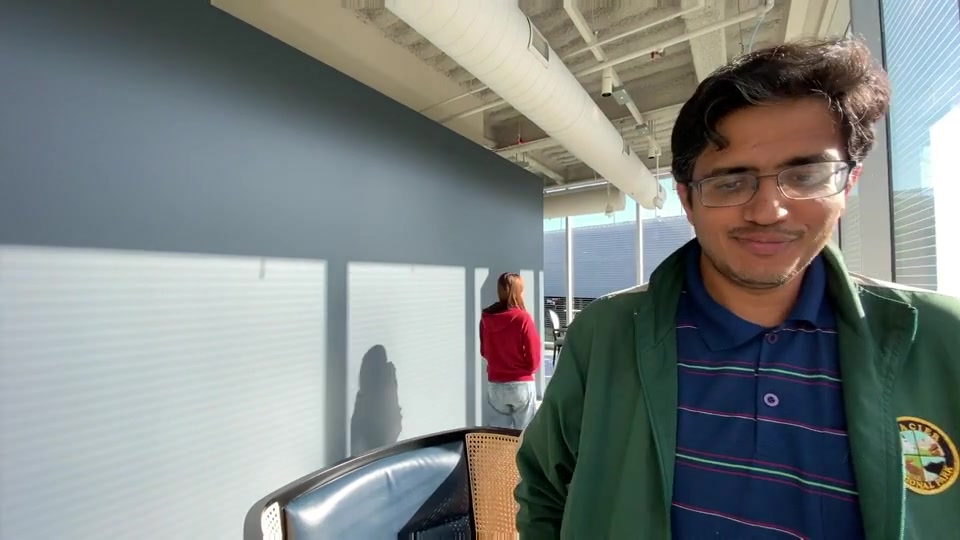}
            \put (3,48) {\white{$105\degree$ FOV}}
            \put (3,40) {\white{Frame $n + 10$}}
        \end{overpic}
        \\
        \multicolumn{2}{c}{Input (105\degree~FOV)}
        \\
        \begin{overpic}[width=0.49\linewidth]{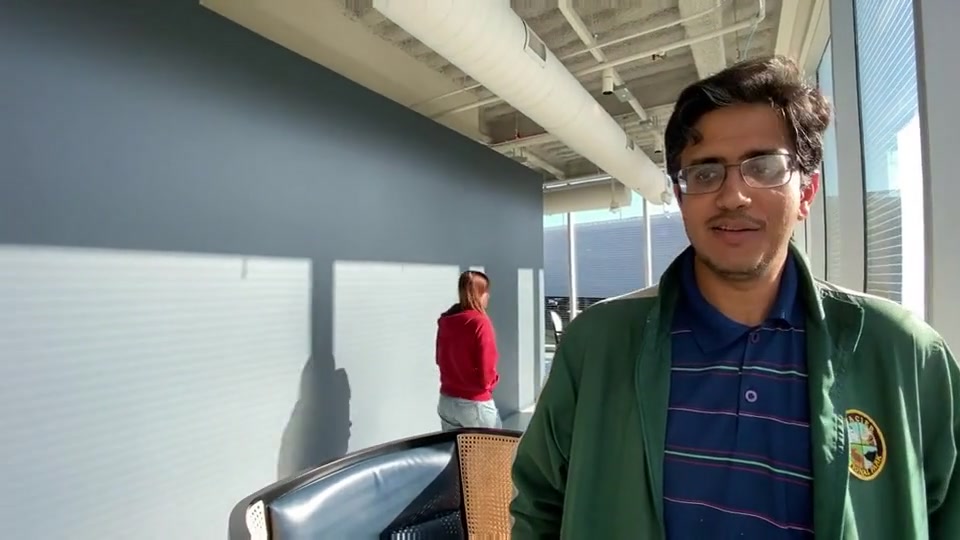}
	        \linethickness{1pt}
            \put(100,40){\color{red}\vector(-1,-1){8}}
        \end{overpic}
        & 
        \begin{overpic}[width=0.49\linewidth]{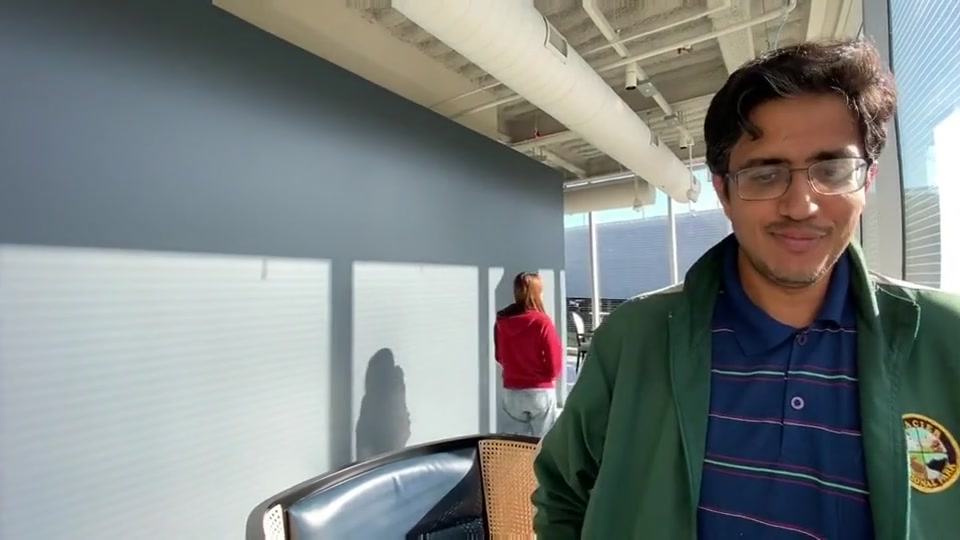}
	        \linethickness{1pt}
            \put(100,53){\color{red}\vector(-1,-2){5}}
        \end{overpic}
        \\
        \multicolumn{2}{c}{Without the line-preservation term in~\eqnref{single_line_term}}
        \\
        \includegraphics[width=0.49\linewidth]{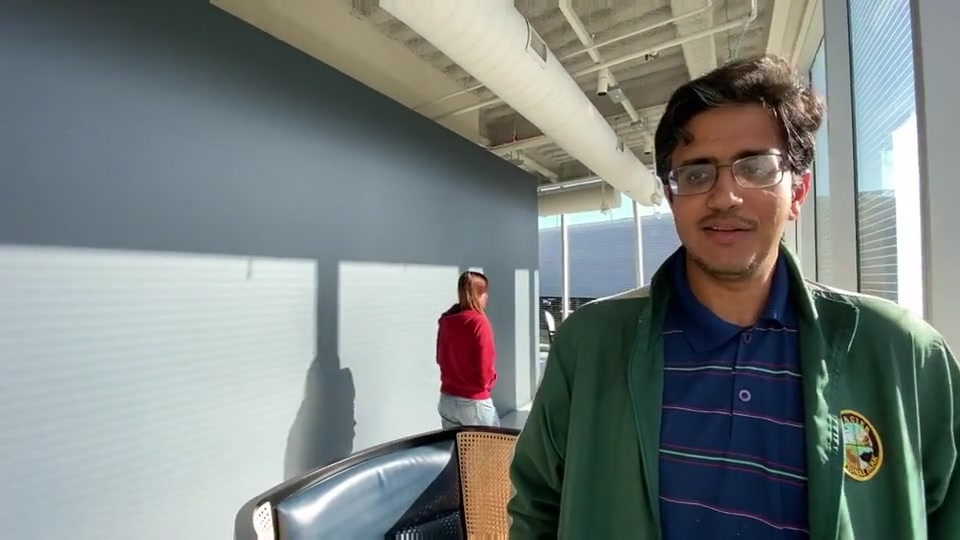} &
        \includegraphics[width=0.49\linewidth]{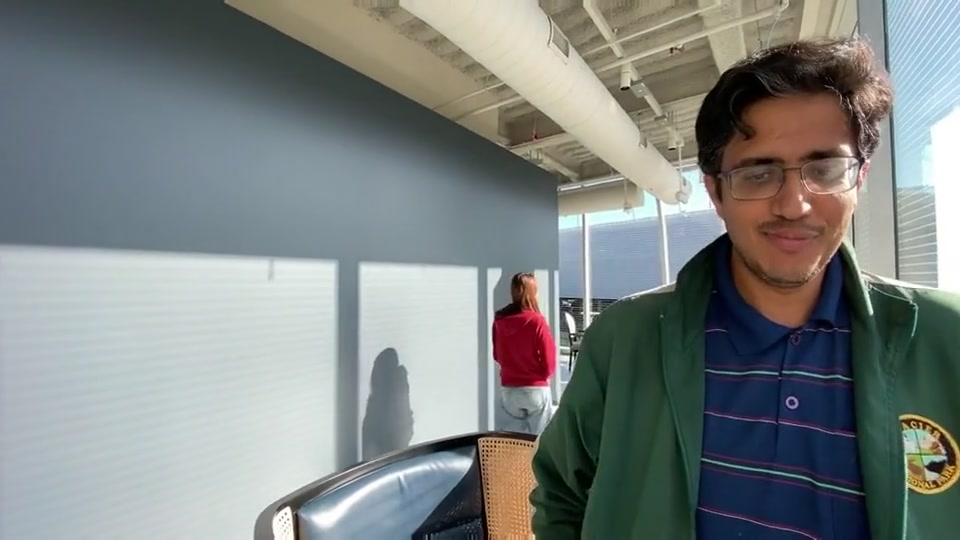}
        \\
        \multicolumn{2}{c}{Our method}
    \end{tabular}
    \vspace{-2mm}
    \caption{
        \textbf{Top:} Two frames from a wide-angle video (105\degree~FOV, 30 FPS) with straight lines in the background.
        \textbf{Middle:} Without the line-preservation term in~\eqnref{single_line_term},
        straight lines may become curved when appearing close to the faces.
        \textbf{Bottom:}
        Our method maintains straight lines in the background.
    }
    \label{fig:effect_of_line_preservation_term}
\end{figure}

\Paragraph{Line preservation.}
In~\figref{effect_of_line_preservation_term}, we visualize the effect of the line preservation term.
Without the line preservation term, background lines closer to the face may be distorted, such as the long straight lines in buildings illustrated in the middle of~\figref{effect_of_line_preservation_term}.

We note that the line-bending term in~\cite{Shih:2019:DFW} (named as grid edge-bending term $\energy_\gridedgeterm$ in this paper) is different from our line-preservation term.
The line-bending term in~\cite{Shih:2019:DFW} regularizes the shape of mesh grids to avoid shearing on grids, \textit{implicitly} preserving the line structure in the background.
When optimizing jointly in the spatial and temporal domains, the meshes are determined by considering both temporal smoothness and grid structure.
We observed that the straight lines may be easily bent even with the regularization of the line-bending term in~\cite{Shih:2019:DFW}, as shown in the second-row of~\figref{effect_of_line_preservation_term}.
To address this, our line-preservation term \textit{explicitly} guides the proposed method to detect the semantic edges in the background and preserve straightness.
Both the line-bending term in~\cite{Shih:2019:DFW} and line-preservation term are necessary for the proposed method to achieve high-quality warping results.

\begin{figure}
    \centering
    \footnotesize
    \renewcommand{\tabcolsep}{1pt} 
    \begin{tabular}{cc}
        \includegraphics[width=0.49\linewidth]{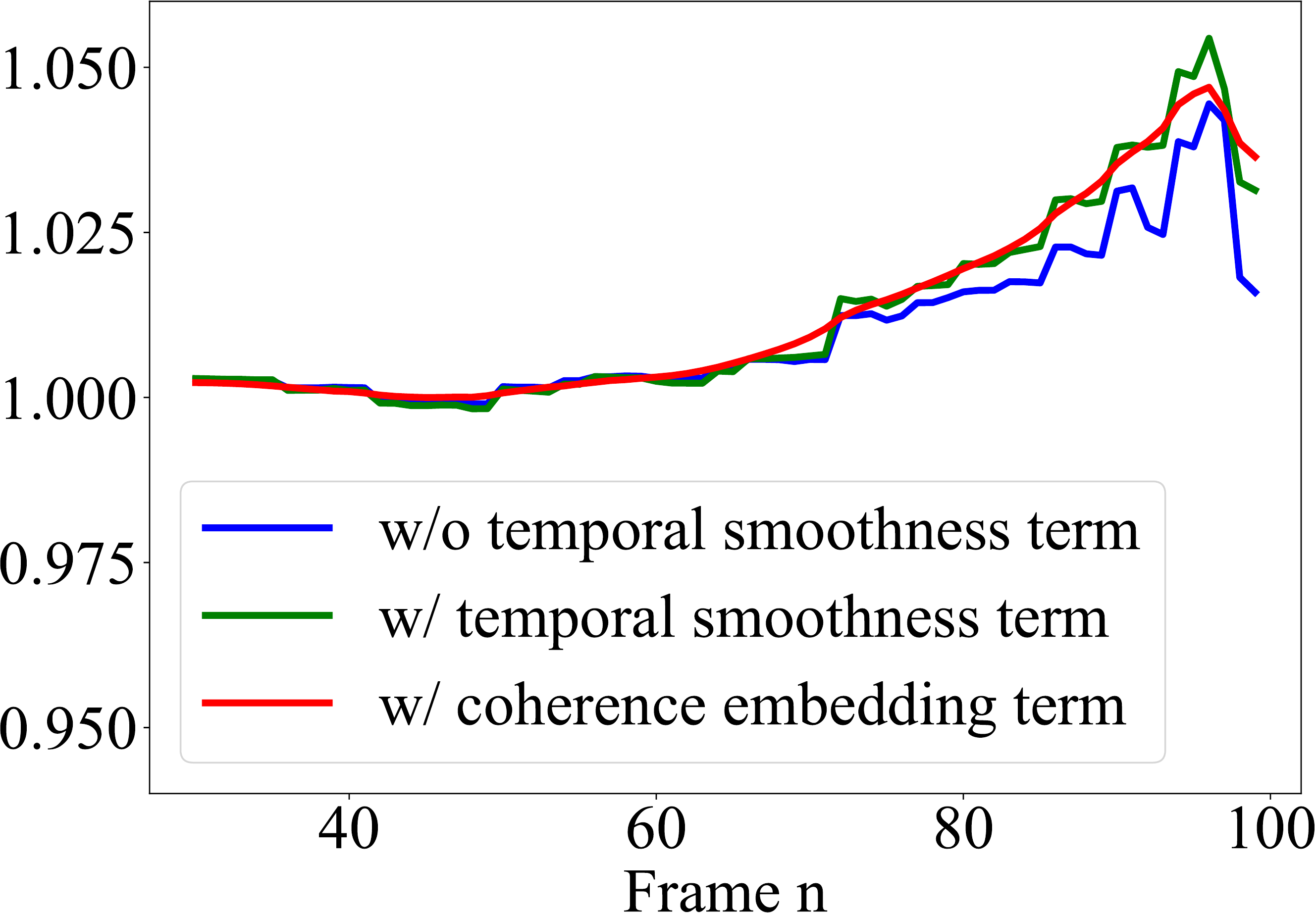} & 
        \includegraphics[width=0.49\linewidth]{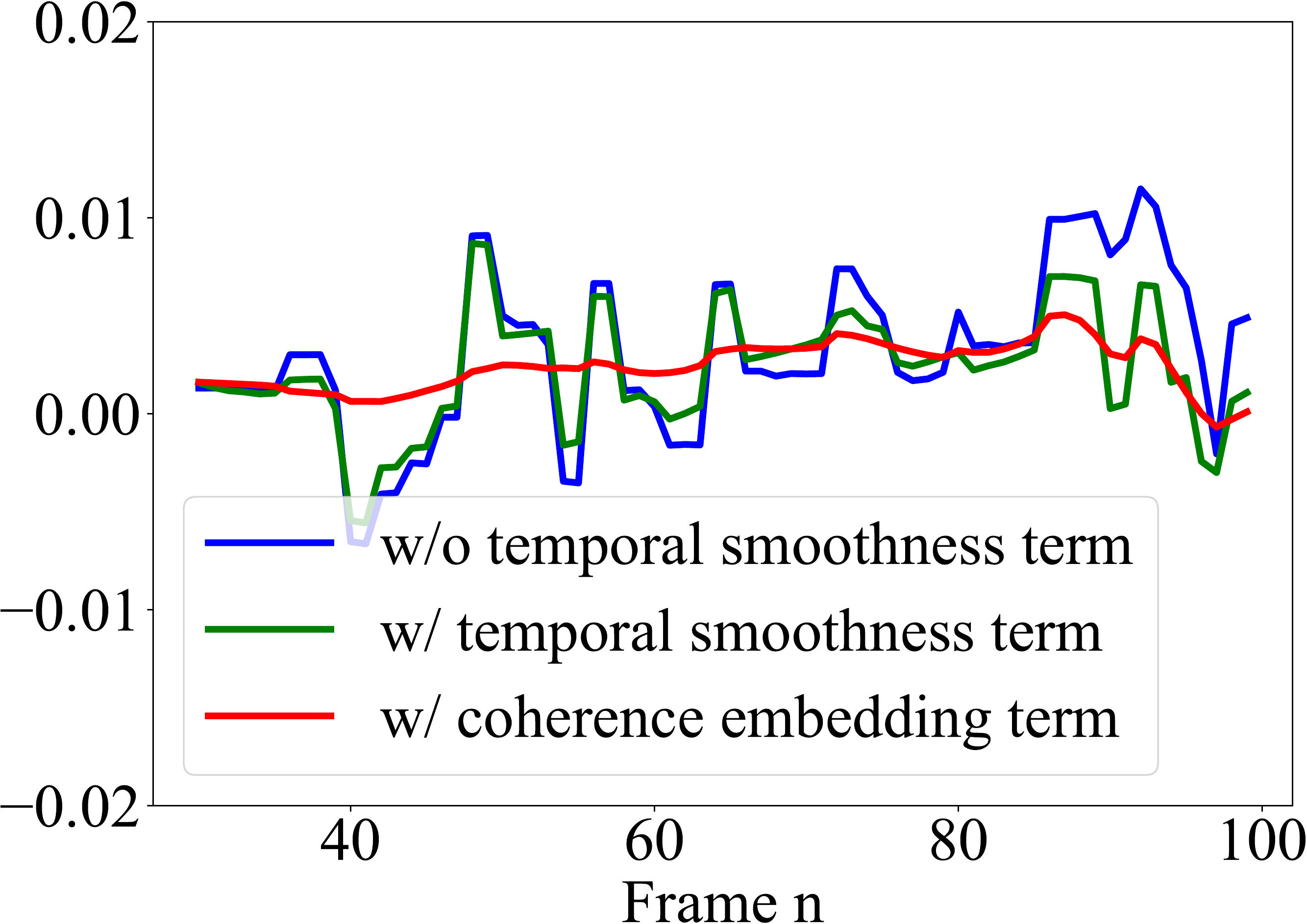} 
        \\
        (a) $a_k$ in~\eqnref{face_affine} & 
        (b) $b_k$ in~\eqnref{face_affine} 
        \\
        \includegraphics[width=0.49\linewidth]{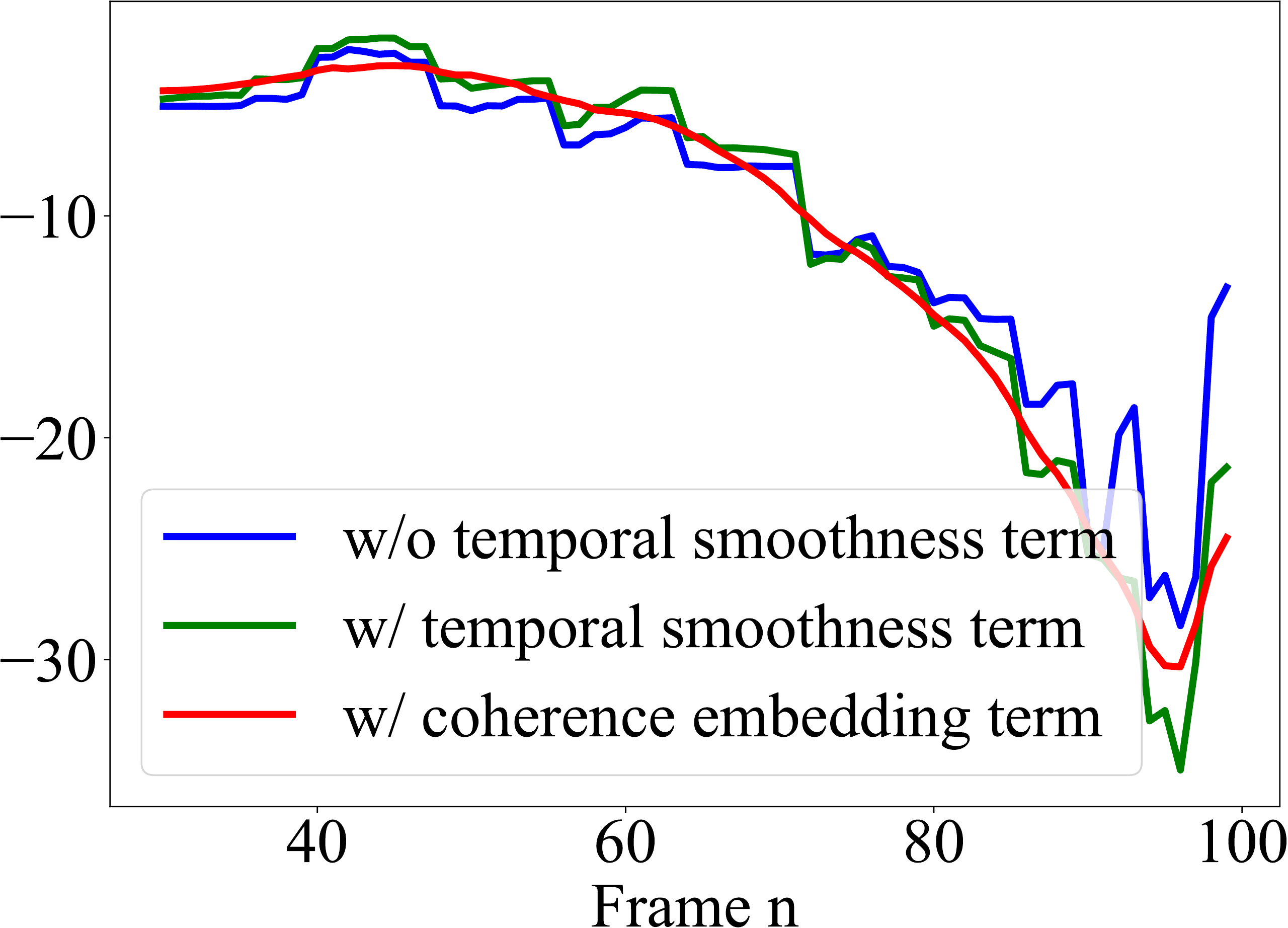} & 
        \includegraphics[width=0.49\linewidth]{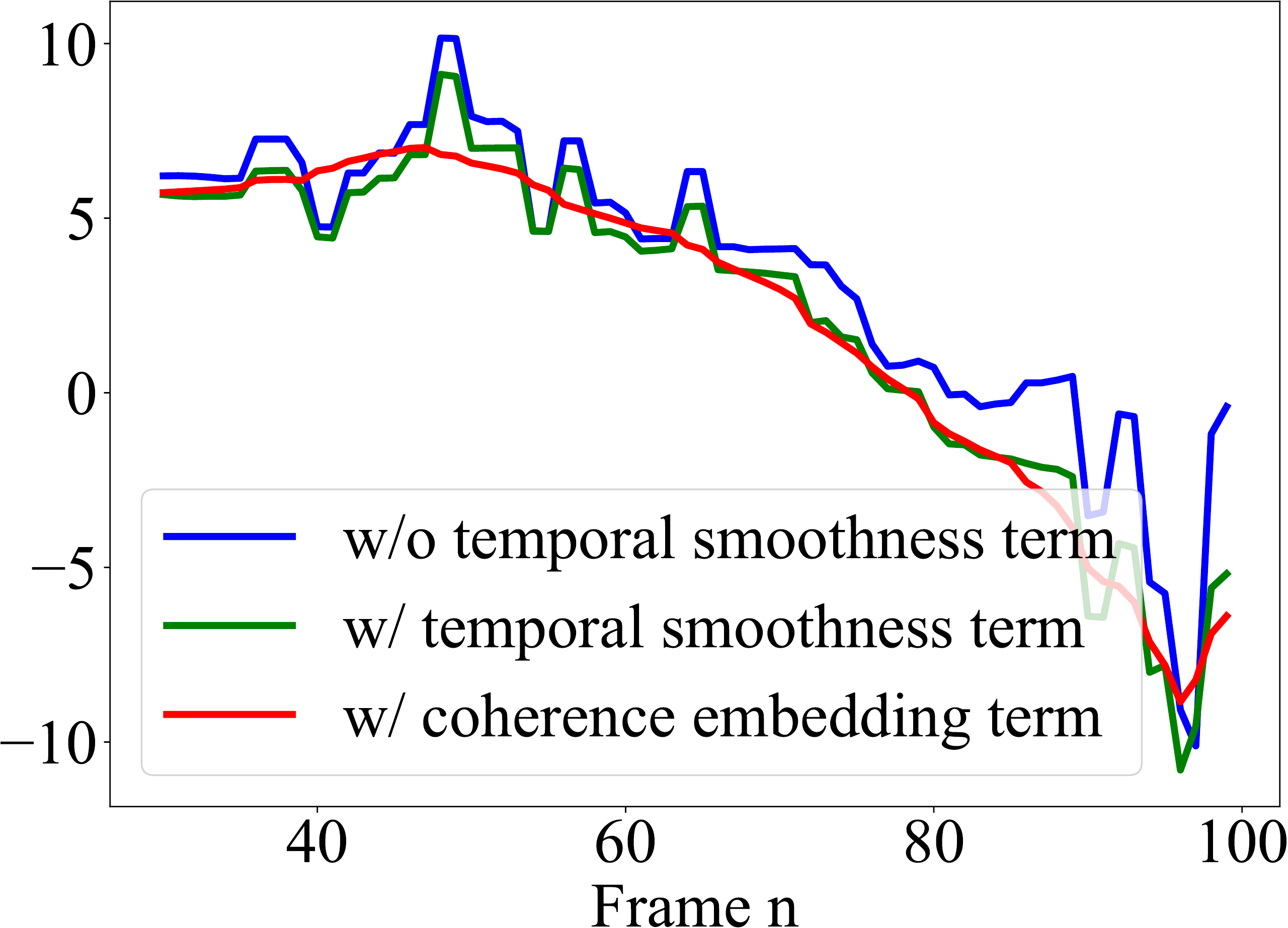}
        \\
        (c) $t_{x,k}$ in~\eqnref{face_affine} & 
        (d) $t_{y,k}$ in~\eqnref{face_affine} 
        \\
    \end{tabular}
    \vspace{-2mm}
    \caption{Without the temporal smoothness term, the similarity transform parameters of faces jitters (\blue{blue curves}).
    Adding the temporal smoothness term in~\eqnref{temporal_term} slightly alleviates the instability (\green{green curves}).
    With the coherent embedding term in~\eqnref{coherence_embedding_term}, the similarity transform becomes temporally stable (\red{red curves}).
    }
    \label{fig:effect_of_coherence_embedding_term}
\end{figure}

\Paragraph{Temporal consistency.}
We plot the similarity transform parameters $a$, $b$, $t_x$, and $t_y$ in~\eqnref{face_affine} of a tracked face in~\figref{effect_of_coherence_embedding_term}.
The temporal smoothness term (green curves) can reduce the jitters in the input (blue curves), and the coherent embedding term in~\eqnref{coherence_embedding_term} facilitates rendering more temporally stable results (red curves).

\begin{figure}[t!]
    \centering
    \footnotesize
    \renewcommand{\tabcolsep}{1pt} 
	\renewcommand{\arraystretch}{0.8} 
    \begin{tabular}{cc}
        \begin{overpic}[width=0.49\linewidth,tics=10]{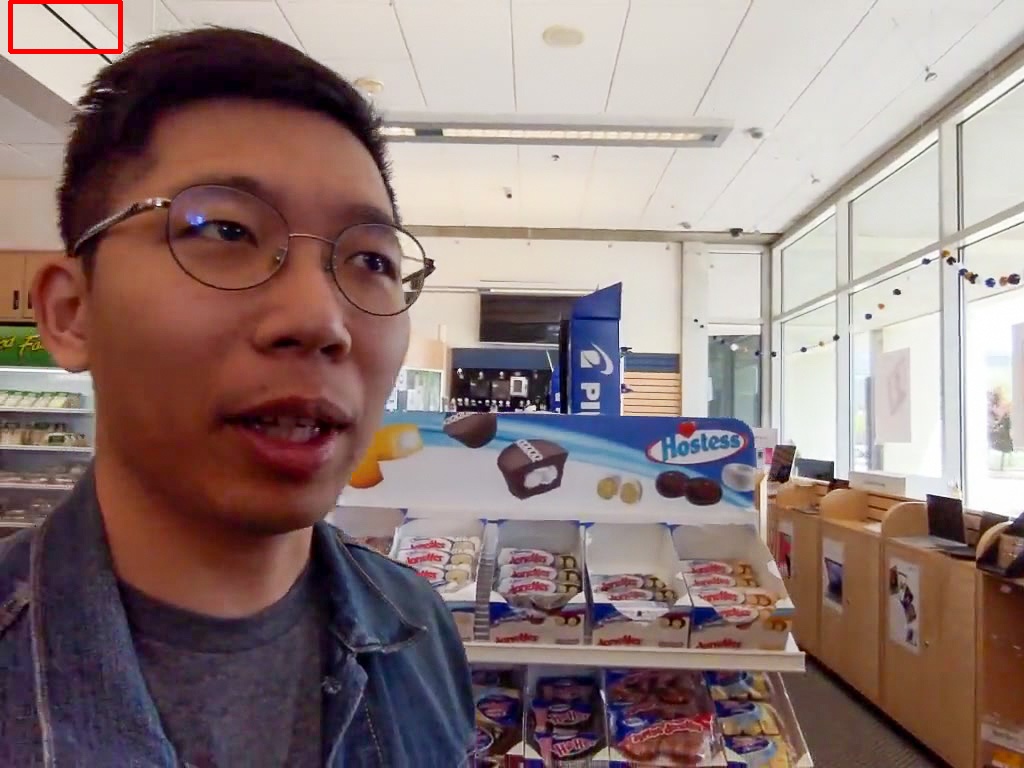}
            \put (3,11) {\white{$97\degree$ FOV}}
            \put (3,3) {\white{Frame $n$}}
        \end{overpic} & 
        \begin{overpic}[width=0.49\linewidth,tics=10]{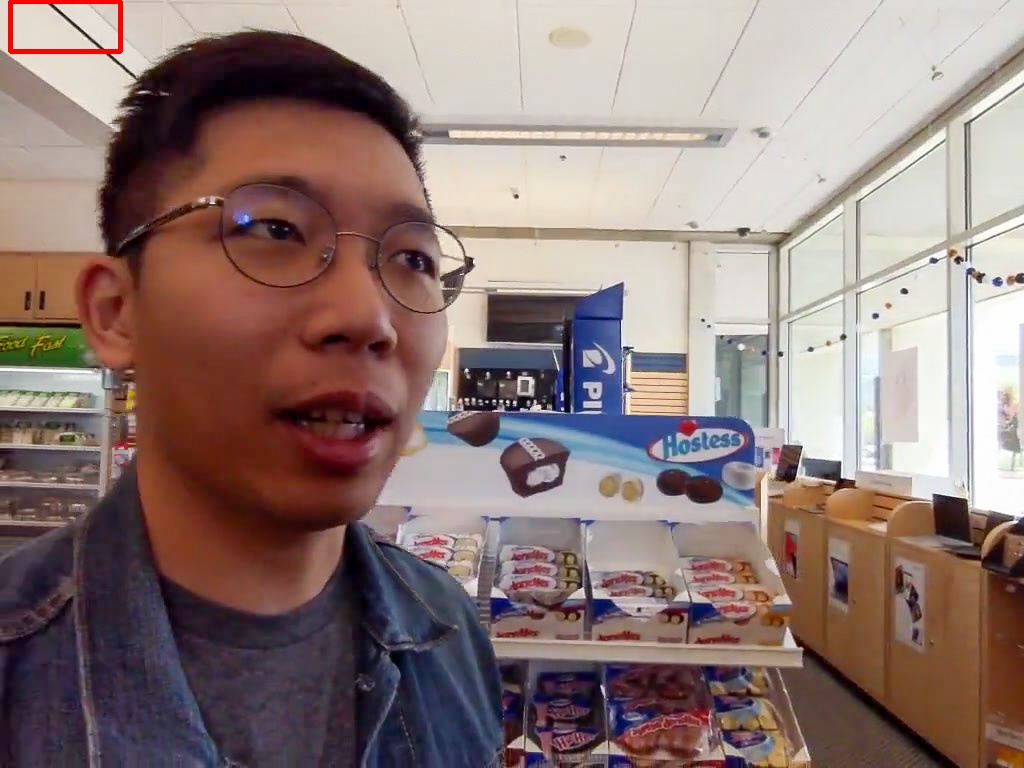}
            \put (3,11) {\white{$97\degree$ FOV}}
            \put (3,3) {\white{Frame $n + 2$}}
        \end{overpic}
        \\
        \multicolumn{2}{c}{Input (97\degree~FOV)}
        \\
        \multicolumn{2}{c}{
            \begin{tabular}{ccc}
                \begin{overpic}[width=0.325\linewidth,tics=10]{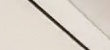}
                    \put (3,3) {\black{Frame $n$}}
                \end{overpic} & 
                \begin{overpic}[width=0.325\linewidth,tics=10]{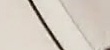}
                    \put (3,3) {\black{Frame $n$}}
                \end{overpic} & 
                \begin{overpic}[width=0.325\linewidth,tics=10]{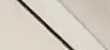}
                    \put (3,3) {\black{Frame $n$}}
                \end{overpic}
                \\
                \begin{overpic}[width=0.325\linewidth,tics=10]{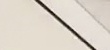}
                    \put (3,3) {\black{Frame $n + 2$}}
                \end{overpic} & 
                \begin{overpic}[width=0.325\linewidth,tics=10]{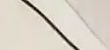}
                    \put (3,3) {\black{Frame $n + 2$}}
                \end{overpic} & 
                \begin{overpic}[width=0.325\linewidth,tics=10]{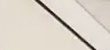}
                    \put (3,3) {\black{Frame $n + 2$}}
                \end{overpic}
                \\
                Input & 
                Optimization in~\eqnref{temporal_term_seq} &
                Our method
            \end{tabular}
        }
    \end{tabular}
    \vspace{-1mm}
    \caption{
        \textbf{Top:} two frames from an input video (30 FPS).
        \textbf{Bottom-left:} A straight line in the background.
        %
        \textbf{Bottom-middle:} The sequential optimization in~\eqnref{temporal_term_seq} is unable to predict the face change and results in visible wobbling and line distortion in the background.
        \textbf{Bottom-right:} Our full-volume optimization method takes all the frames into account to maintain temporal stability for both the background and foreground.
    }
    \label{fig:full-volume}
\end{figure}

\Paragraph{Mesh optimization}
The proposed spatial-temporal optimization task plays a critical role in correcting distortions on face and background regions.  
\figref{full-volume} shows the results of the proposed method based on~\eqnref{total_cost_video} and the alternative based on sequential optimization.
The sequential method starts by optimizing the first frame without a temporal regularization term.
For subsequent frames, the mesh from the previous frame is used as a fixed constraint and temporal regularization between the two consecutive frames is enforced by:
\begin{align} \label{eq:temporal_term_seq}
    \sum_{i} \normtwo{ \vertex_i^{(n)} - \tilde{\vertex}_i^{(n-1)} }\,.
\end{align}
where $\tilde{\vertex}_i^{(n-1)}$ is the optimized mesh vertices at frame $n - 1$ and fixed during optimization. 
While the differences in the formula appear subtle, the results from the sequential optimization approach are not stable (face information cannot be propagated in a reversed-time manner and straight lines near the facial areas are distorted), as shown in~\figref{full-volume}.
Our spatial-temporal optimization generates visually more pleasing results.

While our full-volume optimization can achieve high-quality video results, sequential optimization is more efficient and more suitable for real-time applications.
To include future information in the sequential optimization, we can introduce $K$ look-ahead frames to~\eqnref{temporal_term_seq}, at the cost of $K$-frame delay.
Furthermore, we can apply the full-volume optimization to a short clip instead of the entire video, e.g., optimizing $P$ past frames and $K$ future frames to obtain the optimal mesh at the current frame.
However, this strategy may introduce more computational costs.
The performance and quality trade-off will require further analysis and explorations in a future study.

\begin{figure}
    \centering
    \includegraphics[width=1.0\linewidth]{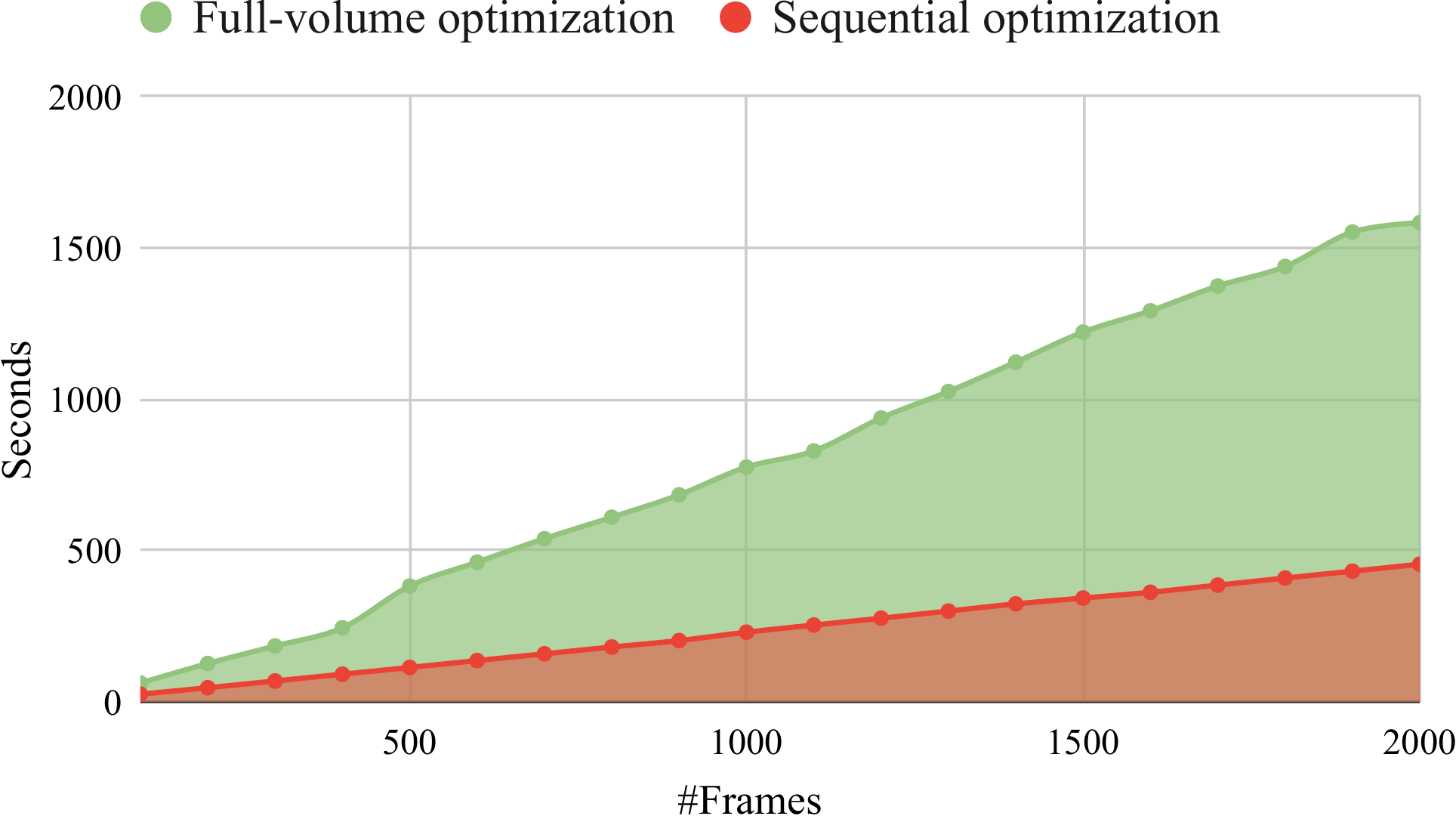}
    \caption{
        \textbf{Execution time.}
        The processing time of both the full-volume and sequential optimization methods grows linearly with the number of frames.
        The sequential optimization scheme is approximately $3\times$ faster than the full-volume optimization approach.
    }
    \label{fig:linear}
\end{figure}

\subsection{Execution time}
We evaluate the processing time of our method on a Linux desktop with an Intel W-2135 CPU.
When processing a video with frame resolution $1024 \times 768$, the processing time is roughly linear to the frame numbers as illustrated in~\figref{linear}, as the temporal energy terms are included sparsely with the neighboring frames. 
Without code optimization of our implementation, it takes $0.8$ seconds to process one frame. 
Table~\ref{tab:performance} shows the processing time of each key step. 
Our method can be further optimized by using native languages such as C++ or leveraging GPUs.

\begin{table}[t!]
    \centering
        \caption{
        \textbf{Processing time.}
        We show the average processing time of each step on a frame with a resolution of $1024 \times 768$.
        The execution time is evaluated on a Linux desktop with an Intel W-2135 CPU.
    }
    \label{tab:performance}
    \footnotesize
    \begin{tabular}{c|cc}
        \toprule
        Stage               &   Time (ms)   & Percentage \\
        \midrule
        Face tracking and segmentation   &    63      &    7.5$\%$ \\
        Line detection and tracking       &    37      &    4.5$\%$ \\
        Mesh optimization   &   642      &   76.3$\%$ \\
        Image warping       &    99      &   11.7$\%$ \\
        Total               &   841      &    100$\%$ \\
        \bottomrule
    \end{tabular}

\end{table}

\begin{figure*}[t!]
    \centering
    \footnotesize
    \renewcommand{\tabcolsep}{1pt} 
	\renewcommand{\arraystretch}{1.0} 
	\renewcommand{\imagewidth}{0.24\linewidth}
    \begin{tabular}{cccc}
        Input frame $n$ &
        Input frame $n + 10$ &
        Input frame $n + 20$ &
        Input frame $n + 30$
        \\
	    \begin{overpic}[width=\imagewidth,tics=10]{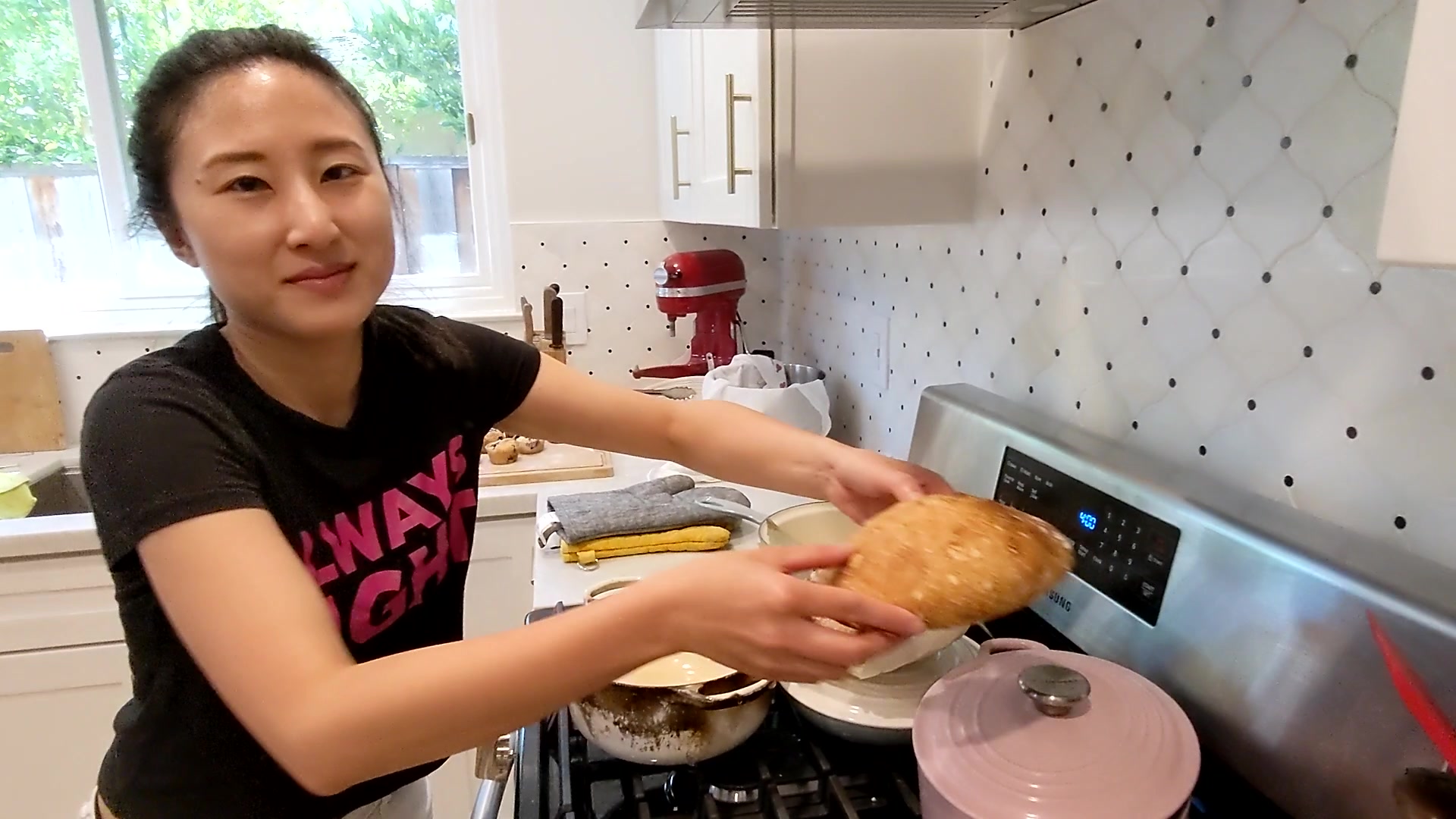}
            \put (3,3) {\white{$90\degree$ FOV}}
        \end{overpic} &
	    \includegraphics[width=\imagewidth,tics=10]{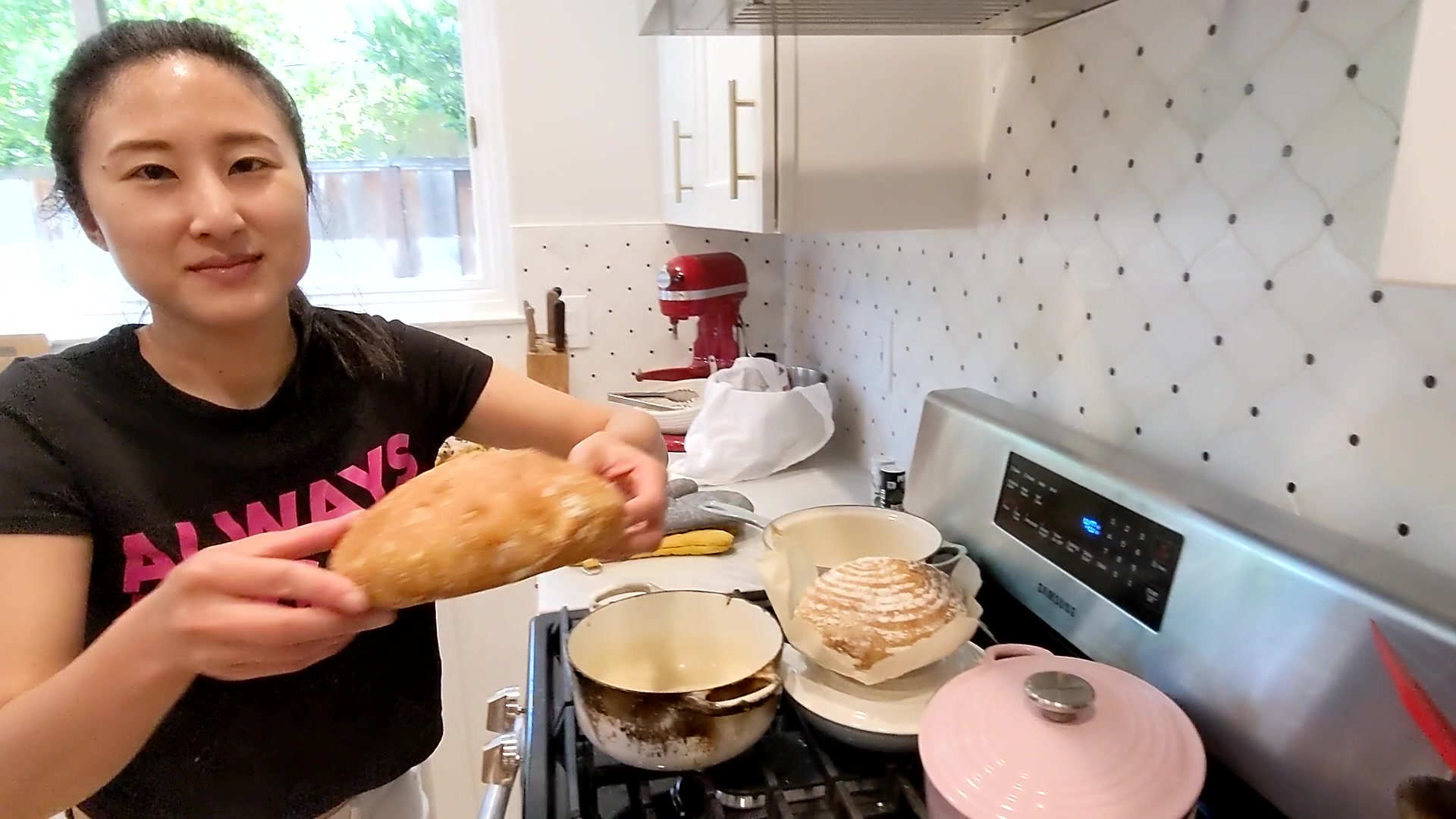} &
	    \includegraphics[width=\imagewidth,tics=10]{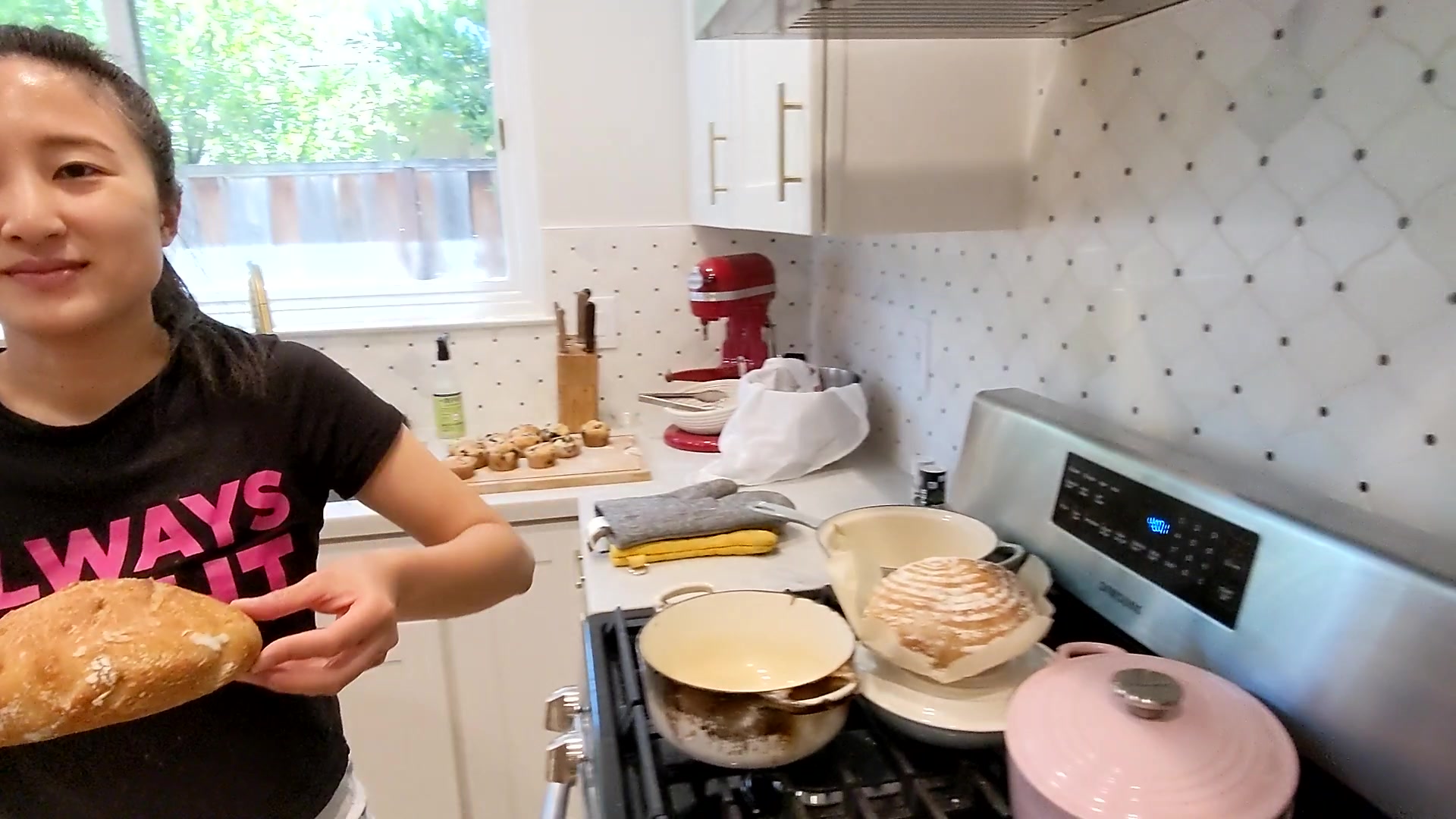} &
	    \includegraphics[width=\imagewidth,tics=10]{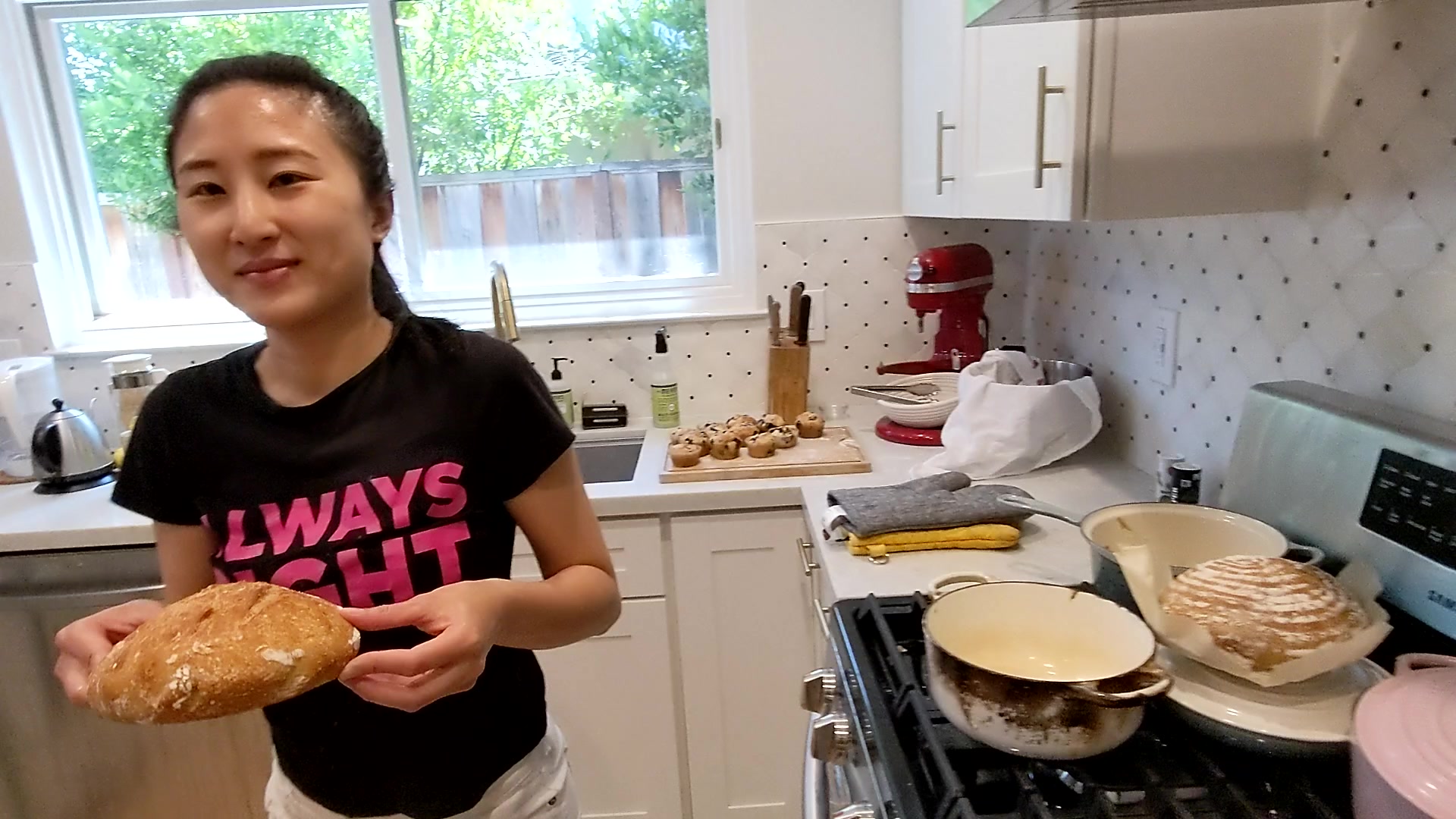}
        \\
	    \multicolumn{4}{c}{Input (perspective projection)}
        \\
        \includegraphics[width=\imagewidth]{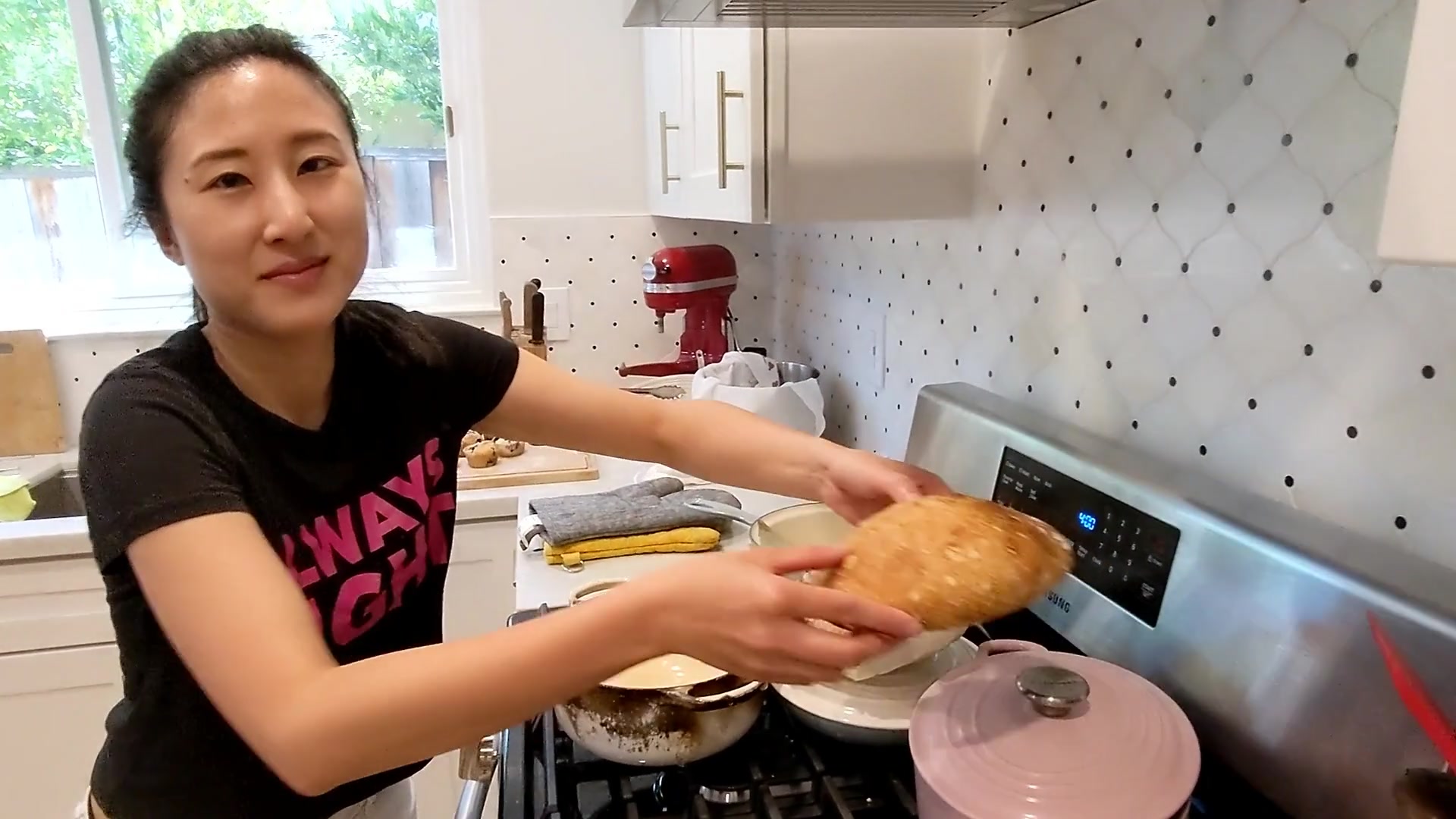} &
        \includegraphics[width=\imagewidth]{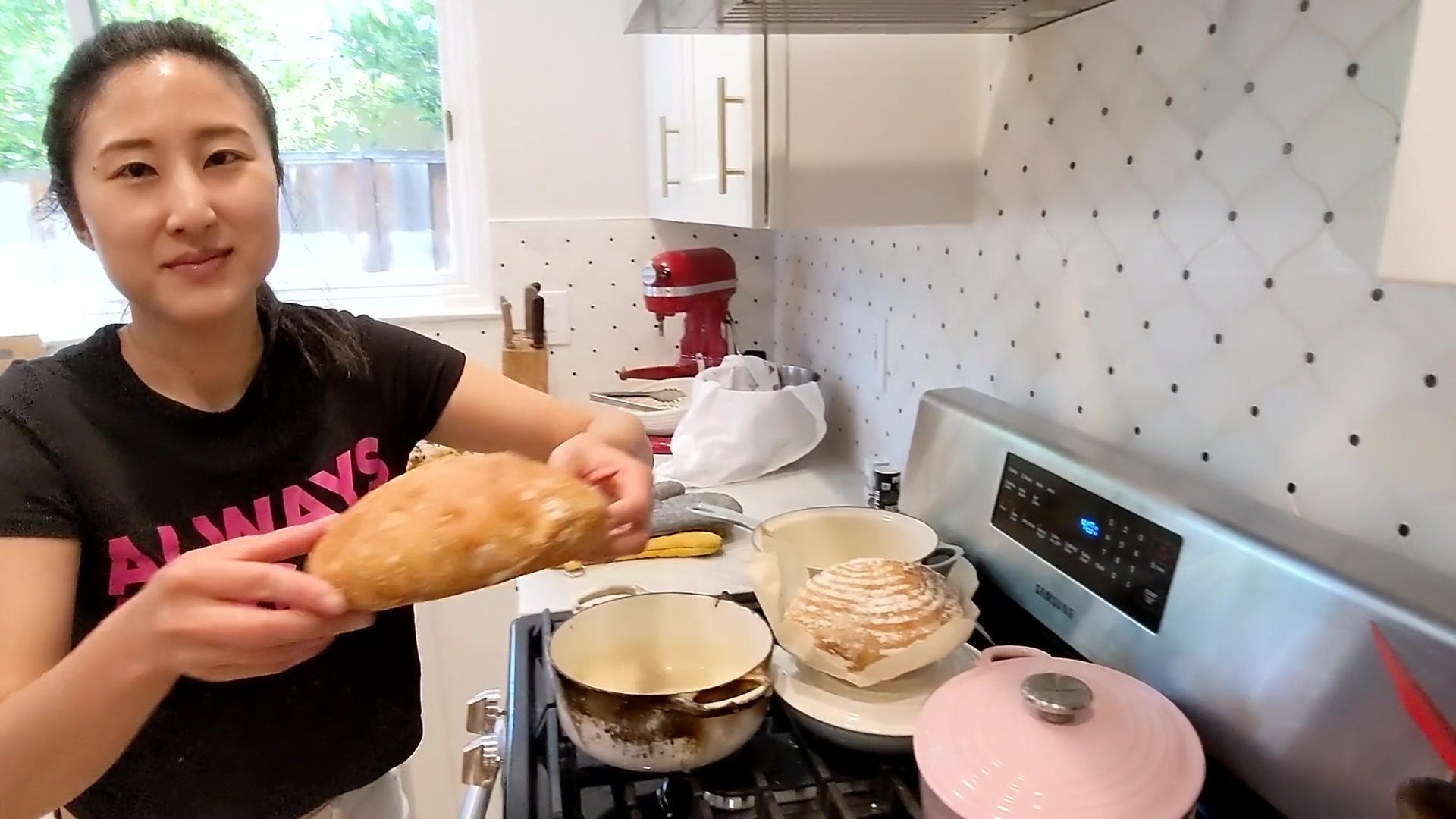} &
        \includegraphics[width=\imagewidth]{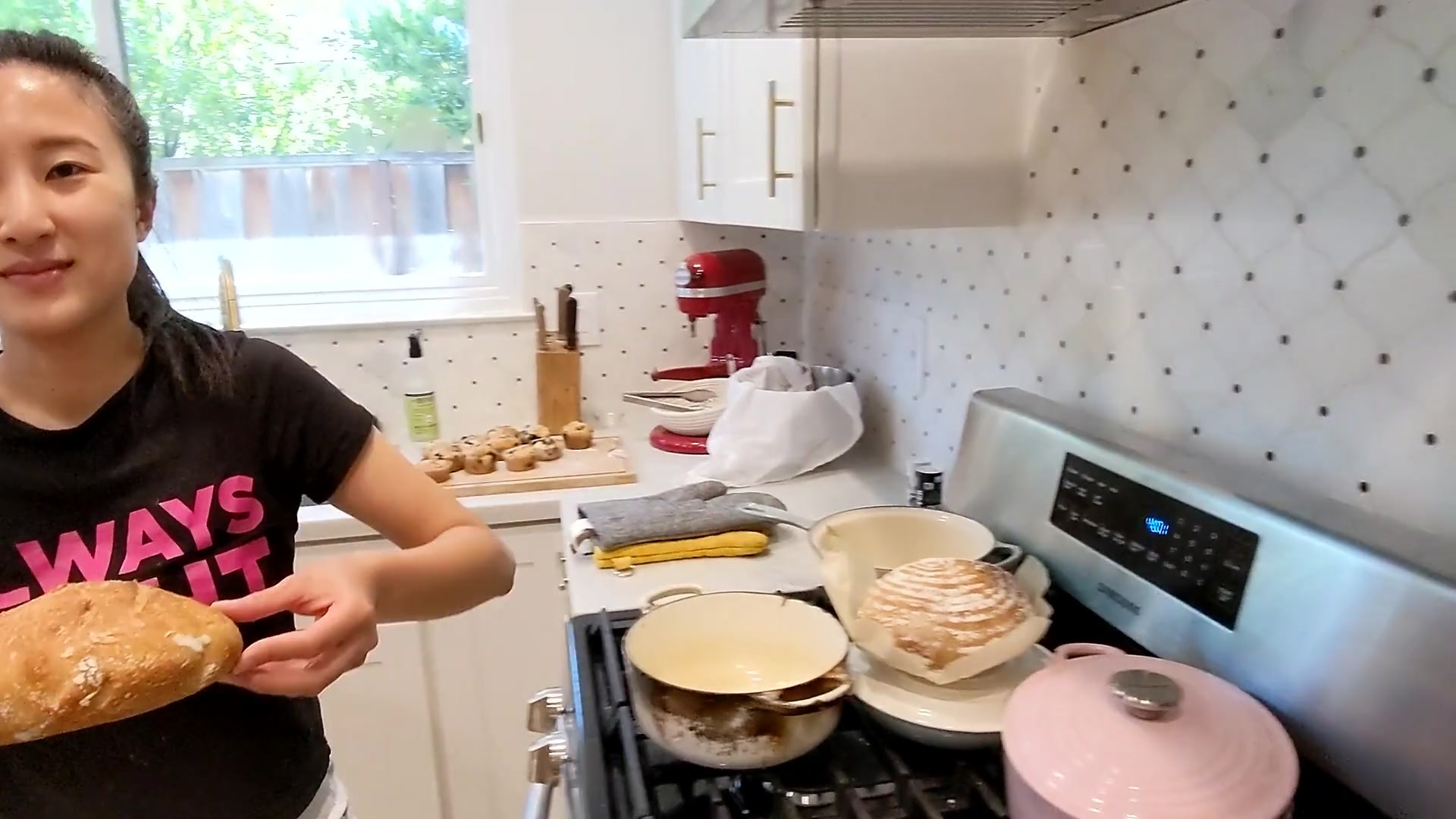} &
        \includegraphics[width=\imagewidth]{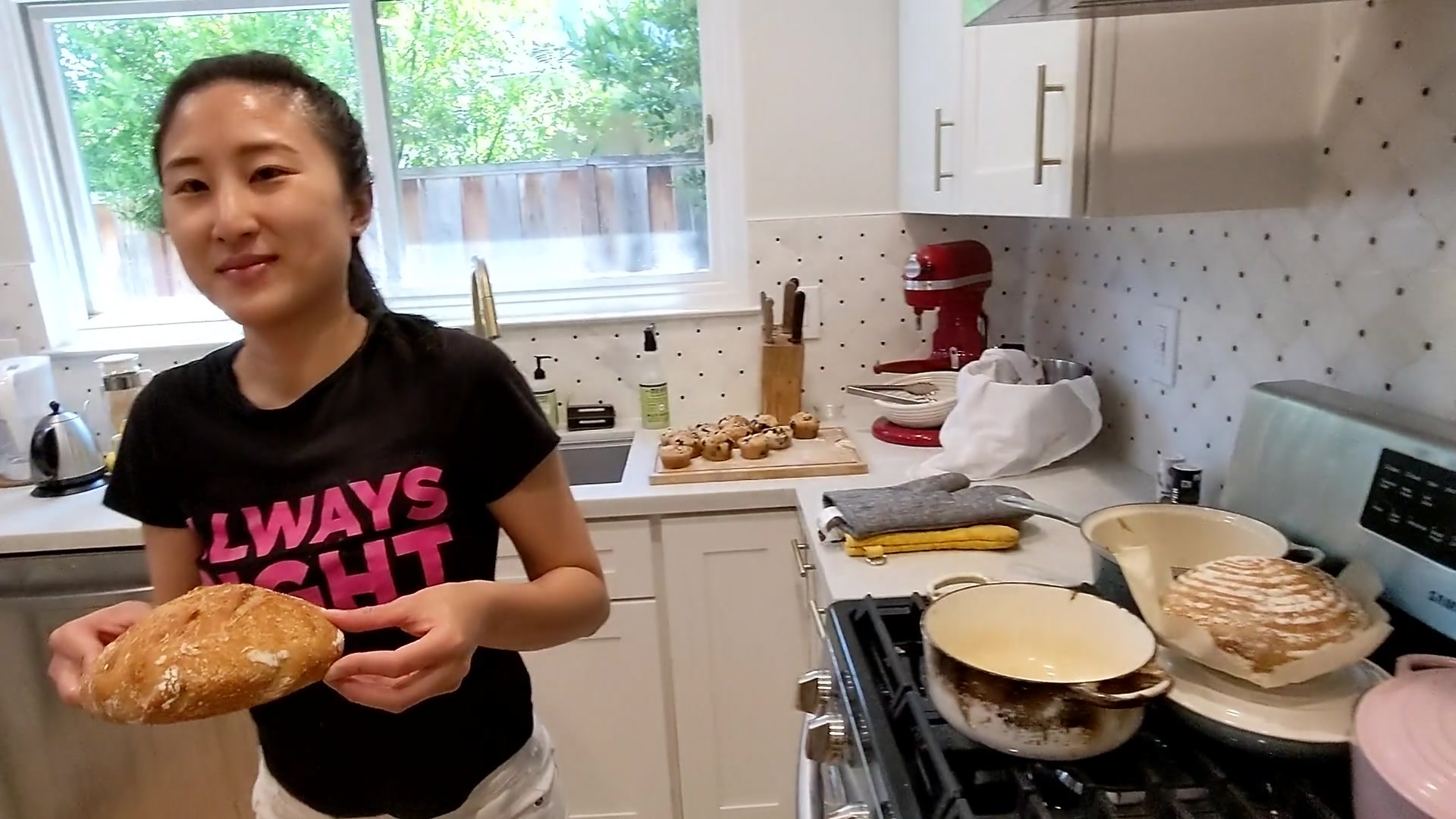} 
        \\
	    \multicolumn{4}{c}{Our method}
    \end{tabular}
    \caption{
        \textbf{Combining the electronic image stabilization (EIS) with our warp.}
        Our method works with the existing EIS on mobile phones in sequential order to produce a smooth and distortion-free video without gimbals.
    }
    \label{fig:eis}
\end{figure*}

\begin{figure*}[t!]
    \centering
    \footnotesize
    \renewcommand{\tabcolsep}{1pt} 
	\renewcommand{\arraystretch}{1.0} 
	\renewcommand{\imagewidth}{0.24\linewidth}
    \begin{tabular}{cccc}
        Input frame $n$ &
        Input frame $n + 20$ &
        Input frame $n + 40$ &
        Input frame $n + 60$
        \\
	    \begin{overpic}[width=\imagewidth,tics=10]{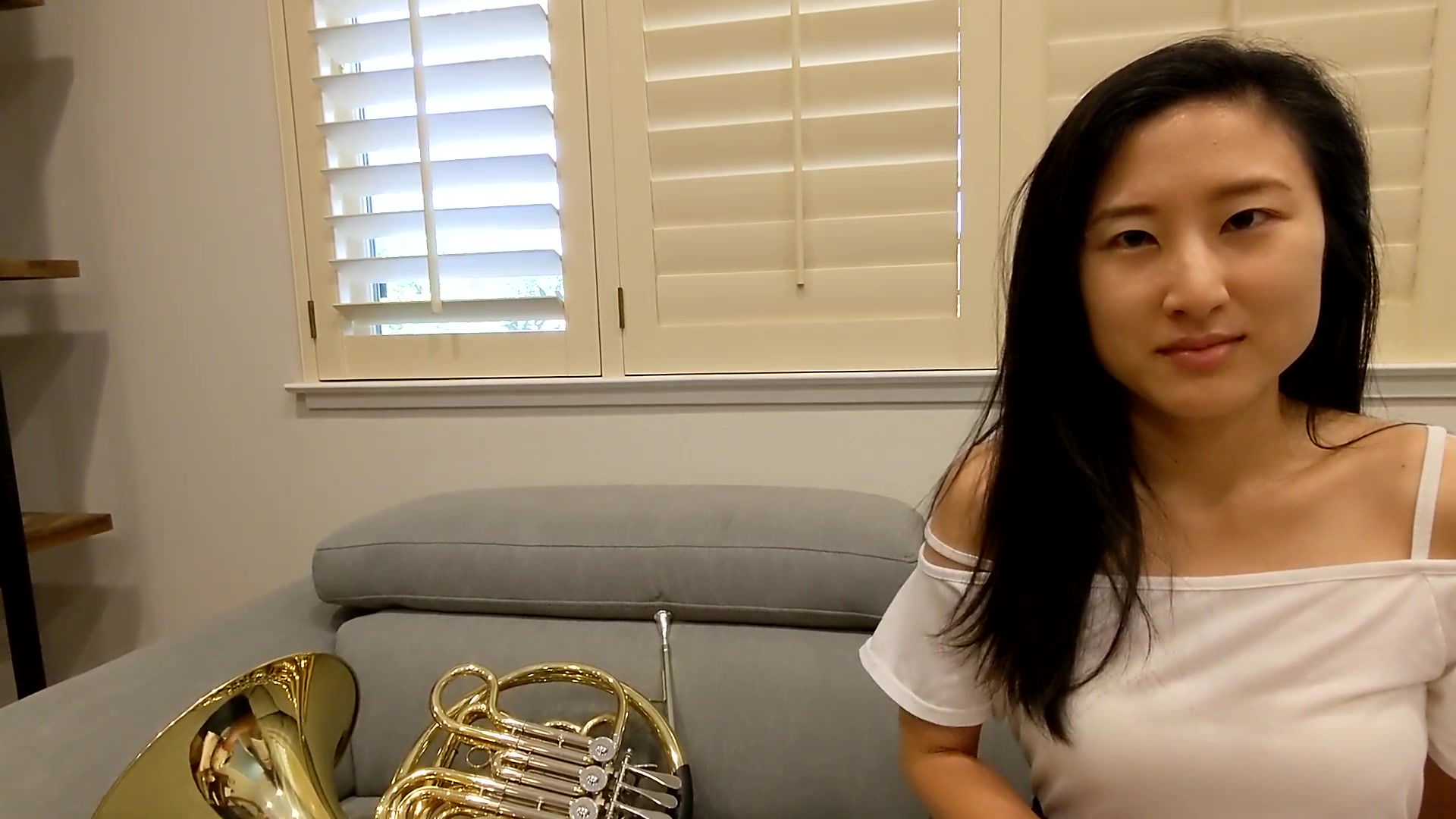}
            \put (3,3) {\white{$90\degree$ FOV}}
        \end{overpic} &
	    \begin{overpic}[width=\imagewidth,tics=10]{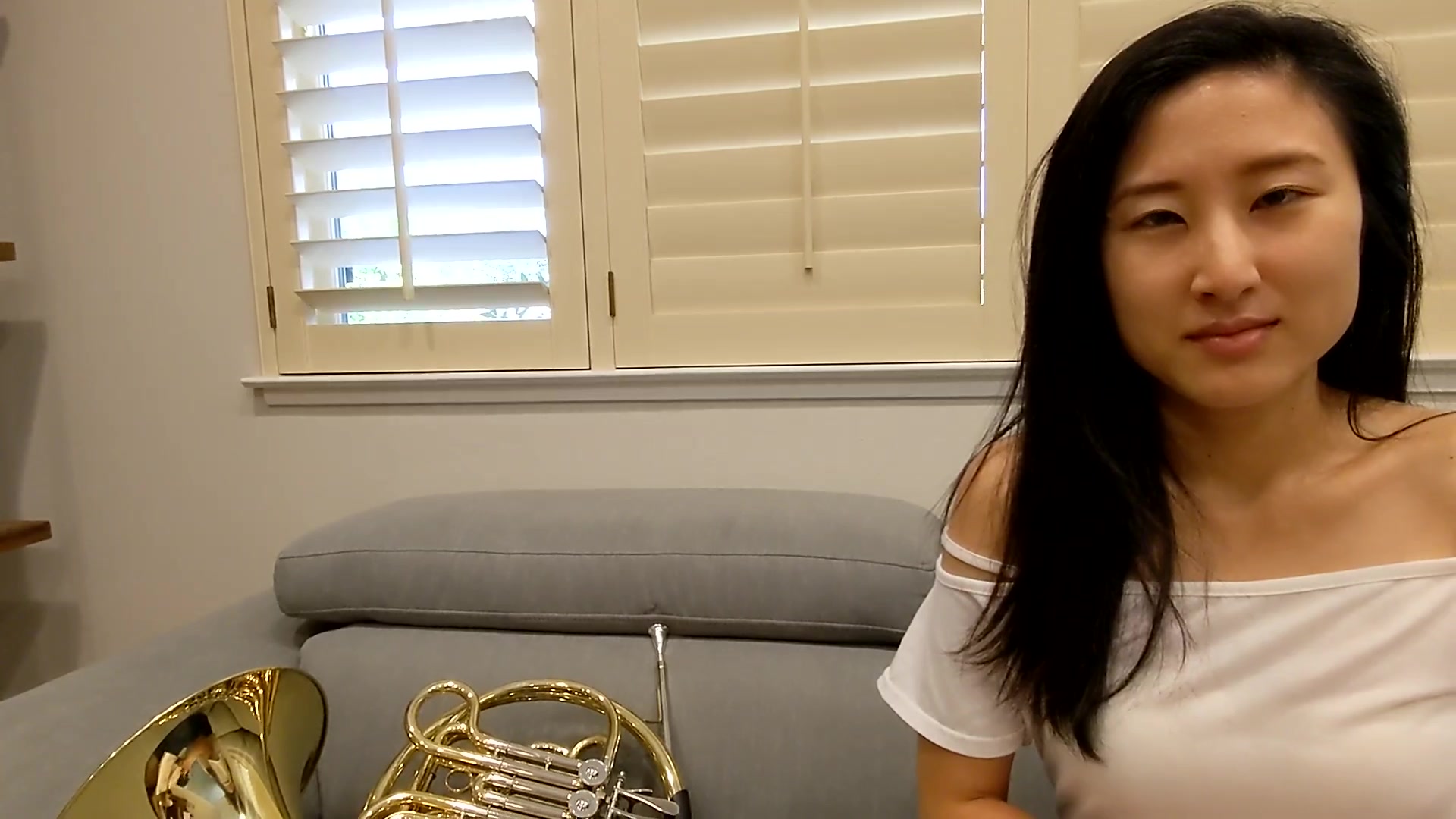}
            \put (3,3) {\white{$79\degree$ FOV}}
        \end{overpic} &
	    \begin{overpic}[width=\imagewidth,tics=10]{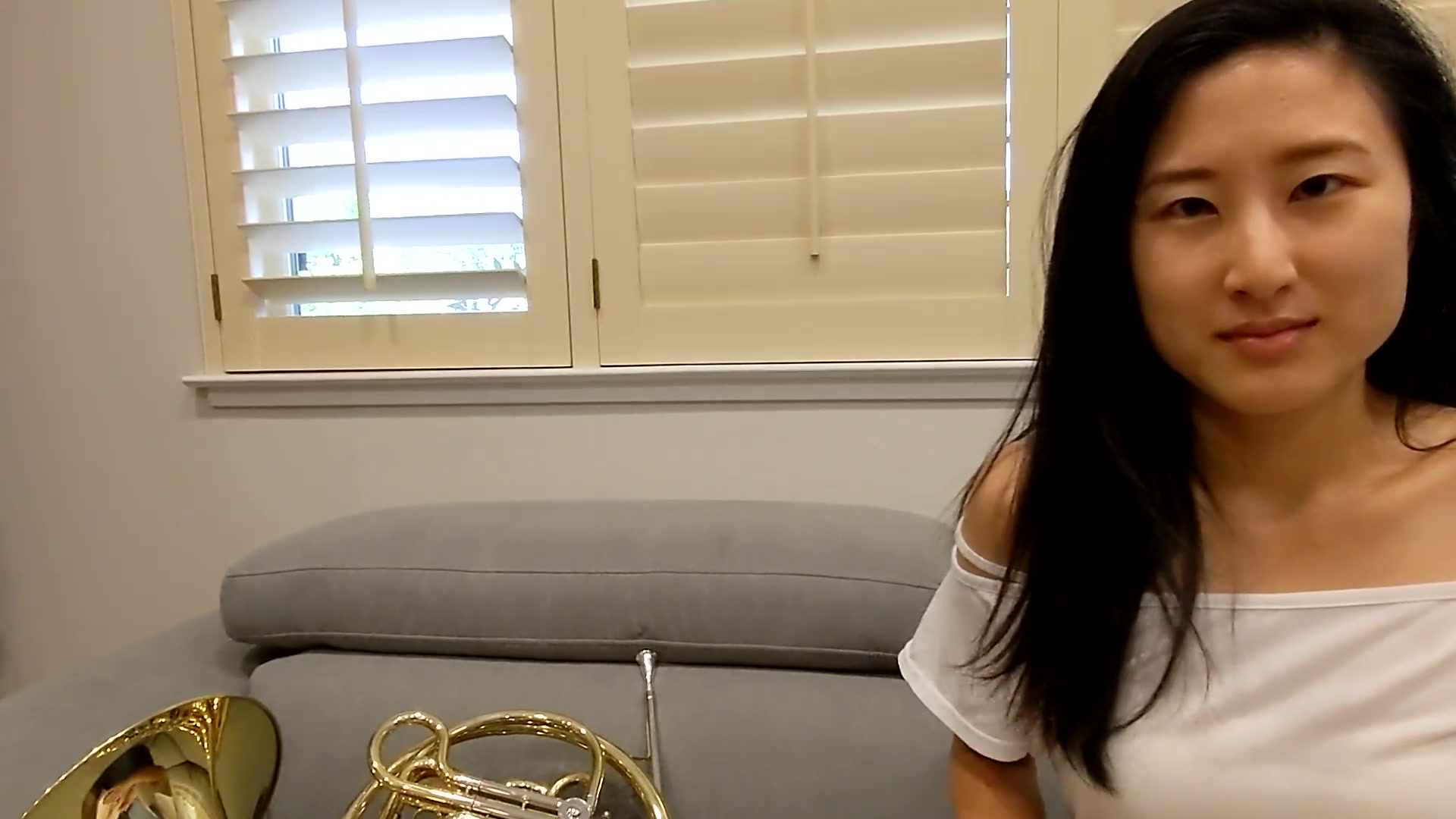}
            \put (3,3) {\white{$70\degree$ FOV}}
        \end{overpic} &
	    \begin{overpic}[width=\imagewidth,tics=10]{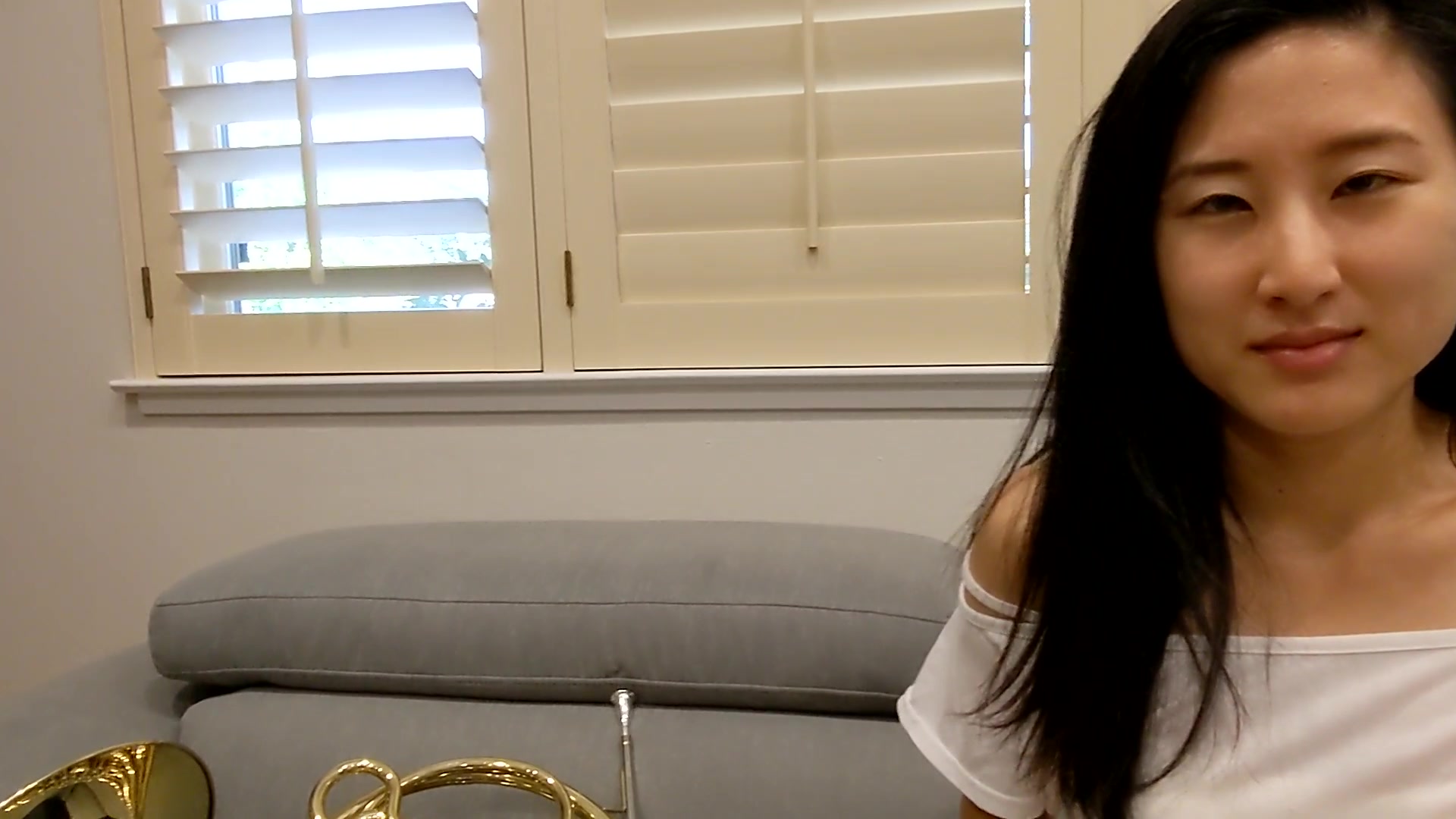}
            \put (3,3) {\white{$62\degree$ FOV}}
        \end{overpic}
        \\
	    \multicolumn{4}{c}{Input (perspective projection)}
        \\
        \includegraphics[width=\imagewidth]{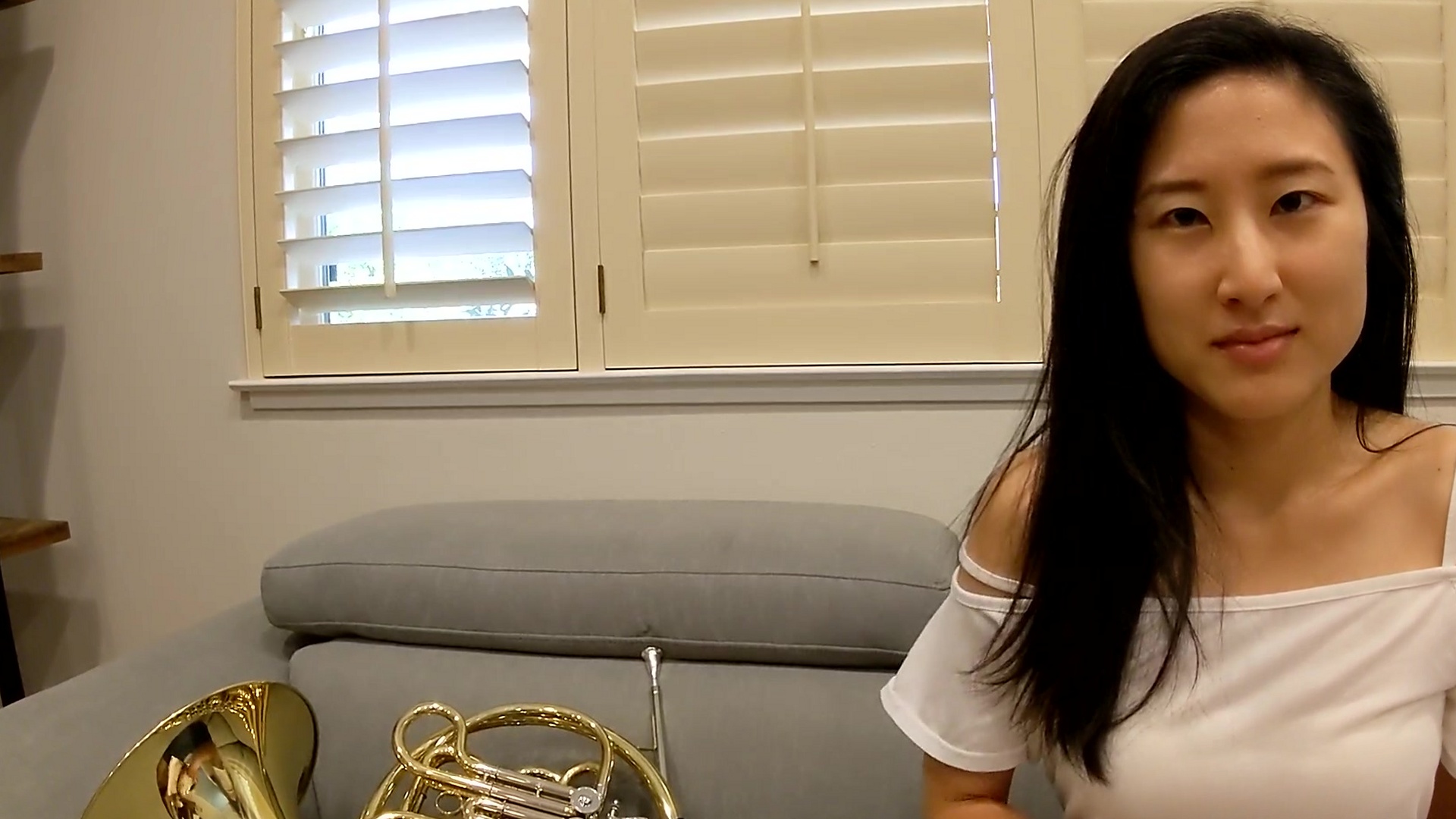} &
        \includegraphics[width=\imagewidth]{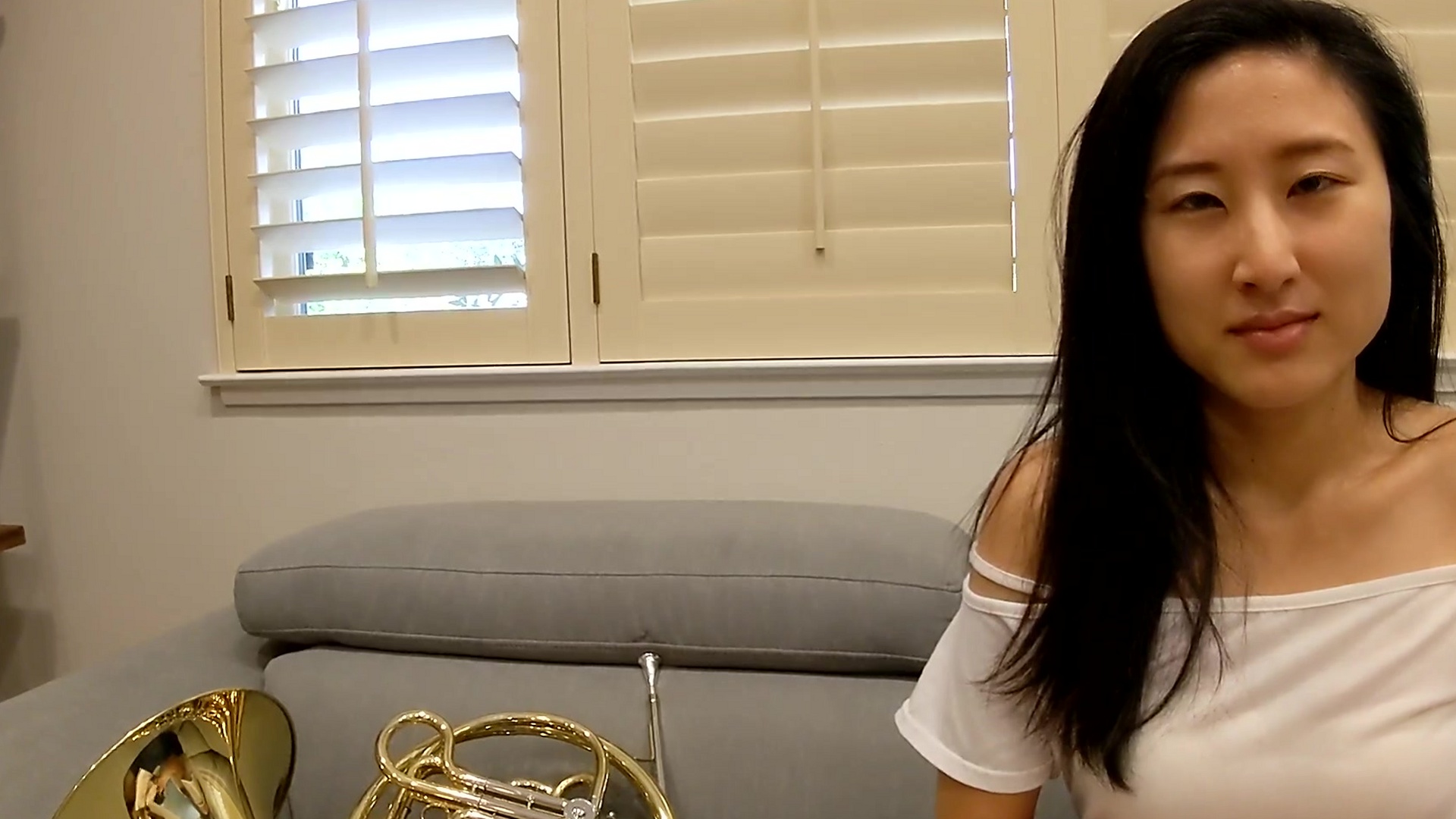} &
        \includegraphics[width=\imagewidth]{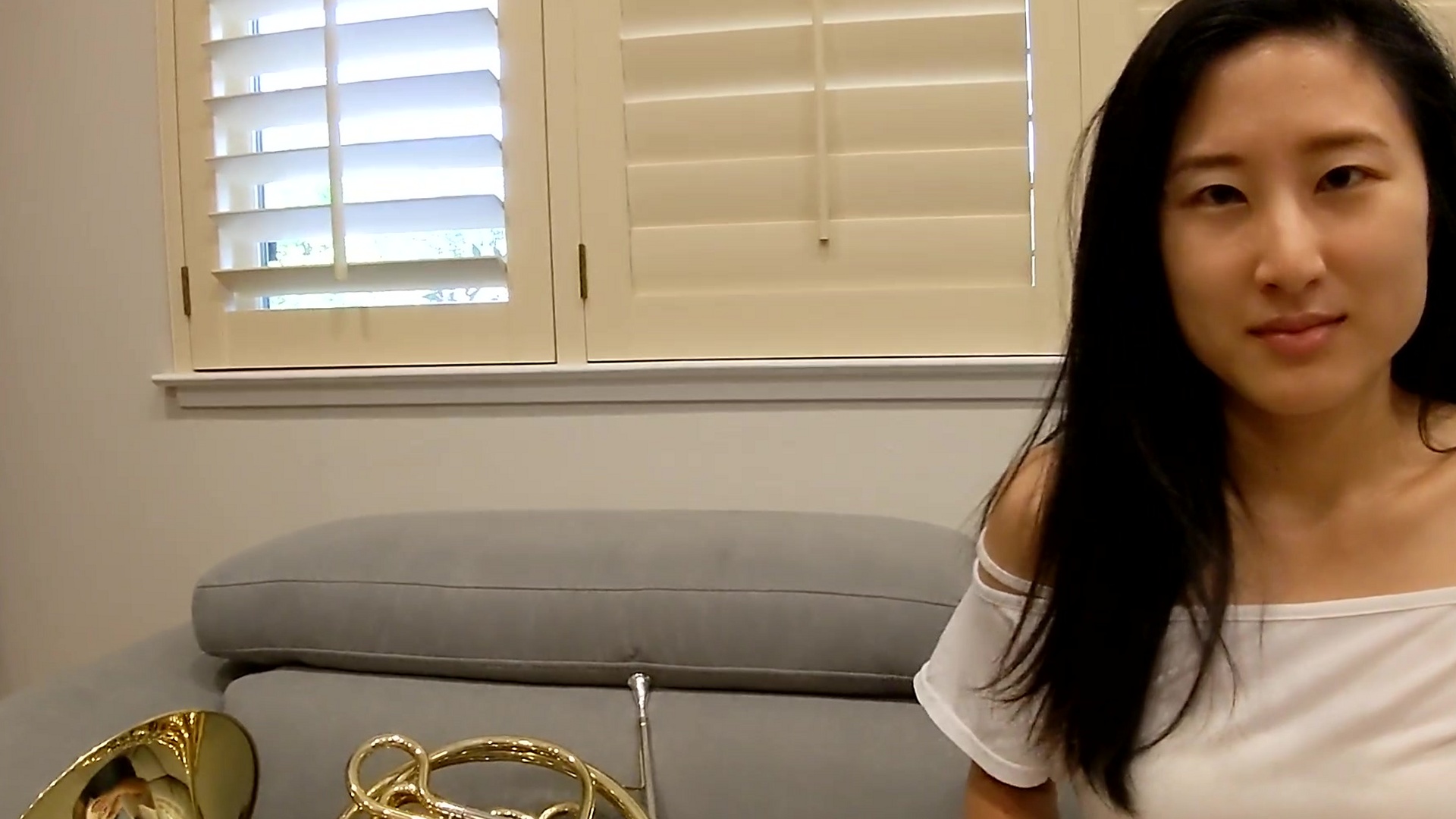} &
        \includegraphics[width=\imagewidth]{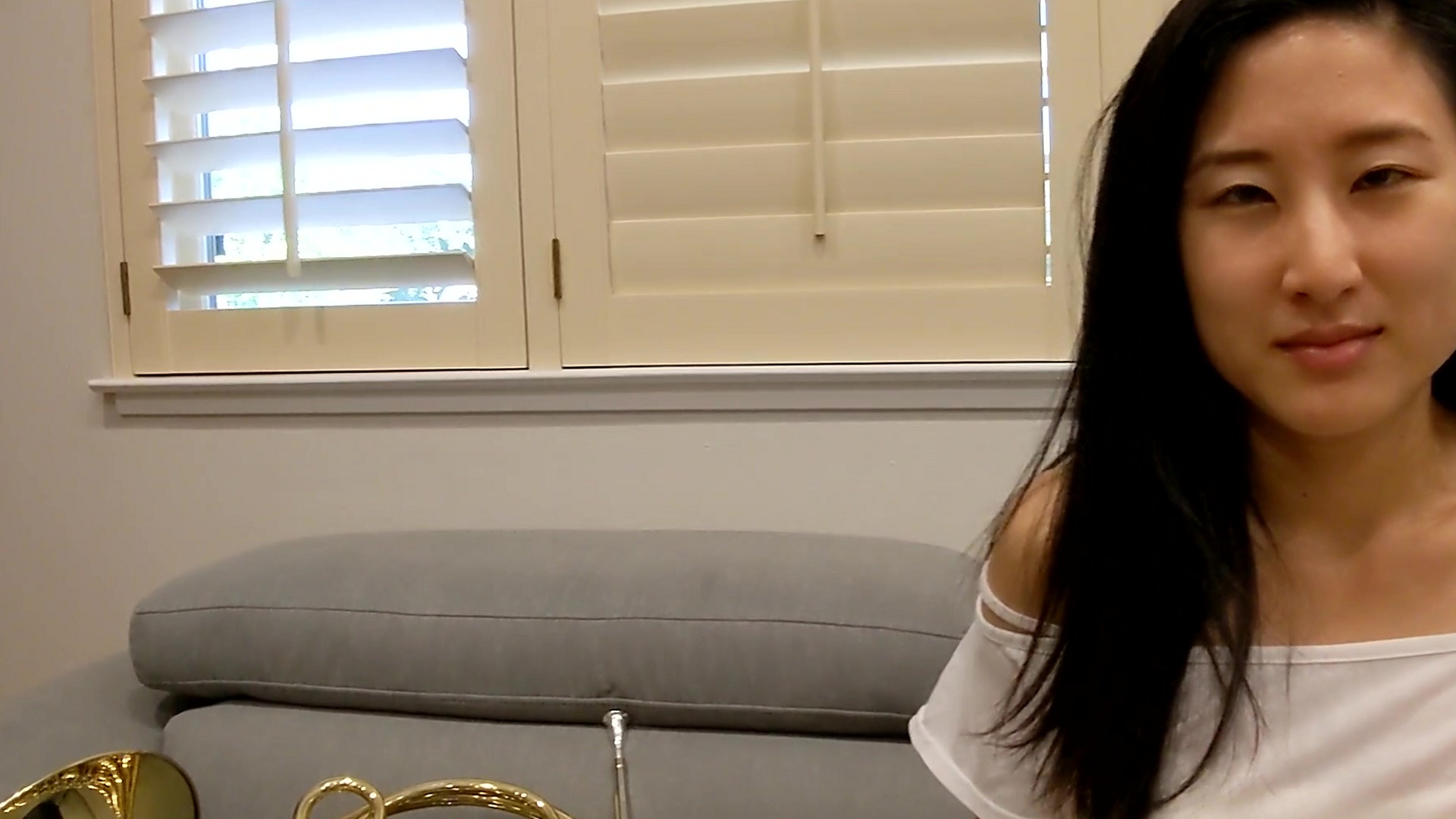} 
        \\
	    \multicolumn{4}{c}{Stereographic projection}
        \\
        \includegraphics[width=\imagewidth]{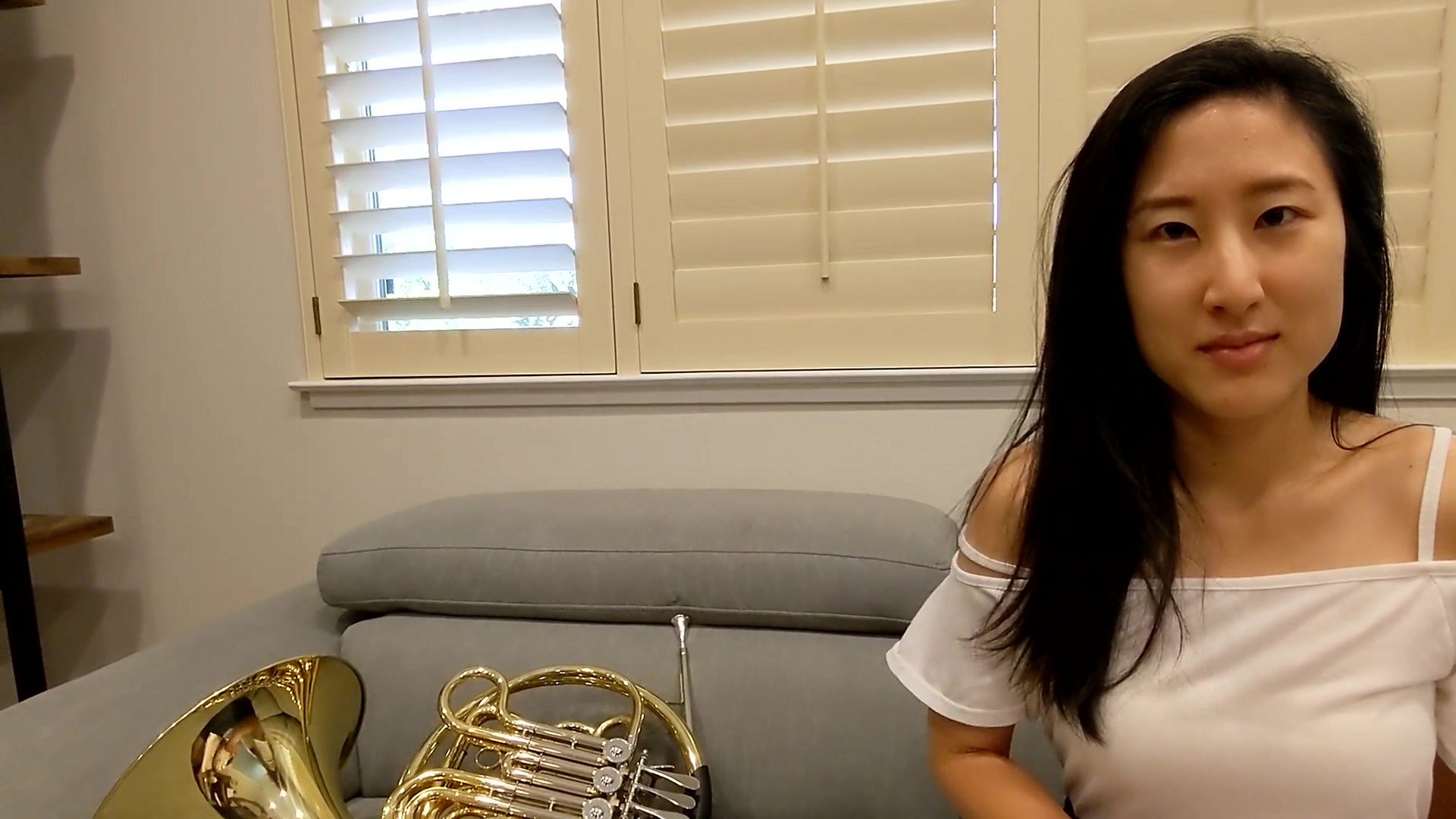} &
        \includegraphics[width=\imagewidth]{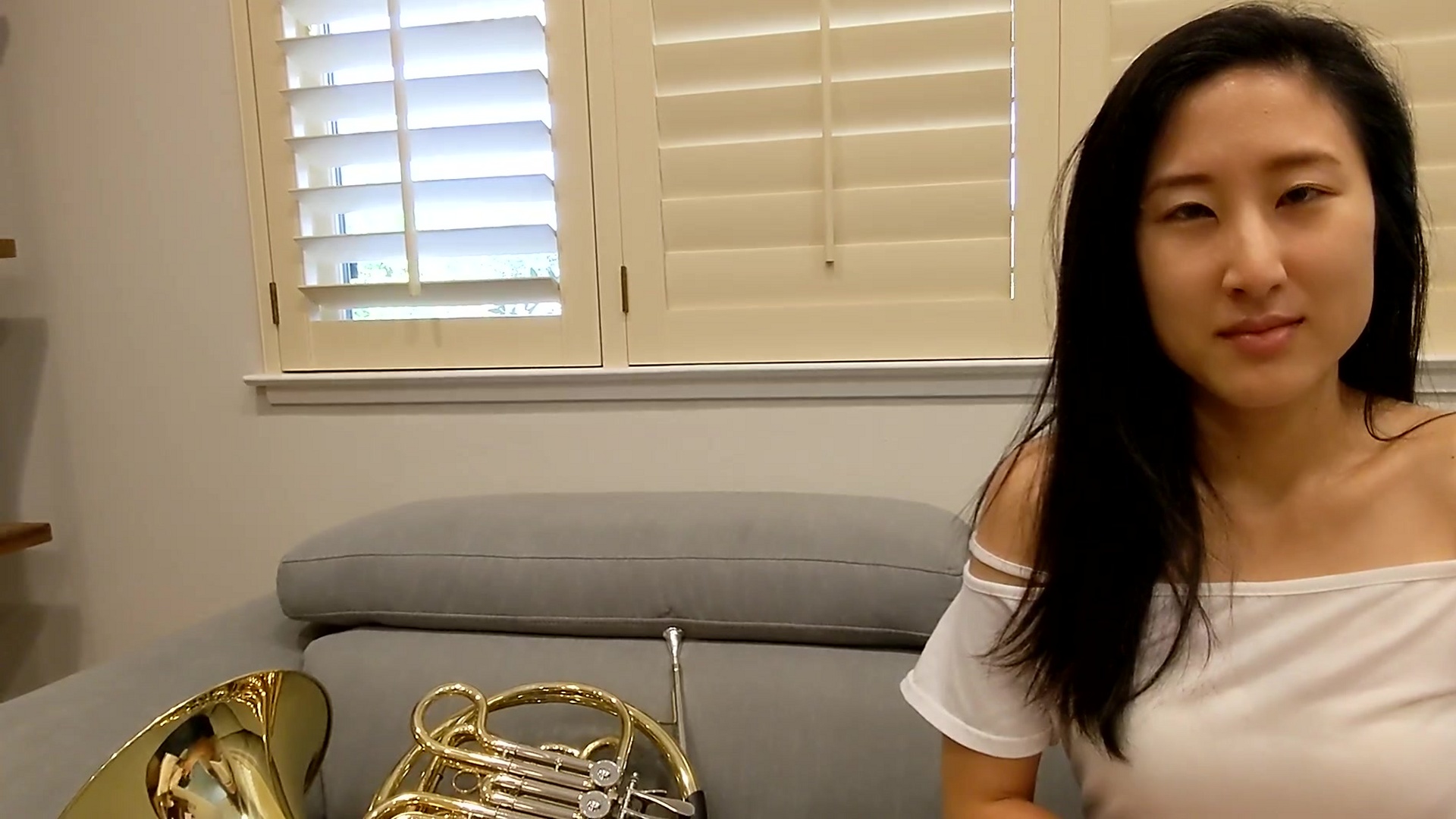} &
        \includegraphics[width=\imagewidth]{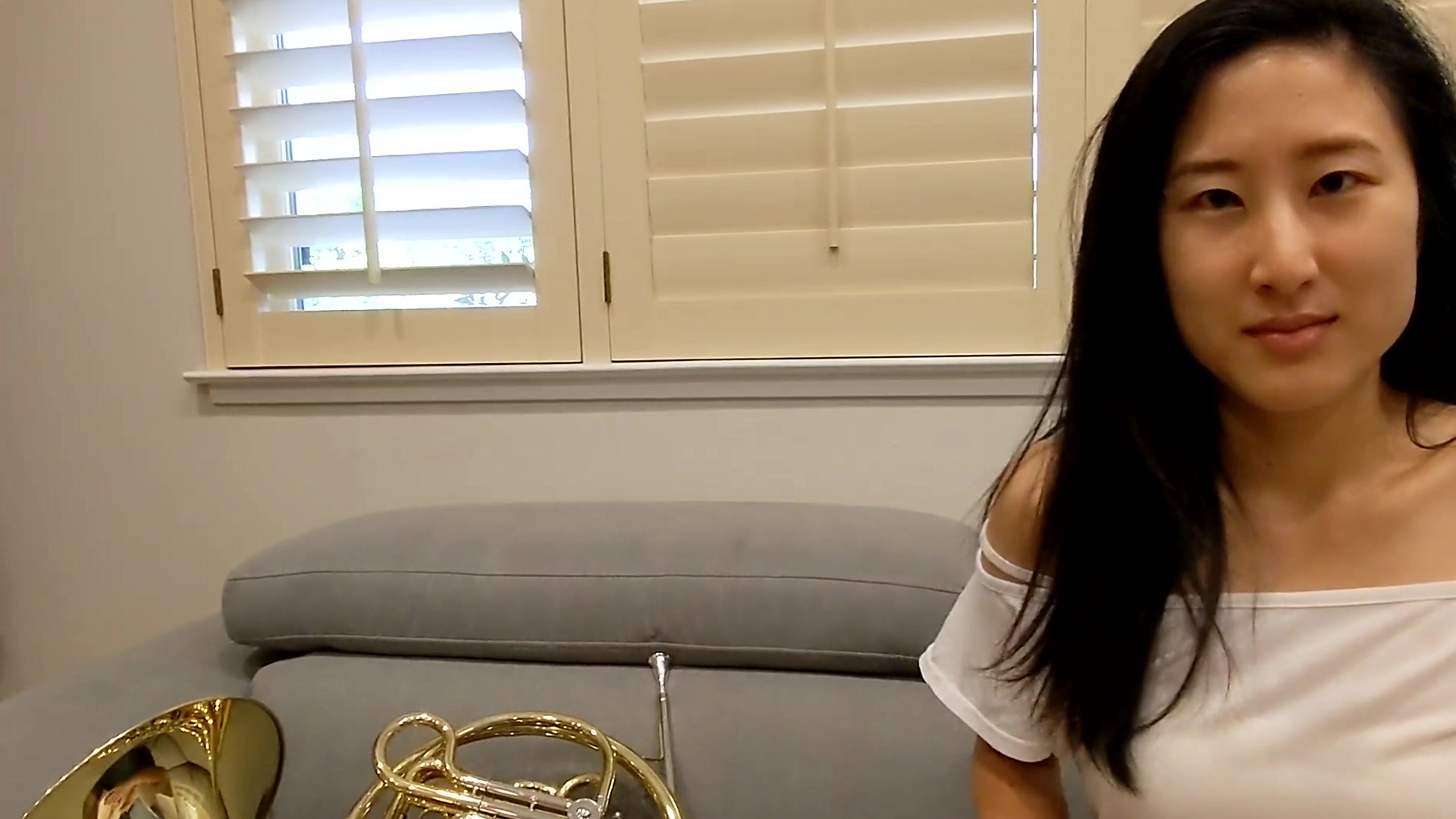} &
        \includegraphics[width=\imagewidth]{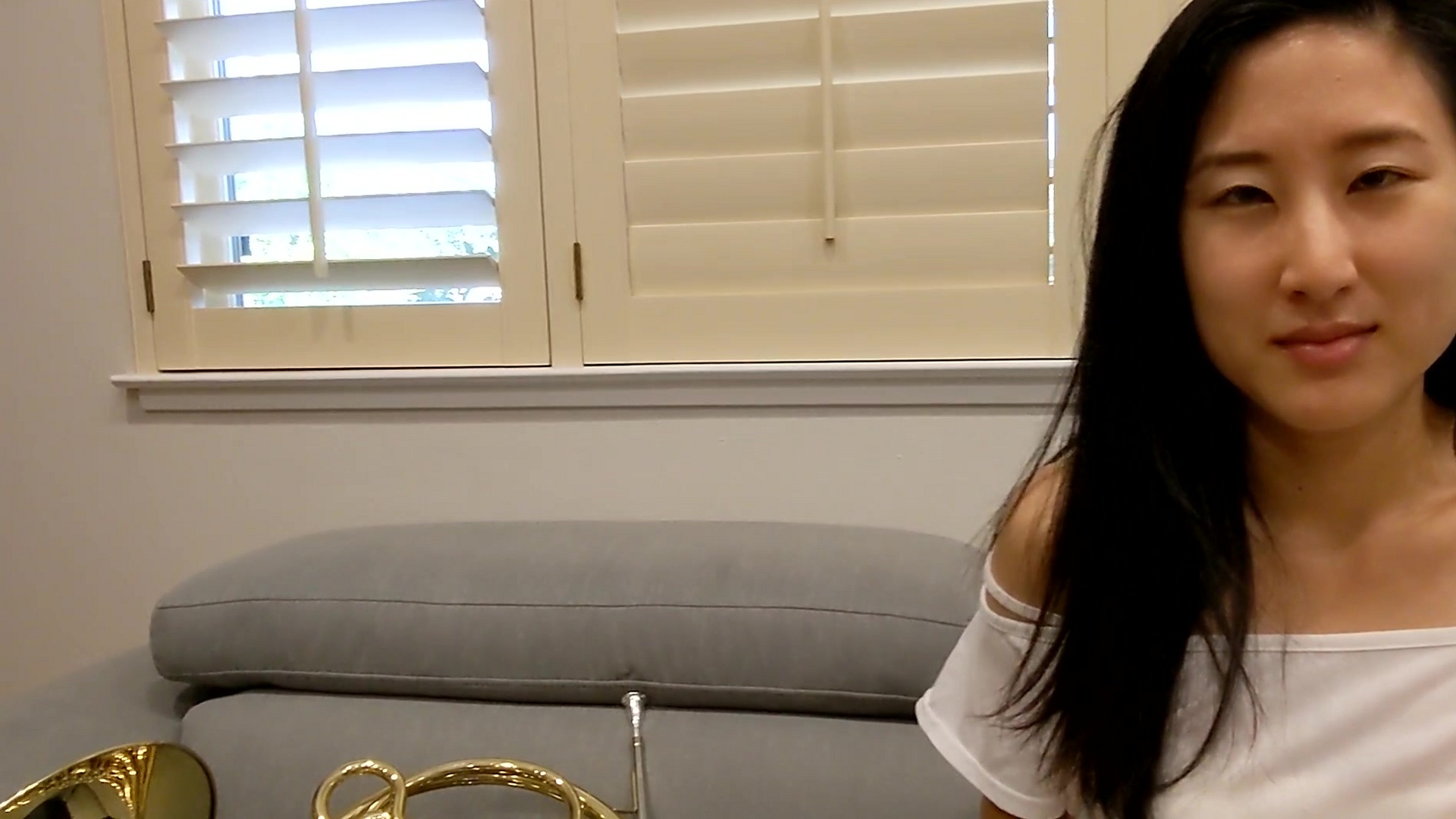} 
        \\
	    \multicolumn{4}{c}{Our method}
        \\
    \end{tabular}
    \caption{
        \textbf{Results with time-varying digital zoom specified by users.}
        The stereographic projection bends the straight lines in the background and cuts the content after cropping.
        Our method outputs temporally stable results even when the camera FOV is varying across time.
        As we generate rectangle output frames, we do not need to crop any content.
    }
    \label{fig:zoom}
\end{figure*}

\subsection{Compatibility with Electronic Image Stabilization}
In real-world scenarios, the videos captured by mobile cameras on modern phones are often recorded with the electronic image stabilization (EIS)~\cite{Shi:2019:SRT}, which applies a frame-based image warping to compensate for motion caused by shaky hands.
We note that there is no clear way to combine the EIS warp and our method in a video.
Furthermore, the camera FOV may be changed due to the digital zoom, which affects the temporal consistency of our warp.
Nevertheless, we show the results by combining the proposed correction method and the EIS warp in sequential order (as a plug-in on top of EIS) to process videos.
We capture videos using the Pixel-4 rear camera with EIS enabled, where the effective FOV is $90\degree$.
Then, we use our algorithm to correct the face distortion, as shown in~\figref{eis}.
In addition, we evaluate our algorithm on videos with varying digital zoom controlled by users, so that the FOV of the video varies with time.
By setting the stereographic projection mesh vertices in~\eqnref{face_term} based on FOV, our method can correct the face distortion precisely and generate temporally stable results.
In contrast, the frame-based method by Shih et al.~\cite{Shih:2019:DFW} contains significant temporal flickering artifacts. 
More results can be found in the supplementary video.

\subsection{Limitations}
\begin{figure}[t!]
    \centering
    \footnotesize
    \renewcommand{\tabcolsep}{1pt} 
	\renewcommand{\arraystretch}{1.0} 
	\subfloat[Input]{
	    \begin{overpic}[width=0.49\linewidth]{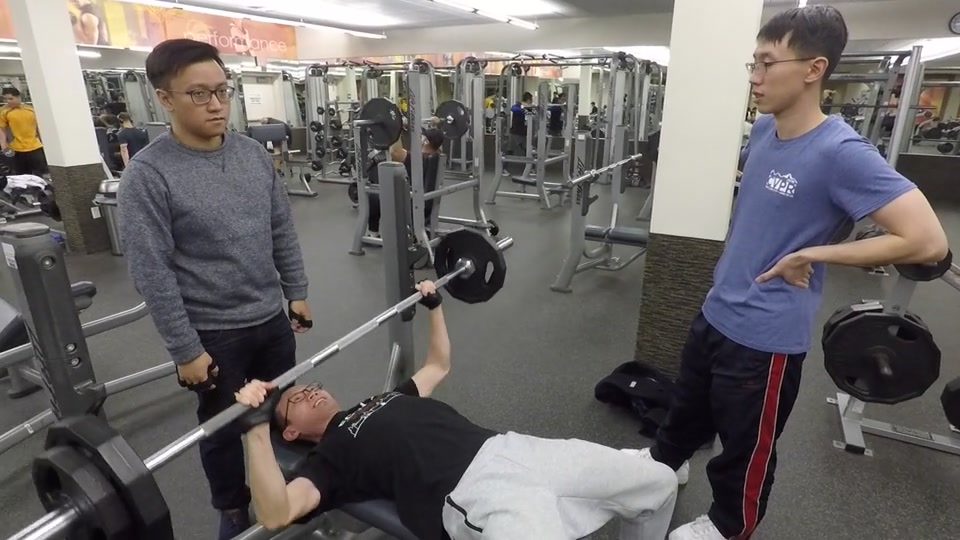}
            \put (3,3) {\white{$103\degree$ FOV}}
        \end{overpic}
	}
	\subfloat[Output]{
	    \begin{overpic}[width=0.49\linewidth]{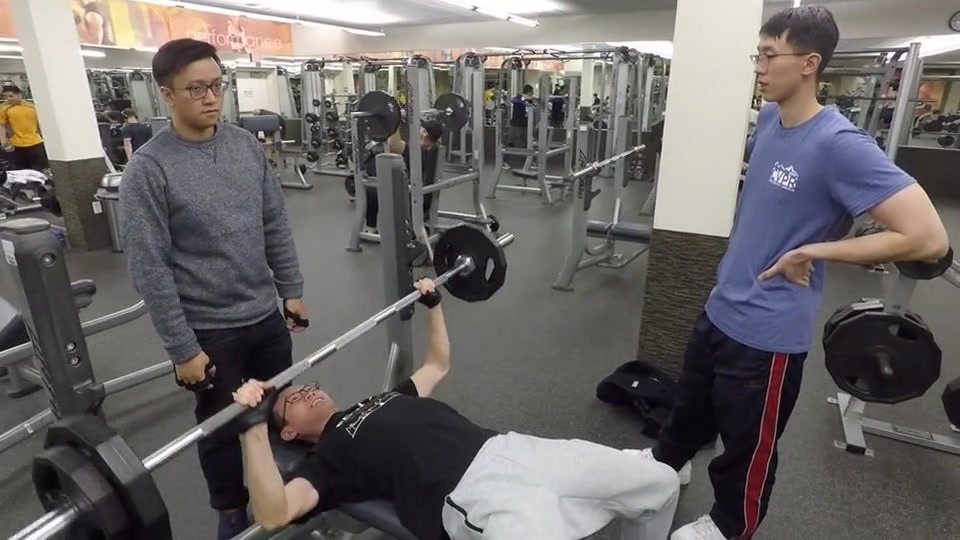}
	        \linethickness{1pt}
            \put(5,55){\color{red}\vector(1,-1){10}}
            \put(100,55){\color{red}\vector(-1,-1){10}}
        \end{overpic}
	}
    \caption{
        \textbf{Limitation.}
        Our method only corrects the face regions rather than the entire body.
        Both the left-most and right-most subjects show unnatural bodies after warping.
    }
    \label{fig:limitation-torso}
\end{figure}

\begin{figure}[t!]
    \centering
    \footnotesize
    \renewcommand{\tabcolsep}{1pt} 
	\renewcommand{\arraystretch}{0.8} 
    \subfloat[Input]{
        \begin{overpic}[width=0.49\linewidth]{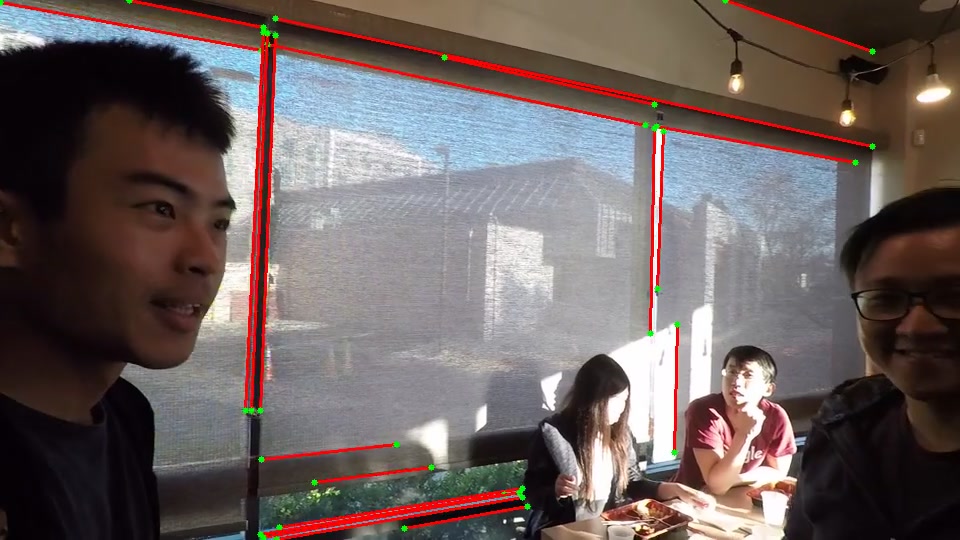}
            \put (3,3) {\white{$103\degree$ FOV}}
        \end{overpic}
    }
	\subfloat[Output]{
	    \begin{overpic}[width=0.49\linewidth]{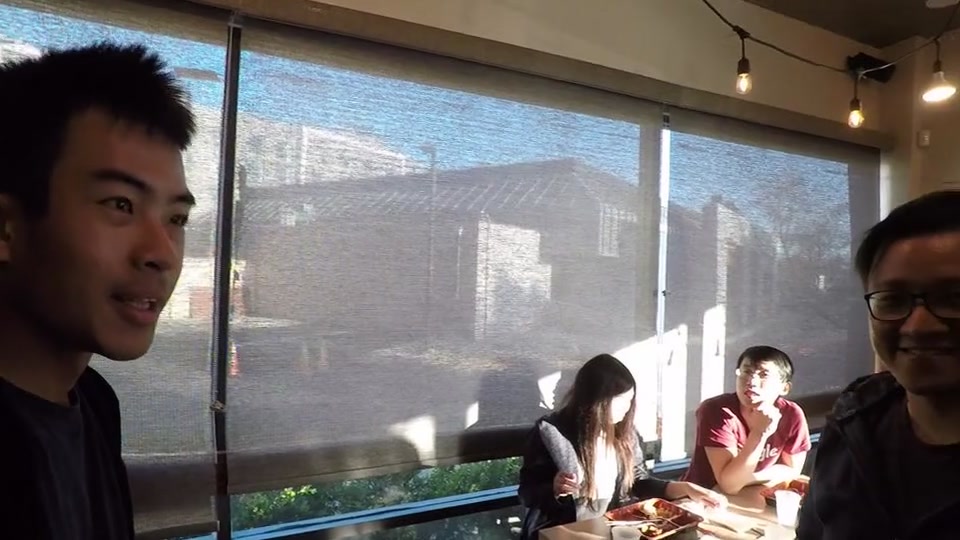}
	        \linethickness{1pt}
            \put(35,20){\color{red}\vector(-1,-0.8){10}}
        \end{overpic}
	}
    \caption{
        \textbf{Failure case.}
        When the line detector misses parts of the straight line, our method may distort the line due to inconsistent line-preservation costs.
    }
    \label{fig:limitation-line}
\end{figure}

As our method is designed to correct the facial area, 
the rendered result may look unnatural when the torso remains uncorrected, as shown in~\figref{limitation-torso}.
Nevertheless, the user study shows that our method is still preferred as faces draw higher interests. 
When a straight line is only partially detected, the undetected segment is not constrained by the line preservation energy in~\eqnref{single_line_term} and may have a different orientation than the detected part after the optimization, as shown in~\figref{limitation-line}.
For front-facing cameras, we find that some users intentionally leverage the perspective effects to make the face look thinner by tilting the camera or moving the face toward the edges and corners of the camera FOV.
In such cases, our distortion correction method serves the opposite purpose. 
Our future work will analyze the balance between the full distortion correction and the aesthetic aspect of these scenarios.

\section{Conclusions}
\label{sec:discussion}
In this work, we propose an automatic algorithm to correct wide-angle distortion on human subjects. 
To the best of our knowledge, it is the first algorithm to solve the wide-angle distortion problem for videos.
Our method employs spatial-temporal optimization for temporal consistency and adopts a line preservation term to keep the geometry in the background.
Our approach significantly improves the video quality captured by wide-angle cameras, such as sports cameras and modern phones, and is suitable for video post-editing software. 

%
While we have demonstrated that applying EIS and our warp in a sequential order produces smooth and distortion-free videos, we are interested in solving the two problems as a joint optimization to improve the undistortion and stabilization quality.
The two problems are related to each other, as the distortion correction changes the rectilinear assumption that video stabilization holds, while the stabilization alters camera FOVs used by the distortion correction algorithm.
Our method is anti-causal and requires the presence of the entire video. 
We plan to develop the preview solution for what-you-see-is-what-you-get user experiences.
 

{\small
\bibliographystyle{IEEEtran}
\bibliography{video_rectiface}

\begin{thebibliography}{10}
\providecommand{\url}[1]{#1}
\csname url@samestyle\endcsname
\providecommand{\newblock}{\relax}
\providecommand{\bibinfo}[2]{#2}
\providecommand{\BIBentrySTDinterwordspacing}{\spaceskip=0pt\relax}
\providecommand{\BIBentryALTinterwordstretchfactor}{4}
\providecommand{\BIBentryALTinterwordspacing}{\spaceskip=\fontdimen2\font plus
\BIBentryALTinterwordstretchfactor\fontdimen3\font minus
  \fontdimen4\font\relax}
\providecommand{\BIBforeignlanguage}[2]{{%
\expandafter\ifx\csname l@#1\endcsname\relax
\typeout{** WARNING: IEEEtran.bst: No hyphenation pattern has been}%
\typeout{** loaded for the language `#1'. Using the pattern for}%
\typeout{** the default language instead.}%
\else
\language=\csname l@#1\endcsname
\fi
#2}}
\providecommand{\BIBdecl}{\relax}
\BIBdecl

\bibitem{Dhanraj:2005:WPL}
D.~Vishwanath, A.~R. Girshick, and M.~S. Banks, ``Why pictures look right when
  viewed from the wrong place,'' \emph{Nature neuroscience}, vol.~8, no.~10, p.
  1401, 2005.

\bibitem{Zorin:1995:CGP}
D.~Zorin and A.~H. Barr, ``Correction of geometric perceptual distortions in
  pictures,'' in \emph{ACM SIGGRAPH}, 1995.

\bibitem{Carroll:2009:OCP}
R.~Carroll, M.~Agrawal, and A.~Agarwala, ``Optimizing content-preserving
  projections for wide-angle images,'' \emph{ACM TOG}, vol.~28, no.~3, p.~43,
  2009.

\bibitem{Shih:2019:DFW}
Y.~Shih, W.-S. Lai, and C.-K. Liang, ``Distortion-free wide-angle portraits on
  camera phones,'' \emph{ACM TOG}, vol.~38, no.~4, pp. 61:1--61:12, 2019.

\bibitem{Niu2010:WPF}
Y.~Niu, F.~Liu, X.~Li, and M.~Gleicher, ``Warp propagation for video
  resizing,'' in \emph{CVPR}, 2010.

\bibitem{Goferman:2011:CAS}
S.~Goferman, L.~Zelnik-Manor, and A.~Tal, ``Context-aware saliency detection,''
  \emph{TPAMI}, vol.~34, no.~10, pp. 1915--1926, 2011.

\bibitem{Judd:2009:LTP}
T.~Judd, K.~Ehinger, F.~Durand, and A.~Torralba, ``Learning to predict where
  humans look,'' in \emph{ICCV}, 2009.

\bibitem{Pan:2016:SAD}
J.~Pan, E.~Sayrol, X.~Giro-i Nieto, K.~McGuinness, and N.~E. O'Connor,
  ``Shallow and deep convolutional networks for saliency prediction,'' in
  \emph{CVPR}, 2016.

\bibitem{Krahenbuhl:2009:ASF}
P.~Kr{\"a}henb{\"u}hl, M.~Lang, A.~Hornung, and M.~Gross, ``A system for
  retargeting of streaming video,'' \emph{ACM TOG}, vol.~28, no.~5, p. 126,
  2009.

\bibitem{Lin:2013:CAV}
S.-S. Lin, C.-H. Lin, I.-C. Yeh, S.-H. Chang, C.-K. Yeh, and T.-Y. Lee,
  ``Content-aware video retargeting using object-preserving warping,''
  \emph{TVCG}, vol.~19, no.~10, pp. 1677--1686, 2013.

\bibitem{Wang:2009:MAT}
Y.-S. Wang, H.~Fu, O.~Sorkine, T.-Y. Lee, and H.-P. Seidel, ``Motion-aware
  temporal coherence for video resizing,'' \emph{ACM TOG}, vol.~28, no.~5, pp.
  127:1--127:10, 2009.

\bibitem{Wang:2011:SAC}
Y.-S. Wang, J.-H. Hsiao, O.~Sorkine, and T.-Y. Lee, ``Scalable and coherent
  video resizing with per-frame optimization,'' \emph{ACM TOG}, vol.~30, no.~4,
  p.~88, 2011.

\bibitem{Wang:2010:MBV}
Y.-S. Wang, H.-C. Lin, O.~Sorkine, and T.-Y. Lee, ``Motion-based video
  retargeting with optimized crop-and-warp,'' \emph{ACM TOG}, vol.~29, no.~4,
  p.~90, 2010.

\bibitem{Wolf:2007:NHC}
L.~Wolf, M.~Guttmann, and D.~Cohen-Or, ``Non-homogeneous content-driven
  video-retargeting,'' in \emph{ICCV}, 2007.

\bibitem{Grundmann:2011:ADV}
M.~Grundmann, V.~Kwatra, and I.~Essa, ``Auto-directed video stabilization with
  robust l1 optimal camera paths,'' in \emph{CVPR}, 2011.

\bibitem{Liu:2009:CPW}
F.~Liu, M.~Gleicher, H.~Jin, and A.~Agarwala, ``Content-preserving warps for 3d
  video stabilization,'' \emph{ACM TOG}, vol.~28, no.~3, p.~44, 2009.

\bibitem{Liu:2013:BCP}
S.~Liu, L.~Yuan, P.~Tan, and J.~Sun, ``Bundled camera paths for video
  stabilization,'' \emph{ACM TOG}, vol.~32, no.~4, p.~78, 2013.

\bibitem{Grundmann:2012:CFR}
M.~Grundmann, V.~Kwatra, D.~Castro, and I.~Essa, ``Calibration-free rolling
  shutter removal,'' in \emph{ICCP}, 2012.

\bibitem{Karpenko:2011:DVS}
A.~Karpenko, D.~Jacobs, J.~Baek, and M.~Levoy, ``Digital video stabilization
  and rolling shutter correction using gyroscopes,'' \emph{Stanford University
  Computer Science Tech Report}, vol.~1, p.~2, 2011.

\bibitem{Jiang:2015:VSW}
W.~Jiang and J.~Gu, ``Video stitching with spatial-temporal content-preserving
  warping,'' in \emph{CVPR Workshops}, 2015.

\bibitem{Nie:2017:DVS}
Y.~Nie, T.~Su, Z.~Zhang, H.~Sun, and G.~Li, ``Dynamic video stitching via
  shakiness removing,'' \emph{TIP}, vol.~27, no.~1, pp. 164--178, 2017.

\bibitem{Li:2018:DAS}
B.~Li, C.-W. Lin, B.~Shi, T.~Huang, W.~Gao, and C.-C. Jay~Kuo, ``Depth-aware
  stereo video retargeting,'' in \emph{CVPR}, 2018.

\bibitem{Liu:2015:ARM}
Y.~Liu, L.~Sun, and S.~Yang, ``A retargeting method for stereoscopic 3d
  video,'' \emph{Computational Visual Media}, vol.~1, no.~2, pp. 119--127,
  2015.

\bibitem{Wei:2012:FVC}
J.~Wei, C.-F. Li, S.-M. Hu, R.~R. Martin, and C.-L. Tai, ``Fisheye video
  correction,'' \emph{TVCG}, vol.~18, no.~10, pp. 1771--1783, 2012.

\bibitem{Shi:2019:SRT}
F.~Shi, S.-F. Tsai, Y.~Wang, and C.-K. Liang, ``Steadiface: Real-time
  face-centric stabilization on mobile phones,'' in \emph{ICIP}, 2019.

\bibitem{Yu:2018:SVS}
J.~Yu and R.~Ramamoorthi, ``Selfie video stabilization,'' in \emph{ECCV}, 2018.

\bibitem{Wadhwa:2018:SDW}
N.~Wadhwa, R.~Garg, D.~Jacobs, B.~Feldman, N.~Kanazawa, R.~Carroll,
  Y.~Movshovitz-Attias, J.~Barron, Y.~Pritch, and M.~Levoy, ``Synthetic
  depth-of-field with a single-camera mobile phone,'' \emph{ACM TOG}, vol.~37,
  no.~4, pp. 64:1--64:13, 2018.

\bibitem{Chang:2011:CAD}
C.-H. Chang, C.-K. Liang, and Y.-Y. Chuang, ``Content-aware display adaptation
  and interactive editing for stereoscopic images,'' \emph{TMM}, vol.~13,
  no.~4, pp. 589--601, 2011.

\bibitem{Wang:2008:OSS}
Y.-S. Wang, C.-L. Tai, O.~Sorkine, and T.-Y. Lee, ``Optimized scale-and-stretch
  for image resizing,'' \emph{ACM TOG}, vol.~27, no.~5, p. 118, 2008.

\bibitem{Liao:2014:SAV}
J.~Liao, R.~S. Lima, D.~Nehab, H.~Hoppe, and P.~V. Sander, ``Semi-automated
  video morphing,'' \emph{CGF}, vol.~33, no.~4, pp. 51--60, 2014.

\bibitem{Valente:2015:PDM}
J.~Valente and S.~Soatto, ``Perspective distortion modeling, learning and
  compensation,'' in \emph{CVPR}, 2015.

\bibitem{fried2016perspective}
O.~Fried, E.~Shechtman, D.~B. Goldman, and A.~Finkelstein, ``Perspective-aware
  manipulation of portrait photos,'' \emph{ACM TOG}, vol.~35, no.~4, pp. 1--10,
  2016.

\bibitem{nagano2019deep}
K.~Nagano, H.~Luo, Z.~Wang, J.~Seo, J.~Xing, L.~Hu, L.~Wei, and H.~Li, ``Deep
  face normalization,'' \emph{ACM TOG}, vol.~38, no.~6, 2019.

\bibitem{zhao2019learning}
Y.~Zhao, Z.~Huang, T.~Li, W.~Chen, C.~LeGendre, X.~Ren, A.~Shapiro, and H.~Li,
  ``Learning perspective undistortion of portraits,'' in \emph{CVPR}, 2019.

\bibitem{Hugemann:2010:CLD}
W.~Hugemann, ``Correcting lens distortions in digital photographs,''
  \emph{Ingenieurb{\"u}ro Morawski+ Hugemann: Leverkusen, Germany}, p.~12,
  2010.

\bibitem{Li:2019:DRI}
X.~Li, B.~Zhang, J.~Liao, and P.~V. Sander, ``Document rectification and
  illumination correction using a patch-based cnn,'' \emph{ACM TOG}, vol.~38,
  no.~6, 2019.

\bibitem{Markovitz:2020:CYR}
A.~Markovitz, I.~Lavi, O.~Perel, S.~Mazor, and R.~Litman, ``Can you read me
  now? content aware rectification using angle supervision,'' 2020.

\bibitem{Tehrani:2016:CPP}
M.~A. Tehrani, A.~Majumder, and M.~Gopi, ``Correcting perceived perspective
  distortions using object specific planar transformations,'' in \emph{ICCP},
  2016.

\bibitem{Bousaid:2020:PDM}
A.~Bousaid, T.~Theodoridis, S.~Nefti-Meziani, and S.~Davis, ``Perspective
  distortion modeling for image measurements,'' \emph{IEEE Access}, vol.~8, pp.
  15\,322--15\,331, 2020.

\bibitem{Yin:2018:FAM}
X.~Yin, X.~Wang, J.~Yu, M.~Zhang, P.~Fua, and D.~Tao, ``Fisheyerecnet: A
  multi-context collaborative deep network for fisheye image rectification,''
  in \emph{ECCV}, 2018.

\bibitem{Del:2020:BFO}
N.~P. Del~Gallego, J.~Ilao, and M.~Cordel, ``Blind first-order perspective
  distortion correction using parallel convolutional neural networks,''
  \emph{Sensors}, vol.~20, no.~17, p. 4898, 2020.

\bibitem{Carroll:2010:IWF}
R.~Carroll, A.~Agarwala, and M.~Agrawala, ``Image warps for artistic
  perspective manipulation,'' \emph{ACM TOG}, vol.~29, no.~4, p. 127, 2010.

\bibitem{Von:2010:LSD}
R.~G. Von~Gioi, J.~Jakubowicz, J.-M. Morel, and G.~Randall, ``Lsd: A fast line
  segment detector with a false detection control,'' \emph{TPAMI}, vol.~32,
  no.~4, pp. 722--732, 2010.

\bibitem{Chang:2012:ALA}
C.-H. Chang and Y.-Y. Chuang, ``A line-structure-preserving approach to image
  resizing,'' in \emph{CVPR}, 2012.

\bibitem{He:2013:RPI}
K.~He, H.~Chang, and J.~Sun, ``Rectangling panoramic images via warping,''
  \emph{ACM TOG}, vol.~32, no.~4, p.~79, 2013.

\bibitem{Li:2015:AGP}
D.~Li, K.~He, J.~Sun, and K.~Zhou, ``A geodesic-preserving method for image
  warping,'' in \emph{CVPR}, 2015.

\bibitem{bouguet2001pyramidal}
J.-Y. Bouguet, ``Pyramidal implementation of the affine lucas kanade feature
  tracker description of the algorithm,'' \emph{Intel Corporation}, vol.~5, no.
  1-10, p.~4, 2001.

\bibitem{liu2016ssd}
W.~Liu, D.~Anguelov, D.~Erhan, C.~Szegedy, S.~Reed, C.-Y. Fu, and A.~C. Berg,
  ``Ssd: Single shot multibox detector,'' in \emph{ECCV}, 2016.

\bibitem{Ronneberger:2015:UNC}
O.~Ronneberger, P.~Fischer, and T.~Brox, ``U-net: Convolutional networks for
  biomedical image segmentation,'' in \emph{International Conference on Medical
  image computing and computer-assisted intervention}, 2015.

\bibitem{Tkachenka:2019:RTH}
A.~Tkachenka, G.~Karpiak, A.~Vakunov, Y.~Kartynnik, A.~Ablavatski,
  V.~Bazarevsky, and S.~Pisarchyk, ``Real-time hair segmentation and recoloring
  on mobile gpus,'' \emph{arXiv:1907.06740}, 2019.

\bibitem{Fong:2011:LAI}
D.~C.-L. Fong and M.~Saunders, ``Lsmr: An iterative algorithm for sparse
  least-squares problems,'' \emph{SIAM Journal on Scientific Computing},
  vol.~33, no.~5, pp. 2950--2971, 2011.

\bibitem{Sharpless:2010:PAN}
T.~K. Sharpless, B.~Postle, and D.~M. German, ``Pannini: a new projection for
  rendering wide angle perspective images,'' in \emph{International conference
  on Computational Aesthetics in Graphics, Visualization and Imaging}, 2010.

\bibitem{mercator}
``Mercator projection,''
  \url{https://en.wikipedia.org/wiki/Mercator_projection}.

\bibitem{moment}
``Moment lens,'' \url{https://www.shopmoment.com/shop/wide-18-mm-lens}, 2019.

\bibitem{osmo}
``Dji osmo,'' \url{https://www.dji.com/osmo}, 2019.

\end{thebibliography}
}

\begin{IEEEbiography}[{\includegraphics[width=1in,height=1.25in,clip,keepaspectratio]{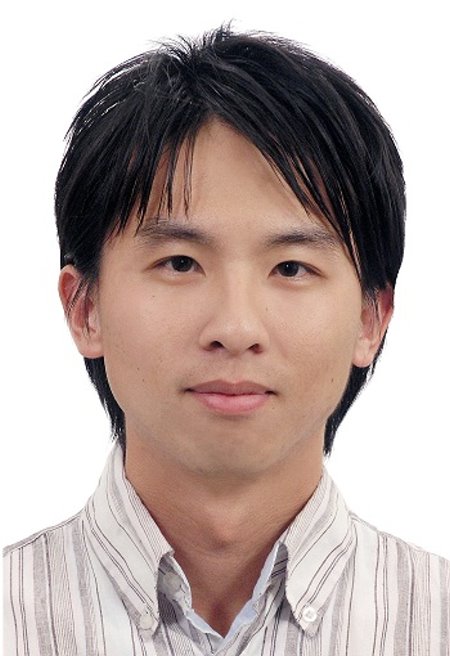}}]{Wei-Sheng Lai}
	is a software engineer at Google, USA. He received the B.S. and M.S. degree in Electrical Engineering from the National Taiwan University, Taipei, Taiwan, and his PhD degree in Electrical Engineering and Computer Science at the University of California Merced in 2019.
\end{IEEEbiography}

\begin{IEEEbiography}[{\includegraphics[width=1in,height=1.25in,clip,keepaspectratio]{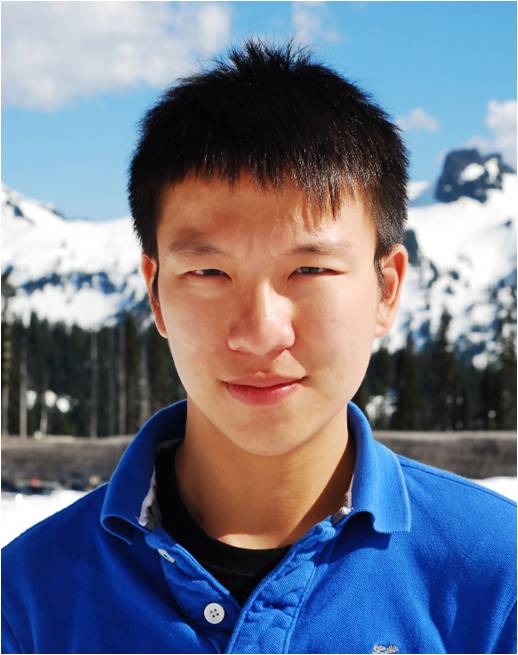}}]{YiChang Shih} is a software engineer at Google since 2017. Prior to Google, YiChang worked at Light on small form-factor multi-lens camera system. He received the PhD degree from MIT CSAIL in 2015.
\end{IEEEbiography}

\begin{IEEEbiography}[{\includegraphics[width=1in,height=1.25in,clip,keepaspectratio]{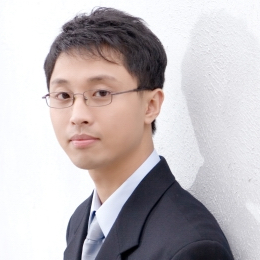}}] {Chia-Kai Liang} is a software engineer at Google since 2015. He worked at Lytro on consumer light field cameras from 2010 to 2015. He received the PhD degree from National Taiwan University, Taipei, Taiwan in 2009. Dr. Liang is an associated editor of the IEEE Transactions on Image Processing and recipient of IEEE Trans. CSVT Best Paper Award in 2008.
\end{IEEEbiography}

\begin{IEEEbiography}[{\includegraphics[width=1in,height=1.25in,clip,keepaspectratio]{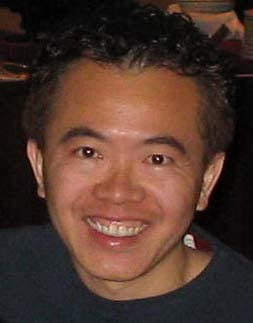}}]{Ming-Hsuan Yang}
	is a research scientist at Google. He received the PhD degree in computer science from the University of Illinois at Urbana-Champaign in 2000. 
	He received the best paper honorable mention from IEEE Conference on Computer Vision and Pattern Recognition in 2018.
	Yang served as a program co-chair of IEEE International Conference on Computer Vision in 2019.
	He is a Fellow of the IEEE. 
\end{IEEEbiography}

\end{document}